\documentclass[format=acmsmall, review=false]{acmart}
\usepackage{acm-ec-23}
\usepackage{algorithmic}
\usepackage[ruled,vlined]{algorithm2e}
\usepackage[utf8]{inputenc} 
\usepackage[T1]{fontenc}    
\usepackage{hyperref}       
\usepackage{url}            
\usepackage{booktabs}       
\usepackage{amsfonts}       
\usepackage{nicefrac}       
\usepackage{microtype}      
\usepackage{wrapfig}
\usepackage{xcolor}    
\usepackage{mathtools}
\usepackage{bbm}
\usepackage{caption} 
\usepackage{subcaption}
\usepackage{pbox}
\usepackage{amsmath}

\DeclareMathOperator*{\argmin}{arg\,min}
\usepackage{amsthm}

\usepackage{float}

\newtheorem*{remark}{Remark}

\usepackage{amsmath,amsfonts,bm}

\def\eqref#1{equation~\ref{#1}}

\def\1{\bm{1}}

\def\vb{{\bm{b}}}
\def\vc{{\bm{c}}}

\def\vf{{\bm{f}}}
\def\vg{{\bm{g}}}
\def\vh{{\bm{h}}}

\def\vw{{\bm{w}}}
\def\vx{{\bm{x}}}
\def\vy{{\bm{y}}}
\def\vz{{\bm{z}}}

\def\mA{{\bm{A}}}
\def\mB{{\bm{B}}}

\def\mI{{\bm{I}}}

\DeclareMathAlphabet{\mathsfit}{\encodingdefault}{\sfdefault}{m}{sl}
\SetMathAlphabet{\mathsfit}{bold}{\encodingdefault}{\sfdefault}{bx}{n}

\setcitestyle{authoryear}

\title[Achieving Better Regret against Strategic Adversaries]{Achieving Better Regret against Strategic Adversaries}

\author{Le Cong Dinh, Tri-Dung Nguyen, Alain Zemkoho, Long Tran-Thanh}

\begin{abstract}
We study online learning problems in which the learner has extra knowledge about the adversary's behaviour, i.e., in game-theoretic settings where opponents typically follow some no-external regret learning algorithms. Under this assumption, we propose two new online learning algorithms, Accurate Follow the Regularized Leader (AFTRL) and Prod-Best Response (Prod-BR), that intensively exploit this extra knowledge while maintaining the no-regret property in the worst-case scenario of 
having inaccurate extra information. Specifically, AFTRL achieves $O(1)$ external regret or $O(1)$ \emph{forward regret} against no-external regret adversary in comparison with $O(\sqrt{T})$ \emph{dynamic regret} of Prod-BR. To the best of our knowledge, our algorithm is the first to consider forward regret that achieves $O(1)$ regret against strategic adversaries. When playing zero-sum games with Accurate Multiplicative Weights Update (AMWU), a special case of AFTRL, we achieve \emph{last round convergence} to the Nash Equilibrium. We also provide numerical experiments to further support our theoretical results. In particular, we demonstrate that our methods achieve significantly better regret 
bounds and rate of last round convergence, compared to the state of the art (e.g., Multiplicative Weights Update (MWU) and its optimistic counterpart, OMWU).
\end{abstract}

\begin{document}

\begin{titlepage}

\maketitle

\end{titlepage}

\section{Introduction}
No-regret algorithms are popular in the online learning and algorithmic game theory literature due to their attractive worst-case performance guarantees \cite{cesa2006prediction}. In particular, using these algorithms to choose the strategies to play provably guarantees the average payoff will not be (significantly) worse than the best-fixed strategy in the hindsight, regardless of the sequences encountered. Due to this property, these no-regret algorithms are commonly used in playing against adversary and solving two-player zero-sum games, in which it will eventually lead to average convergence to a Nash Equilibrium (NE) under self-play settings~\cite{zinkevich2007regret,lanctot2017unified,le2021online}. However, in order to keep the regret bound small, no-regret algorithms (e.g., Multiplicative Weights Update, Follow the Regularized Leader, Mirror Descent~\cite{abernethy2009competing,nemirovskij1983problem}) need to keep their learning rate small, leading to a slow change in the strategy profile. This makes the sequence of strategies played by no-regret algorithms predictable since each strategy profile will be correlated to its predecessors.  Thus, against a no-regret learning opponent, the loss sequence encountered by the learner/player is not entirely arbitrarily adversarial in each round and therefore the worst-case performance guarantees are too pessimistic for the learner. Therefore, in situations such as playing against no-regret algorithms (strategic adversaries), it is desirable to develop a learning algorithm that can exploit the extra structure while maintaining the no-regret property in the worst-case scenario and answer the question:
\[\emph{ Can we exploit no-regret algorithms?}\]
Besides aiming for better regret bounds, we are also interested in last round convergence 
instead of just average convergence to the NE. In two-player zero-sum games, no-regret algorithms such as Multiplicative Weights Update (MWU)~\cite{Freund99} or Follow the Regularized Leader (FTRL) will only lead to average convergence instead of last round convergence. 
In fact, recent results in~\cite{Bailey2018,mertikopoulos2018} show that MWU and FTRL will lead to divergence from the NE in many situations. The average convergence will not only increase the computational and memory overhead but also make things difficult when using a neutral network in the solution process in which averaging is not always possible~\cite{bowling2015heads}. For game theory and modern applications of online learning in optimization such as training Generative Adversarial Networks~\cite{daskalakis2017training}, last round convergence plays a vital role in the process, thus it is crucial to develop algorithms that can lead to last round convergence.

To investigate both of the above-mentioned goals in this paper, under the setting of online linear optimization, we \textbf{first}
develop a new algorithm, Accurate Follow the Regularized Leader (AFTRL), that can exploit no-external regret adversary to achieve $O(1)$ external regret or $O(1)$ forward regret while maintaining state-of-the-art regret bound of $O\Big(\sqrt{\sum_{t=1}^T \|\vx_t-\vx_{t-1}\|_{q}^2}\Big)$ in the worst-case scenario. We also show the generality of our method by extending the result to another online learning class and propose a new algorithm, Accurate Mirror Descent (AMD) with a similar forward regret bound for it. To the best of our knowledge, we are the first to consider \emph{intensive exploitation} and achieve $O(1)$ \emph{forward regret} against no-external regret adversary. \textbf{Secondly}, 
we explore the idea of (A,B)-Prod algorithm in \citep{sani2014exploiting} and suggests a new algorithm, Prod-Best Response (Prod-BR) that achieves a stronger performance guarantee in our setting. In particular, Prod-BR achieves $O(\sqrt{T})$ dynamic regret against no-external regret adversary while maintaining $O(\sqrt{T \log(T)})$ external regret in the worst case. \textbf{Thirdly}, in a special case of AFTRL with entropy regularizer, called Accurate Multiplicative Weights Update (AMWU), we prove that this new algorithm  will lead to last round convergence in two-player zero-sum games, thus can be an efficient game-solver in many practical applications. In addition, this provides novel contributions to the last round convergence literature. 
\textbf{Finally}, to demonstrate the practical efficiency of AMWU, we show that our algorithm significantly outperforms MWU and OMWU~\cite{rakhlin2013online,Daskalakis2018c}  on a number of random matrix games and meta games such as Connect Four or Disc~\cite{czarnecki2020real} by a large margin, achieving smaller average loss, dynamic regret and faster last round convergence. 


\begin{section}{Related Work}
\textbf{Online learning against no-regret learners:} Deng et al.~\cite{deng2019strategizing} studies a similar setting in which the agent plays against a no-external regret adversary in a repeated game. Under the assumption that the agent knows the game structure (i.e., payoff matrix, player's utility), \citep{deng2019strategizing} suggested a fixed strategy for the agent (through solving an optimization problem) such that the agent can guarantee a Stackelberg value, which is optimal in certain games (e.g., general-sum games). Although the work in \citep{deng2019strategizing} provides a planning solution against no-external regret adversary, it can not be applied in many practical situations in which the environment or game structure is unknown (i.e., the agent can not calculate the Stackelberg strategy in advance) or the adversary does not follow no-regret algorithms (i.e., there is no performance guarantee against general adversary). Chiang et al.~\cite{chiang2012online} and Rakhlin et al.~\cite{rakhlin2013online} study a different setting in which the agent has access to the prediction $M_t$ of $\vx_t$ before making a decision at round $t$.~\footnote{As we prove in Lemma \ref{lemma: consecutive strategis of no-regret algorithm}, playing against strategic adversary can result in an accurate prediction of $\vx_t$.} The new algorithm, Optimistic Follow the Regularized Leader (OFTRL), has the external regret that depends linearly on $\sqrt{\sum_{t=1}^T \|\vx_t-M_t\|_{*}^2}$. However, with an accurate prediction (i.e., $M_t\approx \vx_t$), one could expect a stronger performance guarantee rather than no-external regret of OFTRL. Intuitively, since OFTRL sets a fixed weight $1$ for prediction $M_t$~\footnote{The exploiting rate $\alpha$ in Algorithm \ref{alg: AFTRL}.}, it restricts the advantage of the extra knowledge in the learning process. Our new algorithms (AFTRL and AMD) generalize the work of \cite{rakhlin2013online} to further exploit the extra knowledge in the learning process while maintaining a no-forward regret property~\citep{saha2012interplay} in the worst-case scenario. 

\textbf{Last round convergence: }While average convergence of no-regret learning dynamics has been studied extensively in game theory and online learning communities (e.g.,~\cite{Freund99,cesa2006prediction}), last round convergence has only been a topic of research in the last few years due to its application in game theory and optimization.  This started with the negative result of~\cite{Bailey2018,mertikopoulos2018}, who showed that in games with interior equilibria, if the agents use MWU, then the last round strategy moves away from the NE and towards the boundary. More recently, \cite{Daskalakis2018c,wei2020linear} proved that in a two-player zero-sum game with unique NE, if both players follow a variant of MWU, called optimistic multiplicative weight update (OMWU), then the dynamic will converge in last round to the NE. 
In asymmetric setting, \cite{dinh2020last} proposed last round convergence in asymmetric games algorithm (LRCA), which requires one agent to have an estimate of the minimax equilibrium and therefore limit the use of the algorithm. In our work, we prove that 
our method AMWU will converge in last round to the NE of a two-player zero-sum game without such a requirement, and it does this faster than OMWU and MWU. 
\end{section}
\begin{section}{Preliminary}
We consider the online linear optimization setting in which at round $t$, the learner chooses a strategy $\vf_t \in \mathcal{F}$, where $ \mathcal{F} \subset [0,1]^n$~\footnote{All the results remains true for bounded domain of strategy and loss vector.} is a convex compact set. Simultaneously, the environment reviews a loss vector $\vx_t \in [0,1]^n$ and the learner suffers the loss: $\langle \vf_t, \vx_t \rangle$. 
The goal of the learner is to minimize the total loss after $T$ rounds: $\min_{\vf_1,\dots \vf_T} \sum_{t=1}^T \langle \vf_t, \vx_t \rangle$, which can be translated into minimizing the following dynamic regret: 
  
\begin{definition}[\textbf{Dynamic Regret}~\citep{besbes2015non}]
The dynamic regret is defined as:
\begin{equation*}
    \begin{aligned}
DR_T:=\sum_{t=1}^T   \left( \langle \vf_t,\vx_t \rangle - \argmin_{\vg \in \mathcal{F}}\langle \vg, \vx_t \rangle\right)    
    \end{aligned}
\end{equation*}
\end{definition}
In situations where there is no knowledge about $\vx_t$, it is often impossible to achieve no-dynamic regret. Thus, it is more tractable to aim for no-external regret~\cite{cesa2006prediction}:
\begin{definition}[\textbf{No-external regret}]
Let $\vx_1, \vx_2, \ldots$ be a sequence of mixed losses played by the environment. An algorithm of the learner that generates a sequence of mixed strategies $\vx_1, \vx_2, \ldots$ is called a {\em no-external regret} algorithm if we have: 
\begin{equation*}
    \lim_{T \rightarrow \infty} \frac{R_T}{T} = 0, \; \text{where} \; R_T := \min_{\vf \in \mathcal{F}}\sum_{t=1}^\top   \left( \langle \vf_t,\vx_t \rangle - \langle \vf, \vx_t \rangle\right).
\end{equation*}
\vspace{-10pt}
\end{definition}
In this paper, since we assume the learner has extra knowledge about the adversary, the learner can achieve a stronger notion of performance, compared to the conventional no-external regret, namely:
\begin{definition}[\textbf{Forward Regret}~\citep{saha2012interplay}]
The forward regret is defined as:
\begin{equation*}
\begin{aligned}
FR_T := \sum_{t=1}^T   \left( \langle \vf_t,\vx_t \rangle - \langle \vg_t, \vx_t \rangle\right), \text{where}\; \vg_{t+1}= \argmin_{\vg \in \mathcal{F}} G_{t+1}(\vg)= \langle \vg, \sum_{s=1}^t \vx_s +\vx_{t+1}\rangle +\frac{R(\vg)}{\eta}. 
\end{aligned}
\end{equation*}
\end{definition}
In particular, the following lemma implies that if an algorithm has no-forward regret property, then it is a no-external regret algorithm as well, but not vice versa~\footnote{ See \ref{proof: lemma forward regret vs external regret} for the proof of this lemma.}.

\begin{lemma}\label{lemma: forward regret vs external regret}
Let $\vg_t$ be defined as above, then the following relationship holds for any $\vf \in \mathcal{F}$:
\[\sum_{t=1}^T \langle \vg_t, \vx_t \rangle \leq \langle \vf, \sum_{t=1}^T \vx_t \rangle + \frac{R(\vf)}{\eta}.\]
\end{lemma}

In Section \ref{sec:last round convergence}, we study a simpler form of online linear optimization in which the loss function has the form: $\mA^T \vy$ where $\vy$ is a point in the simplex $\Delta_m$. We also consider $\mathcal{F}$ to be the simplex $\Delta_n$ and the game is often referred as the zero-sum matrix game $\mA$. The NE in two-player zero-sum game $\mA$ can be expressed by John von Neumann's minimax theorem~\cite{Neumann1928}:
\begin{equation} \label{minimax theorem}
\max_{\vy\in \Delta_m}\min_{\vf \in \Delta_n} \vf^\top \mA \vy  = \min_{\vf \in \Delta_n}\max_{\vy\in \Delta_m} \vf^\top \mA \vy =  v
\end{equation}
for some $v\in \mathbb{R}$. The point $(\vf^*, \vy^*)$ satisfying Equation (\ref{minimax theorem}) is the NE of the game.
\end{section}
\begin{section}{Accurate Follow the Regularized Leader}
In order to have a no-(external) regret property, popular algorithms such FTRL and OMD need to have small learning rate $\eta$ (i.e., see \citep{shalev2012online}): $\eta=O(\frac{1}{\sqrt{T}})$. From this observation, we can prove the following lemma, which plays an important role in our analyses:
\begin{lemma}\label{lemma: consecutive strategis of no-regret algorithm}
Let $\vf_t$, $\vf_{t+1}$ be two consecutive strategies of no-external regret algorithms (i.e., FTRL, OMD). Then we have for any norm $\|.\|_q$:
\[\|\vf_{t+1}-\vf_t\|_q =O(\frac{1}{\sqrt{T}}). \quad\]
The full proof is given in Appendix \ref{Proof of lemma: consecutive strategis of no-regret algorithm}.
\end{lemma}

Now, 
let $R$ be $\beta$-strongly convex function with respect to $\|.\|_p$ norm. W.l.o.g.~we assume that $\min_{\vf \in \mathcal{F}} R(\vf)=0$.

\begin{algorithm}[H]
\textbf{Input:} \text{learning rate $\eta > 0$,
exploiting rate $\alpha \geq 1$,}

$\vf_1=\argmin_{\vf \in \mathcal{F}}R(\vf)$.

\textbf{Output:} \text{next strategy update}
\[\vf_{t+1}=\argmin_{\vf \in \mathcal{F}} F_{t+1}(\vf)= \langle \vf, \sum_{s=1}^t \vx_s +\alpha x_{t}\rangle + \frac{R(\vf)}{\eta}.\]
\caption{Accurate Follow the Regularized Leader}\label{alg: AFTRL}
\end{algorithm}

The Accurate Follow the Regularized Leader algorithm (AFTRL) contains two important parameters: the exploiting rate $\alpha$ and the learning rate $\eta$. While the learning rate $\eta$ stabilizes the strategy update to avoid exploitation, the exploiting rate $\alpha$ measures the relative 
weights between the historical data $\sum_{s=1}^t \vx_s$ and the prediction $\vx_t$. Intuitively, with an accurate prediction $\vx_t$, a large $\alpha$ will boost the performance of AFTRL since $\vx_{t}$ describes the next loss vector $\vx_{t+1}$ better compared to the historical data $\sum_{s=1}^{t}\vx_s$. Varying $\alpha$ provides different algorithms in the literature.  With $\alpha =0$, the algorithm becomes the classical FTRL~\cite{abernethy2009competing}. With $\alpha =1$, AFTRL recovers the optimistic FTRL method (OFTRL) of~\cite{rakhlin2013online}. 
We can have the following regret bound of the AFTRL algorithm:
\begin{theorem}\label{thm: AFTRL for convec regularizer}
Let $ \mathcal{F} \subset [0,1]^n$ be a convex compact set and let $R$ be a $\beta$-strongly convex function with respect to $\|.\|_p$ norm and $\min_{\vf \in \mathcal{F}} R(\vf)=0.$ Denote $\|.\|_q$ the dual norm with $1/p+1/q=1$. Then the AFTRL achieves the external regret of $O(1)$ or forward regret of $O\Big(\sqrt{\sum_{t=1}^T (\|\vx_t-\vx_{t-1}\|_q)^2}\Big)$ against general adversary. More importantly, against no-external regret adversary (i.e., FTRL, OMD), AFTRL achieves $O(1)$ external regret or $O(1)$ forward regret.
\end{theorem}
\begin{proof}[Proof Sketch]
We first prove that for any strategy of the environment, AFTRL satisfies:
\begin{equation}\label{ineq: regret bound of AFTRL}
    \begin{aligned}
    \sum_{t=1}^T \langle \vf_t,\vx_t \rangle-\frac{1}{\alpha}\langle \vf', \sum_{t=1}^T \vx_t \rangle -\frac{\alpha-1}{\alpha}\sum_{t=1}^T\langle \vg_{t}, \vx_t \rangle
    \leq \frac{1}{\eta \alpha}R(\vf') +\frac{\eta \alpha}{\beta} \sum_{t=1}^T (||\vx_t-\vx_{t-1}||_{q})^2. 
    \end{aligned}
\end{equation}
Define $\vh_{t+1}$ as follows:
$\vh_{t+1}= \argmin_{\vf \in \mathcal{F}} H_{t+1}(\vf)=\langle \vf, \sum_{s=1}^t \vx_s +\alpha \vx_{t+1}\rangle +\frac{R(\vf)}{\eta}.$

Intuitively, the strategy $\vh_{t+1}$ will perform much better than the normal FTRL since the agent can observe one step ahead the strategy of the adversary. Note that we can decompose the total loss of the agent as follows
\begin{equation}\label{eq: decomposition}
\begin{aligned}
 \sum_{t=1}^T \langle \vf_t,\vx_t \rangle= \sum_{t=1}^T\langle \vf_t-\vh_t,\vx_t-\vx_{t-1}\rangle 
 \sum_{t=1}^T\langle \vf_t-\vh_t, \vx_{t-1} \rangle +\sum_{t=1}^T\langle \vh_t, \vx_t \rangle.
\end{aligned}
\end{equation}
The key step of the proof is that we can prove by induction:
\begin{equation}\label{inequation 1.1 2}
\begin{aligned}
&\sum_{t=1}^T\langle \vf_t-\vh_t, \vx_{t-1} \rangle +\sum_{t=1}^T\langle \vh_t, \vx_t \rangle \leq
\frac{1}{\alpha}\langle \vf', \sum_{t=1}^T \vx_t \rangle +\frac{\alpha-1}{\alpha}\sum_{t=1}^T\langle \vg_{t}, \vx_t \rangle + \frac{1}{\eta \alpha}R(\vf'), \; \forall \vf'\in \mathcal{F}. 
\end{aligned}
\end{equation}
Furthermore, using the property of $\beta$-strongly convex function, we can derive:
\begin{equation}\label{eq: strongly convex property}
    \begin{aligned}
     &\|\vx_{t-1}-\vx_t\|_q \geq \frac{\beta}{\eta \alpha} \|\vh_t-\vf_t\|_p \\
     &\implies \sum_{t=1}^T\langle \vf_t-\vh_t,\vx_t-\vx_{t-1}\rangle \leq \sum_{t=1}^T ||\vf_t-\vh_t||_p ||\vx_t-\vx_{t-1}||_{q}
     \leq \frac{\eta \alpha}{ \beta}\sum_{t=1}^T (||\vx_{t-1}-\vx_t||_{q})^2.
    \end{aligned}
\end{equation}
Using Inequality (\ref{inequation 1.1 2}) and (\ref{eq: strongly convex property}) in Equality (\ref{eq: decomposition}) we derive the Inequality (\ref{ineq: regret bound of AFTRL}). 

Now, against general adversary, if $\sum_{t=1}^T \langle \vf_t,\vx_t \rangle-\langle \vf', \sum_{t=1}^T \vx_t \rangle \leq 0$ then by definition, AFTRL has $O(1)$ external regret. In case where $\sum_{t=1}^T \langle \vf_t,\vx_t \rangle-\langle \vf', \sum_{t=1}^T \vx_t \rangle \geq 0$, using Inequality (\ref{ineq: regret bound of AFTRL}) and setting $\eta \alpha = \sqrt{\beta R(\vf')/(\sum_{t=1}^T (\|\vx_t-\vx_{t-1}\|^{}_q)^2)}$ we have:
\[\sum_{t=1}^T \langle \vf_t,\vx_t \rangle-\sum_{t=1}^T\langle \vg_{t}, \vx_t \rangle\leq \frac{\alpha}{\alpha-1}\sqrt{R(\vf') \sum_{t=1}^T (\|\vx_t-\vx_{t-1}\|_q)^2/\beta}= O(\sum_{t=1}^T (\|\vx_t-\vx_{t-1}\|_q)^2).\]
For unknown bound $\sum_{t=1}^T (\|\vx_t-\vx_{t-1}\|_q)^2$, we can use the Doubling Trick as shown in Appendix \ref{Doubling Trick} to achieve a similar regret bound.

Against a no-external regret adversary, using Lemma \ref{lemma: consecutive strategis of no-regret algorithm}, we then have:
\[\sum_{t=1}^T (\|\vx_t-\vx_{t-1}\|_q)^2=\sum_{t=1}^T (O(\frac{1}{\sqrt{T}}))^2=O(1).\]
Thus, Inequality (\ref{ineq: regret bound of AFTRL}) becomes:
\[\sum_{t=1}^T \langle \vf_t,\vx_t \rangle-\frac{1}{\alpha}\langle \vf', \sum_{t=1}^T \vx_t \rangle -\frac{\alpha-1}{\alpha}\sum_{t=1}^T\langle \vg_{t}, \vx_t \rangle
    \leq \frac{1}{\eta \alpha}R(\vf') +\frac{\eta \alpha}{\beta} O(1)=O(1).\]
Following a similar reasoning for general adversary, AFTRL achieves $O(1)$ external regret or $O(1)$ forward regret against no-external regret adversary.
The full proof is given in Appendix \ref{Appendix: thm AFRL for convec regularizer}.
\end{proof}

\begin{remark}[\textbf{AFTRL vs OFTRL}]
While both AFTRL and OFTRL share the same idea of exploiting ``predictable sequences", they are significantly different. Firstly, the level of dependency on predictable sequences in OFTRL is fixed to $1$, whereas AFTRL allows a flexible control over the predictable sequences (i.e., via parameter $\alpha$). Thus, AFTRL can achieve much better performance in situation of accurate prediction compared to OFTRL, which can be reassured by experiment results in Figure \ref{fig:average loss in games}. Secondly, in the worst case scenario, AFTRL can guarantee a stronger forward regret bound compared to external regret bound of OFTRL in \cite{rakhlin2013online}. 
\end{remark}
Our techniques can be extended to a different class of algorithm such as Mirror Descent\citep{shalev2012online}. We introduce Accurate Mirror Descent (AMD)~\footnote{The Pseudocode is given in Appendix \ref{appendix: AMD}} with a similar regret bound as AFTRL:
\begin{theorem}\label{thm: MD with aggressive prediction}
Let $\mathcal{F}$ be a convex set in a Banach space $\mathcal{B}$. Let $\mathcal{R}: \mathcal{B} \to \mathbb{R}$ be a $\beta$-strongly convex function on $\mathcal{F}$ with respect to some norm $\|.\|_p$. Denote $\|.\|_q$ the dual norm with $1/p+1/q=1$.  For any strategy of the environment and any $\vf' \in \mathcal{F}$, AMD yields
\begin{equation*}
    \begin{aligned}
     \sum_{t=1}^T \langle \vf_t,\vx_t \rangle -\frac{1}{\alpha} \langle \vf', \vx_t \rangle-\frac{\alpha-1}{\alpha} \langle \vg_{t+1}, \vx_t \rangle
     \leq \frac{\eta \alpha}{2\beta}\sum_{t=1}^T \|\vx_t-\vx_{t-1}\|_q^2 + \frac{\mathcal{R}_{max}^2}{\eta \alpha},
    \end{aligned}
\end{equation*}

where $\mathcal{R}_{max}^2= \max_{\vf \in \mathcal{F}}\mathcal{R}(\vf)-\min_{\vf \in \mathcal{F}}\mathcal{R}(\vf)$.
\end{theorem}
\begin{proof}
We define $\vh_{t+1}$ as follow:
\[\vh_{t+1}=\argmin_{\vh\in \mathcal{F}} H_{t+1}(\vh)= \eta \langle \vh, \alpha \vx_t \rangle + D_{\mathcal{R}}(\vh,\vg_t).\]

For any $\vf' \in \mathcal{F}$,

\begin{equation}\label{BMD eq 1}
\begin{aligned}
    &\langle \vf_t,\vx_t \rangle -\frac{1}{\alpha} \langle \vf', \vx_t \rangle-\frac{\alpha-1}{\alpha}\langle \vg_{t+1}, \vx_t \rangle \\
    &=\langle \vf_t-\vh_{t+1},\vx_t-\vx_{t-1} \rangle + \langle \vf_t-\vh_{t+1}, \vx_{t-1} \rangle \\
    &+\langle \vh_{t+1}-\vg_{t+1},\vx_t \rangle + \frac{1}{\alpha} \langle \vg_{t+1}-\vf', \vx_t \rangle.
\end{aligned}    
\end{equation}

Using property of dual norm, we derive
\begin{equation}\label{BMD eq 2}
\begin{aligned}
 \langle \vf_t-\vh_{t+1},\vx_t-\vx_{t-1} \rangle \leq \|\vf_t-\vh_{t+1}\|_p \|\vx_{t-1}-\vx_t\|_q \\
 \leq \frac{\beta}{2\eta \alpha}\|\vf_t-\vh_{t+1}\|_p^2 + \frac{\eta \alpha}{2\beta}\|\vx_{t-1}-\vx_t\|_q^2. 
\end{aligned}
\end{equation}
We note that for any $\vg \in \mathcal{F}$ and $\vf=\argmin_{\vf \in \mathcal{F}} \langle \vf, \vx \rangle + D_{\mathcal{R}}(\vf,\vc)$, we have the following inequalities (see e.g. \cite{beck2003mirror}):
\[\langle \vf - \vg,\vx \rangle \leq D_{\mathcal{R}}(\vg,\vc)-D_{\mathcal{R}}(\vg,\vf)-D_{\mathcal{R}}(\vf,\vc).\]
This yields
\begin{equation*}
    \begin{aligned}
     &\langle \vf_t-\vh_{t+1},\vx_{t-1} \rangle
     \leq \frac{1}{\eta \alpha}(D_{\mathcal{R}}(\vh_{t+1},\vg_t)-D_{\mathcal{R}}(\vh_{t+1},\vf_t)-D_{\mathcal{R}}(\vf_t,\vg_t)),\\
     &\langle \vh_{t+1}-\vg_{t+1},\vx_t \rangle 
     \leq \frac{1}{\eta \alpha}(D_{\mathcal{R}}(\vg_{t+1},\vg_t)-D_{\mathcal{R}}(\vg_{t+1},\vh_{t+1})-D_{\mathcal{R}}(\vh_{t+1},\vg_t)),\\
     &\langle \vg_{t+1}-\vf', \vx_t \rangle
     \leq \frac{1}{\eta}(D_{\mathcal{R}}(\vf', \vg_t)-D_{\mathcal{R}}(\vf', \vg_{t+1})-D_{\mathcal{R}}(\vg_{t+1},\vg_t)).
    \end{aligned}
\end{equation*}
Summing up the above inequalities we have
\begin{equation}\label{BMD eq 3}
    \begin{aligned}
    &\langle \vf_t-\vh_{t+1}, \vx_{t-1} \rangle +\langle \vh_{t+1}-\vg_{t+1},\vx_t \rangle + \frac{1}{\alpha} \langle \vg_{t+1}-\vf', \vx_t \rangle \\
    &\leq \frac{1}{\eta \alpha}(D_{\mathcal{R}}(\vf', \vg_t)-D_{\mathcal{R}}(\vf', \vg_{t+1})-D_{\mathcal{R}}\left(\vh_{t+1},\vf_t)
    -D_{\mathcal{R}}(\vf_t,\vg_t)-D_{\mathcal{R}}(\vg_{t+1},\vh_{t+1})\right).
    \end{aligned}
\end{equation}
Using the property of strongly convex function, we have
\begin{equation}\label{BMD eq 4}
   D_{\mathcal{R}}(\vh_{t+1},\vf_t) \geq \frac{\beta}{2} \|\vh_{t+1}-\vf_t\|_p^2; \quad D_{\mathcal{R}}(\vf_t,\vg_t) \geq \frac{\beta}{2} \|\vf_t-\vg_t\|_p^2.
\end{equation}
Putting Inequalities (\ref{BMD eq 2}),(\ref{BMD eq 3}) and (\ref{BMD eq 4}) in Equality (\ref{BMD eq 1}) we derive that
\begin{equation*}
    \begin{aligned}
    &\langle \vf_t,\vx_t \rangle -\frac{1}{\alpha} \langle \vf', \vx_t \rangle-\frac{\alpha-1}{\alpha}\langle \vg_{t+1}, \vx_t \rangle \leq \frac{\eta \alpha}{2\beta}\|\vx_{t-1}-\vx_t\|_q^2\\
    &+\frac{1}{\eta \alpha}(D_{\mathcal{R}}(\vf', \vg_t)-D_{\mathcal{R}}(\vf', \vg_{t+1}))-\frac{\beta}{2 \eta \alpha}\|\vf_t-\vg_t\|_p^2
    \end{aligned}
\end{equation*}
Summing over $t=1,\ldots,T$ yields, for any $\vf' \in \mathcal{F}$,
\begin{equation*}
\begin{aligned}
    &\sum_{t=1}^T \langle \vf_t,\vx_t \rangle -\frac{1}{\alpha} \langle \vf', \vx_t \rangle-\frac{\alpha-1}{\alpha} \langle \vg_{t+1}, \vx_t \rangle \\
    &\leq \frac{\eta \alpha}{2\beta}\sum_{t=1}^T \|\vx_t-\vx_{t-1}\|_q^2 + \frac{\mathcal{R}_{max}^2}{\eta \alpha}
    -\frac{\beta}{2\eta \alpha} \sum_{t=1}^T \|\vf_t-\vg_t\|_p^2\\
    &\leq \frac{\eta \alpha}{2\beta}\sum_{t=1}^T \|\vx_t-\vx_{t-1}\|_q^2 + \frac{\mathcal{R}_{max}^2}{\eta \alpha}.
\end{aligned}
\end{equation*}
where $\mathcal{R}_{max}^2= \max_{\vf \in \mathcal{F}}\mathcal{R}(\vf)-\min_{\vf \in \mathcal{F}}\mathcal{R}(\vf).$

Using the following inequality with any given $\vf' \in \mathcal{F}$ and $\vg_{t+1}=\argmin_{\vg \in \mathcal{F}} \eta \langle \vg, \vx_t \rangle + D_{\mathcal{R}}(\vg,\vg_t)$ (e.g., see \cite{beck2003mirror}):
\[\eta\langle \vg_{t+1}-\vf', \vx_t \rangle \leq D_{\mathcal{R}}(\vf', \vg_t)-D_{\mathcal{R}}(\vf', \vg_{t+1})-D_{\mathcal{R}}(\vg_{t+1},\vg_t)\]
we can derive that, for any $\vf' \in \mathcal{F}$,
\[\sum_{i=1}^T \langle \vg_{t+1},\vx_t \rangle \leq \sum_{i=1}^T \langle \vf', \vx_t \rangle +\frac{\mathcal{R}_{max}^2}{\eta} - \frac{\beta}{2\eta} \sum_{t=1}^T \|\vg_{t+1}-\vg_t\|^2.\]
Thus, the regret with respect to $\sum_{i=1}^T \langle \vg_{t+1},\vx_t \rangle$ (i.e., forward regret for AMD) is stronger than the (external) regret with respect to $\sum_{i=1}^T \langle \vf', \vx_t \rangle, \; \forall \vf' \in \mathcal{F}$.
\end{proof}
\section{Prod with Best Response}
While AFTRL gives us a guarantee of no-forward regret, one can wonder whether the agent can achieve a better performance (e.g., no-dynamic regret) given the extra knowledge? In this section, we introduce Prod with Best Response algorithm (Prod-BR) such that the agent can achieve no-dynamic regret against no-external regret adversary while maintaining a no-external regret performance in the worst case. Our variant Prod-BR algorithm gets motivation from (A,B)-Prod algorithm in \cite{sani2014exploiting}, in which we observe that the best response strategy from current feedback  can exploit a no-external regret adversary. The Prod-BR runs two separate algorithms (i.e., FTRL and BR) inside the main algorithm. Intuitively, while FTRL maintains a performance guarantee against the worst case scenario, BR algorithm exploits the extra structure against no-external regret adversary and thus make Prod-BR algorithm efficient. Prod-BR can balance between accurate and inaccurate extra knowledge so that the agent can achieve $O(\sqrt{T})$ dynamic regret against no-external regret adversary while maintaining $O(\sqrt{T}\log(T))$ external regret in the worst case scenario. 
\begin{algorithm}[h]
\textbf{Input:} \text{learning rate $\eta > 0$, $\eta_1 \in (0,1]$, initial weight $w_{1,R}, w_{1,BR},$ regularizer function $R(.)$.}

$\vf_{t+1}=\argmin_{\vf \in \mathcal{F}} F_{t+1}(\vf)= \langle \vf, \sum_{s=1}^t \vx_s \rangle + \frac{R(\vf)}{\eta}; \;\; BR_{t+1}= \argmin_{\vf \in \mathcal{F}} \langle f, \vx_t \rangle$

\textbf{Output:} \text{next strategy update $\vg_{t+1}$ and next weight $w_{t+1,R}$:}

\[\vg_{t+1}=\frac{w_{t, R}}{w_{t, R}+w_{1, BR}} \vf_{t+1}+\frac{w_{1, BR}}{w_{t, FTRL}+w_{1, BR}} BR_{t+1};\;\;w_{t+1,R}=w_{t,R}(1+\eta_1 \langle BR_{t+1}-\vf_{t+1}, \vx_{t+1} \rangle).\]
\caption{Prod-Best Response algorithm}\label{alg: Prod-BR}
\end{algorithm}
We first show that in the case where the adversary follows a no-external regret algorithm (i.e., FTRL, OMD) with optimal learning rate, then the best response with respect to the previous feedback can guarantee the agent the following:
\begin{lemma}\label{lm: BR performance each round}
Let $\vx_t$, $\vx_{t+1}$ be two consecutive strategies of a no-external regret algorithm (i.e., FTRL, OMD). Then, we have
 \[\langle \vb,\vx_{t+1} \rangle- \langle \vc, \vx_{t+1}\rangle = O(\frac{1}{\sqrt{T}}),\; \text{where}\; \vb=\argmin_{\vf \in \mathcal{F}} \langle \vf, \vx_{t}\rangle \;,\;\vc=\argmin_{\vf \in \mathcal{F}} \langle \vf, \vx_{t+1}\rangle.\]
 The full proof is given in Appendix \ref{proof of lm: BR performance each round}.
 \vspace{-5pt}
\end{lemma}

We then can prove the following theorem about the performance of Prod-BR algorithm:
\begin{theorem}
Let the agent follows Prod-BR Algorithm \ref{alg: Prod-BR} with $\eta=n/\sqrt{2T}$, $\eta_1=1/2.\sqrt{\log(T)/T}$ and $w_{1,BR}=1-w_{1,R}=1-\eta_1$. Then it achieves $O(\sqrt{T\log(T)})$ external regret against general adversary while maintaining $O(\sqrt{T})$ dynamic regret against no-external regret adversary.
\vspace{-5pt}
\end{theorem}
\begin{proof}
Following the regret bound analysis of (A,B)-Prod in Corollary 1 in \citep{sani2014exploiting} we have
\begin{subequations}
    \begin{align}
     &\sum_{t=1}^T \langle \vg_t, \vx_t \rangle \leq \sum_{t=1}^T \langle \vf_t, \vx_t \rangle + 2\sqrt{T \log(T)} \; \text{and} \label{eq: Prod-BR 1}\\
     &\sum_{t=1}^T \langle \vg_t, \vx_t \rangle \leq \sum_{t=1}^T \langle BR_t, \vx_t \rangle+ 2\log(2). \label{eq: Prod-BR 2}
    \end{align}
\end{subequations}
Since the agent uses the optimal learning rate for FTRL inside Algorithm \ref{alg: Prod-BR}, following the regret bound analysis of FTRL (i.e., see \citep{shalev2012online}) we have
\[\sum_{t=1}^T \langle \vf_t, \vx_t \rangle-\sum_{t=1}^T \langle \vf, \vx_t \rangle \leq n \sqrt{2T}\;\; \forall \vf \in \mathcal{F}.\]
Along with Inequality (\ref{eq: Prod-BR 1}) we have
\[\sum_{t=1}^T \langle \vg_t, \vx_t \rangle - \sum_{t=1}^T \langle \vf, \vx_t \rangle \leq 2\sqrt{T \log(T)}+n \sqrt{2T}=O(\sqrt{T \log(T)}) \;\; \forall \vf \in \mathcal{F},\]

or Prod-BR achieves $O(\sqrt{T \log(T)})$ external regret against general adversary. For the second part of the proof, using Inequality (\ref{eq: Prod-BR 2}) along with Lemma \ref{lm: BR performance each round} we have
\begin{equation*}
    \begin{aligned}
    &\sum_{t=1}^T \langle \vg_t, \vx_t \rangle -\argmin_{\vf \in \mathcal{F}} \langle \vf,\vx_t \rangle \leq \sum_{t=1}^T \langle BR_t, \vx_t \rangle - \argmin_{\vf \in \mathcal{F}} \langle \vf,\vx_t \rangle +2\log(2)\\
    &= \sum_{t=1}^T O(\frac{1}{\sqrt{T}})+ 2\log(2)
    =O(\sqrt{T}),
    \end{aligned}
\end{equation*}
or Prod-BR has $O(\sqrt{T})$ dynamic regret against no-external regret adversary.
\end{proof}
\begin{remark}[Prod-BR vs AFTRL]
In the worst case scenario, AFTRL provides a better performance guarantee over Prod-BR ($O(\sqrt{T})$ vs $O(\sqrt{T}\log(T))$). However, against no-external regret adversary, Prod-BR provides a much stronger notion of performance guarantee (no-dynamic regret) compared to no-forward regret of AFTRL. Note that both Prod-BR and AFTRL rely on the small distance between two consecutive strategies of the adversary. While it holds true for many no-external regret algorithms as in Lemma \ref{lemma: consecutive strategis of no-regret algorithm}, there are no-external regret algorithms (i.e., AdaHedge~\cite{de2014follow}) such as the distance between two consecutive strategies will have the form: $\|\vf_{t+1}-\vf_t\|_q=O(1/\sqrt{t})$ where $t$ denotes the current iteration. In this situation, following the same argument, AFTRL achieves $O(1)$ external regret or $O(\log(T))$ forward regret while Prod-BR maintains $O(\sqrt{T})$ dynamic regret.
\end{remark}


\section{Accurate Multiplicative Weights Update with Last Round Convergence}\label{sec:last round convergence}
\begin{algorithm}[H]
\textbf{Input:} \text{ learning rate $\eta >0$, exploiting rate $\alpha > 0$,}

\text{$\vf_1=\vf_2=[1/n,\dots,1/n].$}

\textbf{Output:} \text{Next update}
\begin{equation}\label{eq: update rule of AMWU}
    \begin{aligned}
    \vf_{t+1}(i)= \frac{\vf_{t}(i)e^{\eta ((\alpha+1){e_i}^\top \mA \vy_t-\alpha{e_i}^\top \mA\vy_{t-1})}}{\sum_j \vf_{t}(j) e^{\eta ((\alpha+1){e_j}^\top\mA \vy_t-\alpha{e_j}^\top \mA\vy_{t-1})}},
    \end{aligned}
\end{equation}
\text{$e_i$ denotes the unit-vector with weight of $1$ at $i$-component.}
\caption{Accurate Multiplicative Weights Update}
\end{algorithm}
We now turn to the second group of our contributions, namely: to ensure last round convergence with this new algorithmic framework. We show that if both players follow Accurate Multiplicative Weights Update (AMWU), a special case of AFTRL with 
entropy regularizer, then the dynamic converges last round to the NE in zero-sum game with unique NE.~\footnote{With some abuse of notation, in this section we use both $\vf(i)$ and $\vf_i$ to denote the $i$-th element of vector $\vf$.}

Note here that the uniqueness assumption of NE is generic in the following sense: since the set of zero-sum games with non-unique equilibrium has Lebesgue measure zero~\cite{van1991stability}, if the entries of $\mA$ are independently sampled from  some continuous distribution, then with probability one, the game has a unique NE. We leave the relaxation of the uniqueness assumption for future work.  
Our main last round convergence result is as follows:
 \begin{theorem}[\textbf{Last Round Convergence of AMWU}]\label{thm: last round convergence of AMWU}
Let $(\vf^*,\vy^*)$ be a unique Nash Equilibrium of the matrix game $\mA$. Then, with $\alpha=\eta^{b-1}$ for $b \in (0,1]$ and sufficiently small $\eta$, the dynamic of AMWU converges last round to the NE of the game: $\lim_{t \to \infty}(\vf_t,\vy_t)=(\vf^*,\vy^*)$.   
 \end{theorem}
 \begin{proof}[Proof of Sketch]
 We break the proof into three main parts.
First, 
in Section~\ref{sec:KL_distance}, we prove that the K-L divergence~\citep{KL1951} between the $t$-th strategy $(\vf_t,\vy_t)$ and $(\vf^*,\vy^*)$ will decrease by a factor of $\eta^{2+b}$ unless the strategy $(\vf_t,\vy_t)$ is $O(\eta^{b/3})$-close~\footnote{We later define it rigorously in Appendix \ref{appendix: missing definitions}}:
\begin{equation*}
    \begin{aligned}
    RE((\vf^*,\vy^*)||(\vf_{t+1},\vy_{t+1}))\leq
    RE((\vf^*,\vy^*)||(\vf_t,\vy_t)) -\Omega(\eta^{b+2}).
    \end{aligned}
\end{equation*}
The key step is the observation that the quantity $\vf_{t-1}^\top \mA \vy_t- \vf_t^\top \mA \vy_{t-1}$ can be bounded by:
\begin{equation*}
    \begin{aligned}
     &\eta \vf_{t-1}^\top \mA \vy_t-\eta \vf_t^\top \mA \vy_{t-1}
     =-\eta^2 \sum_{i}\vf_t(i)((\vf_t-e_i)^\top \mA ((\alpha+1)\vy_{t}-\alpha \vy_{t-1}))^2
     \\
     &-\eta^2 \sum_{i} \vy_{t}(i)((\vy_t-e_i)^\top \mA^\top((\alpha \vy_{t-1}-(\alpha+1)\vy_{t}))^2
     +O(\eta^{2+b}).
    \end{aligned}
\end{equation*}
From the above result, we then have that if the starting point is uniform (i.e., $\vf_1=(1/n,\dots,1/n)$ and $\vy_1=(1/m,\dots,1/m)$), AMWU will reach $O(\eta^{b/3})$-close in at most: $O\left(\frac{\log(nm)}{\eta^{2+b}}\right)$ time steps.

Second, in Section~\ref{sec:closeness}, we show that $\eta^{b/3}$-close point implies close to the NE with sufficiently small $\eta$. The proof comes closely related to the proof of Theorem 3.2 in \citep{Daskalakis2018c}. Thus, for any starting strategy with non-zero element and a sufficient small learning rate $\eta$, AMWU can get arbitrarily close to the NE. 

Finally, 
in Section~\ref{sec:local_convergence}, by proving that the spectral radius of the unique minimax equilibrium is less than one, we show that the update dynamic of AMWU is a locally converging on the NE point, meaning that there is last round convergence to the NE if the dynamic leads to a point in the neighborhood of the NE. 
Now, applying the first and second points to the dynamic of AMWU algorithm with non-zero element starting strategy, we have that AMWU will get arbitrarily close to the NE $(\vf^*,\vy^*)$ with a sufficiently small learning rate $\eta$. Then, using the locally converging property of AMWU, the last round convergence result in Theorem \ref{thm: last round convergence of AMWU} will follow directly. 

All the missing proofs can be found in Appendix \ref{appendix: missing proof of AMWU}. 

\end{proof}

We now provide the proof of the three key steps above. 
\subsection{Decreasing K-L distance}\label{sec:KL_distance} 
In this subsection, part of our analysis bases on the linear variant of AMWU with the following update rule:
\[\vf_{t+1}(i)= \frac{\vf_{t}(i)(1+\eta ((\alpha+1){e_i}^\top\mA \vy_t-\alpha{e_i}^\top \mA\vy_{t-1}))}{\sum_{j} \vf_{t}(j)(1+\eta ((\alpha+1){e_j}^\top\mA \vy_t-\alpha{e_j}^\top \mA\vy_{t-1}))}.\]
Since the variant' update rule does not contain the exponential part, it reduces the complexity in the analysis.
We first quantify the distance between two consecutive updates of AMWU by the following lemma:
\begin{lemma}\label{distance between two updates}
Let $\vf \in \Delta_n$ be the vector of the max player, $\vw,\vz \in \Delta_m$ such that $\|\vw-\vz\|_1=O(\eta)$, $\eta \alpha=O(1)$ and suppose $\vf', \vf''$ are the next iterates of AMWU and its linear variant with current vector $\vf$ and vectors $\vw, \vz$ of the min player. It holds that
\[\|\vf'-\vf''\|_1\; \text{is}\; O(\eta^2)\;\text{and}\; \|\vf'-\vf\|_1, \; \|\vf''-\vf\|_1\;\text{are}\; O(\eta).\]
Analogously, it holds for vector $\vy \in \Delta_m$ of the min player and its next iterates.

(The full proof is given in Appendix \ref{proof of distance between two updates}.)
\end{lemma}
When analysing the K-L divergence between the $t$-th strategy $(\vf_t,\vy_t)$ and $(\vf^*,\vy^*)$, we will encounter the quantity $\vf_{t-1}^\top \mA \vy_t- \vf_t^\top \mA \vy_{t-1}$. In order to bound this quantity, we need the following lemmas:
\begin{lemma}
Let $\vf \in \Delta_n$ be the vector of the max player, $\vw,\vz \in \Delta_m$ such that $\|\vw-\vz\|_1=O(\eta)$, $\eta \alpha=O(1)$ and suppose $\vf', \vf''$ are the next iterates of AMWU and its linear variant with current vector $\vf$ and vectors $\vw, \vz$ of the min player. It holds that (for $\eta$ sufficiently small)
\begin{equation*}
    \begin{aligned}
     &\;\;\eta (\vf'-\vf)^\top \mA ((\alpha+1)\vw-\alpha\vz)\\
     &=\eta (\vf''-\vf)^\top \mA ((\alpha+1)\vw-\alpha\vz)-O(\eta^3)\\
     &=(1-O(\eta))\eta^2 \sum_{i}\vf_i((\vf-e_i)^\top \mA ((\alpha+1)\vw-\alpha \vz))^2
     -O(\eta^3)\\
     &=(1-O(\eta))\eta^2 \sum_{i}\vf'_i((\vf'-e_i)^\top \mA ((\alpha+1)\vw-\alpha \vz))^2
     -O(\eta^3).
    \end{aligned}
\end{equation*}
\end{lemma}
\begin{proof}
By following Lemma \ref{distance between two updates}, we only need to prove the second equality. Set $\mB=(\mathbbm{1}_n \mathbbm{1}_m^\top+\eta\mA)$. We have that $f_i''=f_i\frac{(\mB((\alpha+1)\vw)-\alpha\vz)_i}{\vf^\top \mB((\alpha+1)\vw)-\alpha\vz)}$ following the definition of linear AMWU. We can derive that
\begin{equation*}
    \begin{aligned}
     &(\vf''^\top \mB((\alpha+1)\vw-\alpha\vz).(\vf^\top \mB((\alpha+1)\vw-\alpha\vz)\\
     &=\sum_{ij}\mB_{ij}\vf''_i((\alpha+1)\vw-\alpha\vz)_j.(\vf^\top \mB((\alpha+1)\vw-\alpha\vz)\\
     &=\sum_{ij}\mB_{ij} f_i\frac{(\mB((\alpha+1)\vw)-\alpha\vz)_i}{\vf^\top \mB((\alpha+1)\vw)-\alpha\vz)}((\alpha+1)\vw-\alpha\vz)_j.(\vf^\top \mB((\alpha+1)\vw-\alpha\vz)\\
     &=\sum_{ij}\mB_{ij} f_i(\mB((\alpha+1)\vw)-\alpha\vz)_i((\alpha+1)\vw-\alpha\vz)_j\\
     &=\sum_{i}f_i (\mB((\alpha+1)\vw-\alpha\vz)_i)^2\\
     &=(\vf^T \mB((\alpha+1)\vw-\alpha\vz))^2+\sum_{i}\vf_i(\vf^T \mB((\alpha+1)\vw-\alpha\vz)-(\mB((\alpha+1)\vw-\alpha\vz))_i)^2.
    \end{aligned}
\end{equation*}
Thus we have:
\begin{equation}\label{step 1 in Lemma 8}
    \begin{aligned}
    &(\vf''^\top \mB((\alpha+1)\vw-\alpha\vz).(\vf^\top \mB((\alpha+1)\vw-\alpha\vz)\\
    &=(\vf^T \mB((\alpha+1)\vw-\alpha\vz))^2+\sum_{i}\vf_i(\vf^T \mB((\alpha+1)\vw-\alpha\vz)-(\mB((\alpha+1)\vw-\alpha\vz))_i)^2.
    \end{aligned}
\end{equation}
Since our assumption that $\|\vw-\vz\|_1=O(\eta)$ and $\eta \alpha=O(1)$, we then have:
\[\|\mA ((\alpha+1)\vw-\alpha\vz)\|=\|\alpha\mA(\vw-\vz)+\mA \vw\|=O(\alpha \eta)+O(1)=O(1).\]
Thus we also have:
\[\vf^T \mB((\alpha+1)\vw-\alpha\vz)=1\pm O(\eta).\]
Devide both sides of Equation (\ref{step 1 in Lemma 8}) by $\vf^T \mB((\alpha+1)\vw-\alpha\vz)$ we have:
\begin{equation}
    \begin{aligned}
     &(\vf''^\top \mB((\alpha+1)\vw-\alpha\vz)\\
     &=(\vf^T \mB((\alpha+1)\vw-\alpha\vz))+(1-O(\eta))\sum_{i}\vf_i(\vf^T \mB((\alpha+1)\vw-\alpha\vz)-(\mB((\alpha+1)\vw-\alpha\vz))_i)^2\\
     &=\eta \vf^\top \mA ((\alpha+1)\vw-\alpha\vz)+(1-O(\eta))\eta^2\sum_{i}\vf_i((\vf-e_i)^T \mA((\alpha+1)\vw-\alpha\vz))^2.
    \end{aligned}
\end{equation}
Thus, the second equality is proven. Other equalities come directly as the result of Lemma \ref{distance between two updates}.
\end{proof}
Furthermore, from the above lemma, if we impose the condition:
\[\eta\alpha =\eta^b,\]
where $b$ is in $(0,1]$. Note that this condition does not contradict to $\eta\alpha=O(1).$ Then from the above lemma we have:
\begin{equation*}
    \begin{aligned}
     &\eta (\vf'-\vf)^\top \mA ((\alpha+1)\vw-\alpha\vz)\\
     &=\eta (\vf'-\vf)^\top \mA\vw +\eta \alpha (\vf'-\vf)^\top \mA (\vw-\vz)\\
     &=\eta (\vf'-\vf)^\top \mA\vw +\eta^b O(\eta^2)\\
     &\implies \eta (\vf'-\vf)^\top \mA\vw
     =(1-O(\eta))\eta^2 \sum_{i}\vf_i((\vf-e_i)^\top \mA ((\alpha+1)\vw-\alpha \vz))^2
     -\eta^b O(\eta^2)\\
     &=(1-O(\eta))\eta^2 \sum_{i}\vf'_i((\vf'-e_i)^\top \mA ((\alpha+1)\vw-\alpha \vz))^2
     -\eta^b O(\eta^2).
    \end{aligned}
\end{equation*}
Similarly, we have the following lemma for the min player:
\begin{lemma}
Let $\vy \in \Delta_m$, $\vw, \vz \in \Delta_n$ and suppose $\vy'$ is the next iterate of AMWU with current vector $\vy$ and inputs $\vw,\vz$. Furthermore, assume that $\|\vw-\vz\|_1=O(\eta)$ and $\eta \alpha=\eta^b$ for $0\leq b\leq 1$. It holds that (for $\eta$ sufficiently small):
\begin{equation*}
    \begin{aligned}
     &\eta (\vy'-\vy)^\top \mA^\top (-\vw)\\
     &=(1-O(\eta))\eta^2 \sum_{i} \vy'_i((\vy'-e_i)^\top \mA^\top((\alpha \vz-(\alpha+1)\vw))^2
     -\eta^b O(\eta^2).
    \end{aligned}
\end{equation*}
\end{lemma}
We then can prove the following lemma:
\begin{lemma}\label{Lemma 10}
Let $(\vf_t, \vy_t)$ be the t-th iteration of AMWU dynamic. For each time step $t\geq 2$ it holds that
\begin{equation*}
    \begin{aligned}
     &\eta \vf_{t-1}^\top \mA \vy_t-\eta \vf_t^\top \mA \vy_{t-1} =
     -\eta^2 \sum_{i}\vf_t(i)((\vf_t-e_i)^\top \mA ((\alpha+1)\vy_{t}-\alpha \vy_{t-1}))^2 \\
     &-\eta^2 \sum_{i} \vy_{t}(i)((\vy_t-e_i)^\top \mA^\top((\alpha \vy_{t-1}-(\alpha+1)\vy_{t}))^2
     +O(\eta^{2+b}).
    \end{aligned}
\end{equation*}
\end{lemma}
\begin{proof}

\begin{equation}
    \begin{aligned}
     &\eta \vf_{t-1}^\top \mA \vy_t-\eta \vf_t^\top \mA \vy_{t-1}\\
     &\leq -(1-O(\eta))\eta^2 \sum_{i}\vf_t(i)((\vf_t-e_i)^\top \mA ((\alpha+1)\vy_{t-1}-\alpha \vy_{t-2}))^2 +\\
     &-(1-O(\eta))\eta^2 \sum_{i} \vy_{t}(i)((\vy_t-e_i)^\top \mA^\top((\alpha \vy_{t-2}-(\alpha+1)\vy_{t-1}))^2+\eta^b O(\eta^2)\\
     &=-(1-O(\eta))\eta^2 \sum_{i}\vf_t(i)((\vf_t-e_i)^\top \mA ((\alpha+1)\vy_{t}-\alpha \vy_{t-1}))^2 -(1-O(\eta))\eta^2 \eta^{2b}+\\
     &-(1-O(\eta))\eta^2 \sum_{i} \vy_{t}(i)((\vy_t-e_i)^\top \mA^\top((\alpha \vy_{t-1}-(\alpha+1)\vy_{t}))^2 -(1-O(\eta))\eta^2 \eta^{2b}+\eta^b O(\eta^2)\\
     &=-(1-O(\eta))\eta^2 \sum_{i}\vf_t(i)((\vf_t-e_i)^\top \mA ((\alpha+1)\vy_{t}-\alpha \vy_{t-1}))^2 +\\
     &-(1-O(\eta))\eta^2 \sum_{i} \vy_{t}(i)((\vy_t-e_i)^\top \mA^\top((\alpha \vy_{t-1}-(\alpha+1)\vy_{t}))^2+\eta^b O(\eta^2)
    \end{aligned}
\end{equation}
\end{proof}
From Lemma \ref{Lemma 10}, we can derive our main theorem:
\begin{theorem}\label{thm: theorem KL decrease}
Let $(\vf^*,\vy^*)$ be the unique optimal minimax equilibrium and $\eta$ suffciently small. Assume that $\alpha =\eta^{b-1}$ where $b\in (0,1]$. Then
$RE((\vf^*,\vy^*)||(\vf_t,\vy_t))$
is decreasing with time $t$ by $\eta^{2+b}$ unless $(\vf_t, \vy_t)$ is $O(\eta^{b/3})$-close.
\end{theorem}
\begin{proof}[Proof Sketch]
Using the definition of relative entropy and the following inequality
\[{\vf^*}^\top \mA ((\alpha+1)\vy_t-\alpha \vy_{t-1})\geq {\vf^*}^\top \mA \vy^*,\]
we can derive the following relationship 
\begin{equation*}
    \begin{aligned}
     &RE((\vf^*,\vy^*)||(\vf_{t+1},\vy_{t+1}))-RE((\vf^*,\vy^*)||(\vf_t,\vy_t))\\
     &\leq \log\left(\sum_{i}\vf_t(i)e^{\eta((e_i-\vf_t)^\top\mA ((\alpha+1)\vy_t-\alpha \vy_{t-1}))} \right)+
     \log\left(\sum_{i}\vy_t(i)e^{\eta ((-(\alpha+1)\vf_t+\alpha \vf_{t-1})^\top \mA (e_i-\vy_t))}\right)\\
     &+\eta^b (\vf_{t-1}^{\top}\mA\vy_t-\vf_t^{\top}\mA\vy_{t-1}).
    \end{aligned}
\end{equation*}
Apply Lemma \ref{Lemma 10} along with the property of $\eta^{b/3}$-close gives us the result (the full proof is in Appendix \ref{proof of thm: theorem KL decrease}).
\end{proof}
\begin{remark}
From the above theorem, if the starting point is uniform (i.e., $\vf_1=(1/n,\dots,1/n)$ and $\vy_1=(1/m,\dots,1/m)$), AMWU will reach $O(\eta^{b/3})-close$ in at most:
    $O\left(\frac{\log(nm)}{\eta^{2+b}}\right)$ time steps.
\end{remark}
\subsection{$\eta^{b/3}$-closeness implies closeness to optimum}\label{sec:closeness}
We first need the following lemma:
\begin{lemma}\label{lemma: close lemma 1}
Let $i \in \operatorname{Supp}(\vf^*)$ and $j \in \operatorname{Supp}(\vy^*)$. It holds that $x_T(i)\geq \frac{1}{2}\eta^{b/3}$ and $y_T(i)\geq \frac{1}{2}\eta^{b/3}$ as long as
\[\eta^{b/3} \leq \min_{s \in \operatorname{Supp}(\vf^*)}\frac{1}{(nm)^{1/\vf^*(s)}},\; \min_{s \in \operatorname{Supp}(\vy^*)}\frac{1}{(nm)^{1/\vy^*(s)}}.\]
\end{lemma}
\begin{proof}
By definition of T, the K-L divergence is decreasing for $2 \leq t \leq T-1$, thus
\[RE((\vf^*,\vy^*)||(\vf_{T-1},\vy_{T-1}))< RE((\vf^*,\vy^*)||(\vf_{1},\vy_{1})).\]
This implies that:
\begin{equation*}
    \begin{aligned}
     &\vf^*(i)\log\big(\frac{1}{\vf_{T-1}}(i)\big) \leq \sum_{j} \vf^*(j)\log\big(\frac{1}{\vf_{T-1}}(j)\big)\\
     &\leq \sum_{i} \vf^*(i)\log\big(\frac{1}{\vf_1(i)}\big)+\sum_{i}\vy^*(i)\log\big(\frac{1}{\vy_1(i)}\big)=\log(nm) \\
     &\implies \vf_T(i) > \frac{1}{(mn)^{1/\vf^*(i)}}\geq \eta^{b/3}.
    \end{aligned}
\end{equation*}
Since $|\vf_T(i)-\vf_{T-1}(i)|$ is $O(\eta)$, the result follows.
\end{proof}

Using the above lemma, we can follow the same argument as in Theorem 3.2 of \citet{Daskalakis2018c} to prove the following theorem:
\begin{theorem}[$\eta^{b/3}$-closeness implies closeness to optimum]\label{theorem close implies close to optimum}
Assume $(\vf^*, \vy^*)$ is unique optimal solution of the problem. Let T be the first time KL divergence does not decrease by $\Omega(\eta^{b+2})$. It follows that as $\eta \to 0$, the $\eta^{b/3}$-close point $(\vf_T,\vy_T)$ has distance from $(\vf^*,\vy^*)$ that goes to zero:
\[\lim_{\eta \to 0} \|(\vf^*, \vy^*)-(\vf_T, \vy_T)\|_1=0.\]
\end{theorem}
For the completeness of the paper, we provide the full proof in Appendix \ref{proof of theorem close implies close to optimum}.

\subsection{Proof of local convergence}\label{sec:local_convergence}
We use the following well-known fact in dynamical systems to prove the local convergence:
\begin{proposition}[see \cite{galor2007discrete}]\label{prop: dynamical system convergence 1}
If the Jacobian of the continuously differential update rule $w$ at a fixed point $\vz$ has spectral radius less than one, then there exists a neighborhood $U$ around $\vz$ such that for all $\vx \in U$, the dynamic converges to $\vz$.
\end{proposition}
Given this, our local convergence theorem states: 
\begin{theorem}\label{theorem contraction}
Let$(\vf^*, \vy^*)$ be the unique minimax equilibrium of the game $\mA$. There exists a neighborhood of $(\vf^*, \vy^*)$ such that the AMWU dynamics converge.
\end{theorem}
\begin{proof}[Proof Sketch]
We first construct a dynamical system of AMWU update from Equation (\ref{eq: update rule of AMWU}), in which the variable is two consecutive strategies (e.g., see Equation (\ref{eq: dynamical system update rule})). It is easy to show that $(\vf^*, \vy^*,\vf^*, \vy^*)$ is a fixed point in the dynamical system. Then, following Proposition \ref{prop: dynamical system convergence}, in order to prove the local convergence property, we only need to prove that the Jacobian of the dynamical system computed at $(\vf^*, \vy^*,\vf^*, \vy^*)$ has spectral radius less than one i.e, every eigenvalue of the Jacobian computed at $(\vf^*, \vy^*,\vf^*, \vy^*)$ is less than $1$. The full proof is given in Appendix \ref{proof of theorem contraction}.
\end{proof}

\begin{figure*}[t!]
     \centering
\begin{subfigure}[l]{.49\textwidth}
         \centering
         \includegraphics[width=1.\textwidth]{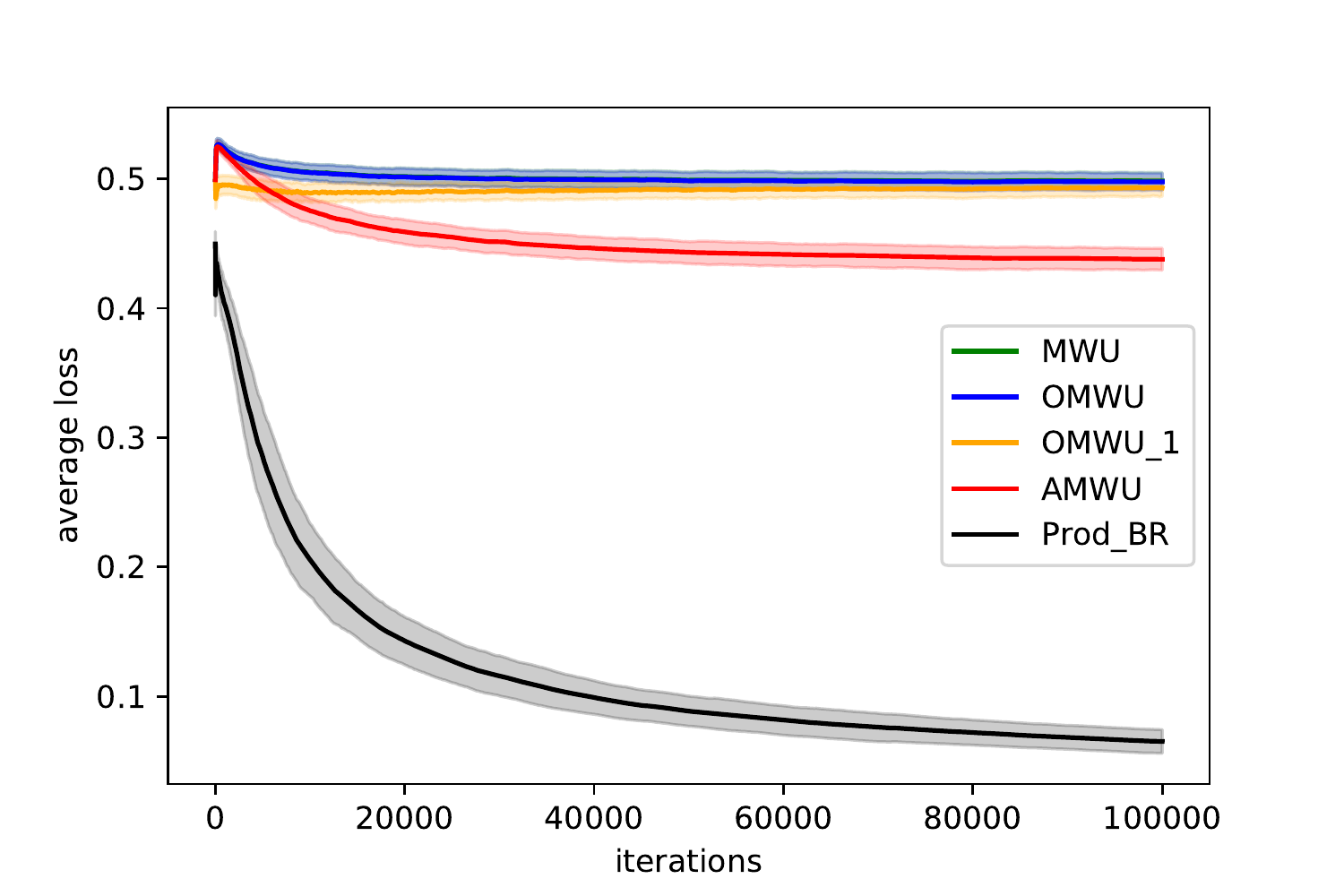}
\caption{0.5 learning rate MWU adversary in random game}
         \label{fig:0.5 MWU adversary in random games}
\end{subfigure}
\begin{subfigure}[l]{.49\textwidth}
         \centering
         \includegraphics[width=1.\textwidth]{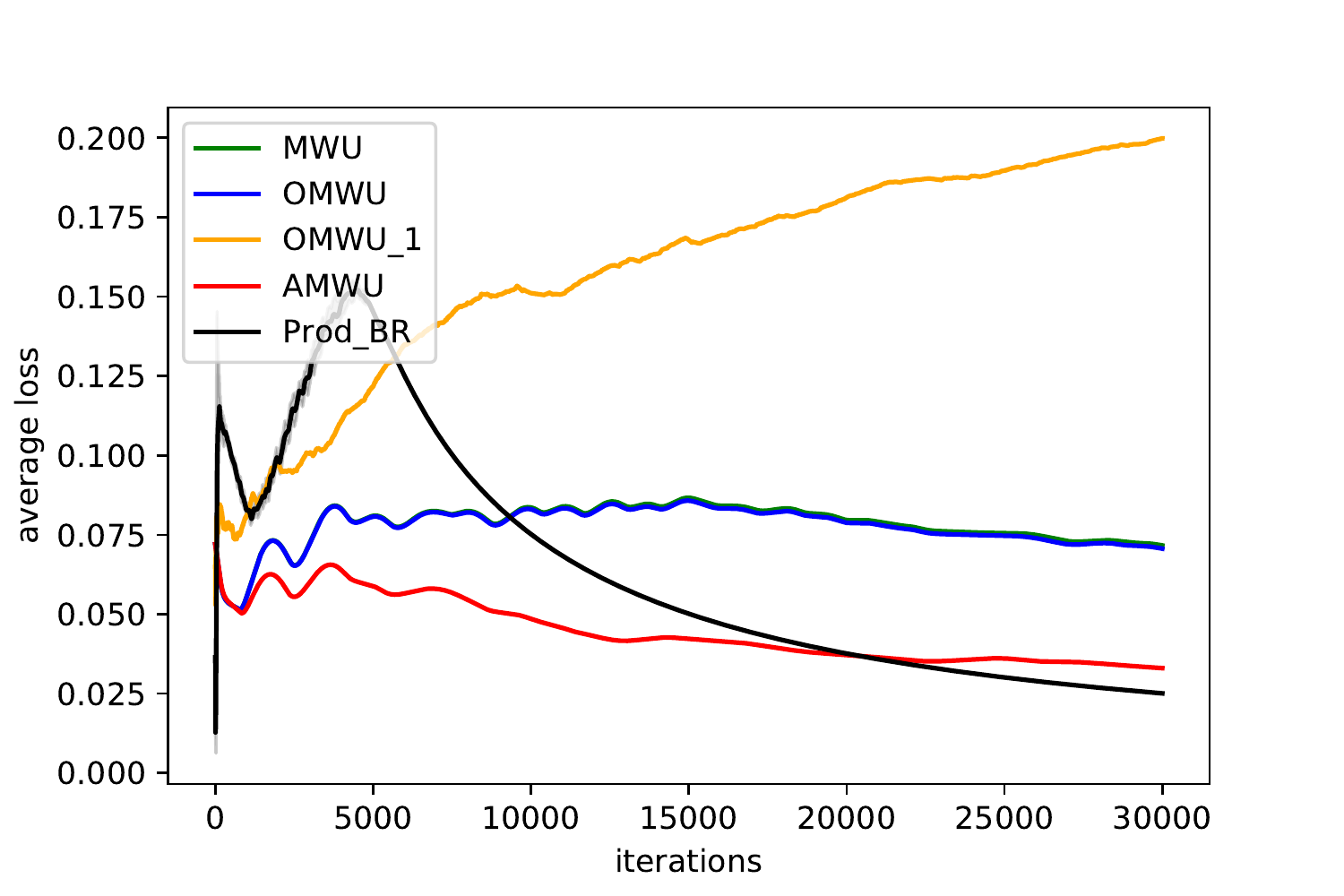}
\caption{0.5 learning rate MWU adversary in Connect Four}
         \label{fig:0.5 MWU adversary in meta games}
\end{subfigure}
\caption{Average Loss Against Oblivious MWU adversary}
  \label{fig:average loss in games}
\end{figure*}
\begin{figure*}[t!]
     \centering
\begin{subfigure}[l]{.49\textwidth}
         \centering
         \includegraphics[width=1.\textwidth]{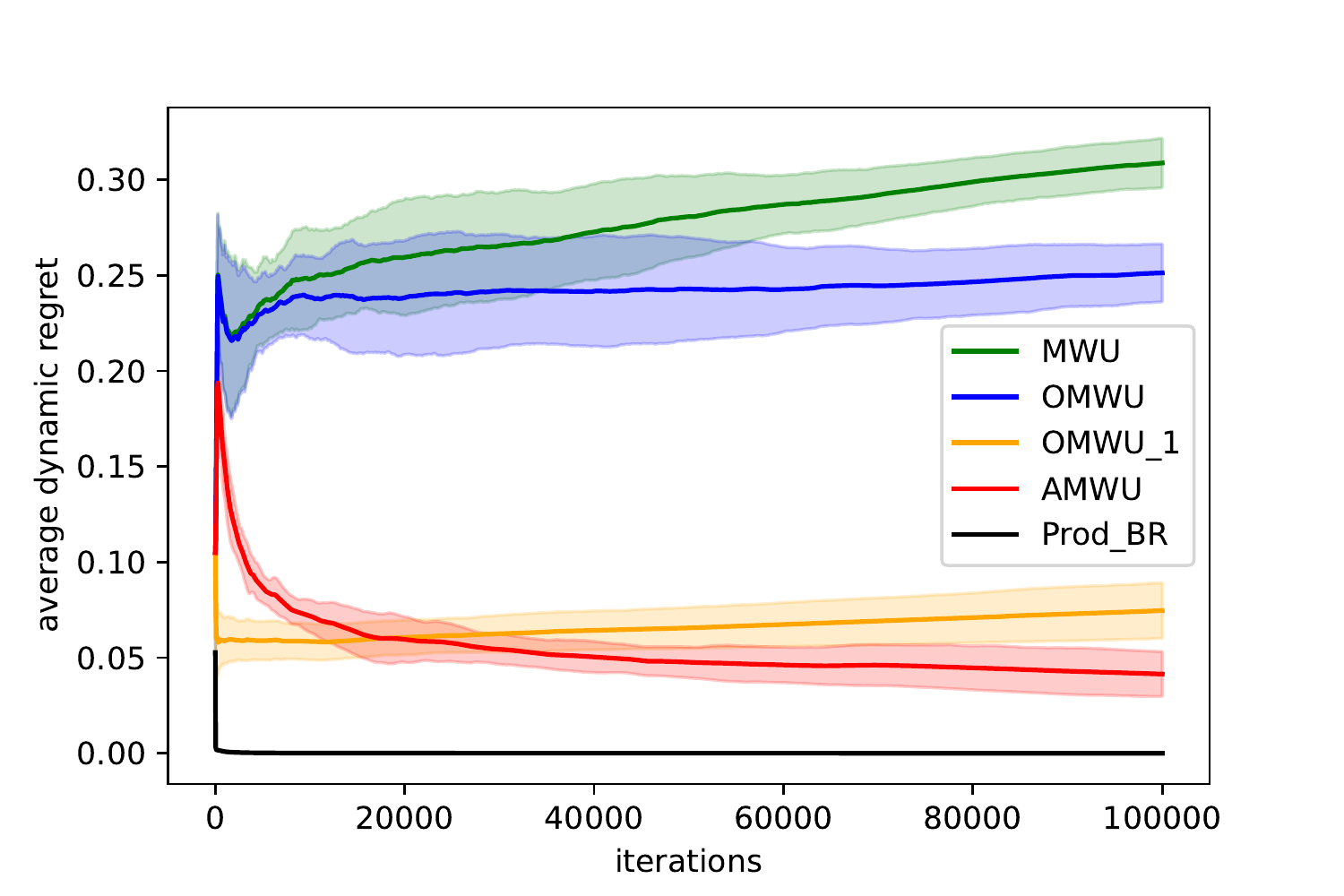}
\caption{non-oblivious MWU adversary in random game}
         \label{fig:0.5 MWU non-oblivious adversary in random games}
\end{subfigure}
\begin{subfigure}[l]{.49\textwidth}
         \centering
         \includegraphics[width=1.\textwidth]{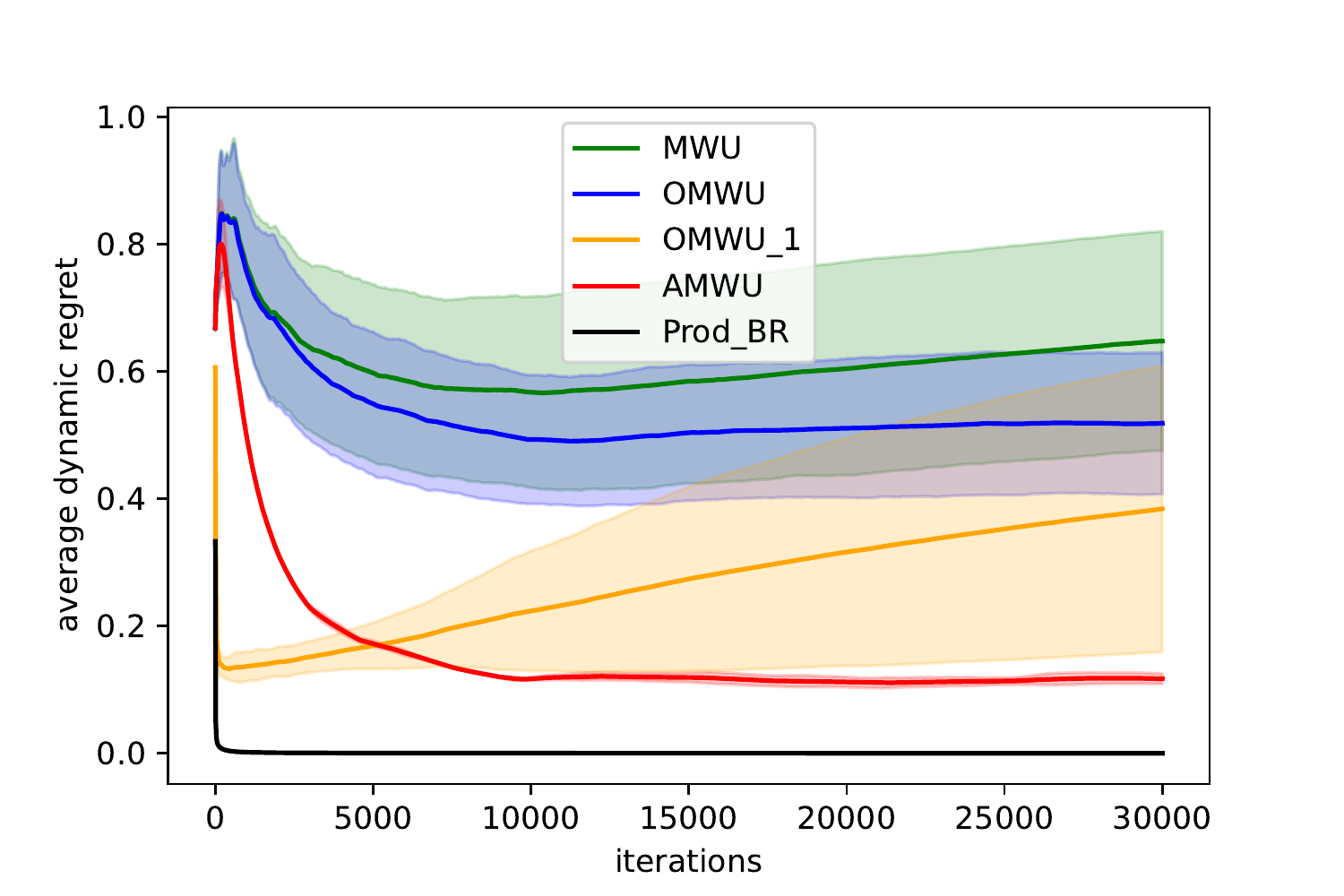}
\caption{non-oblivious MWU adversary in Connect Four}
         \label{fig:0.5 MWU non-oblivous adversary in meta games}
\end{subfigure}
\caption{Average Loss Against Non-Oblivious MWU adversary}
  \label{fig:average loss against non-oblivious in games}
\end{figure*}
\begin{figure*}[t!]
     \centering
\begin{subfigure}[l]{.49\textwidth}
         \centering
         \includegraphics[width=1.\textwidth]{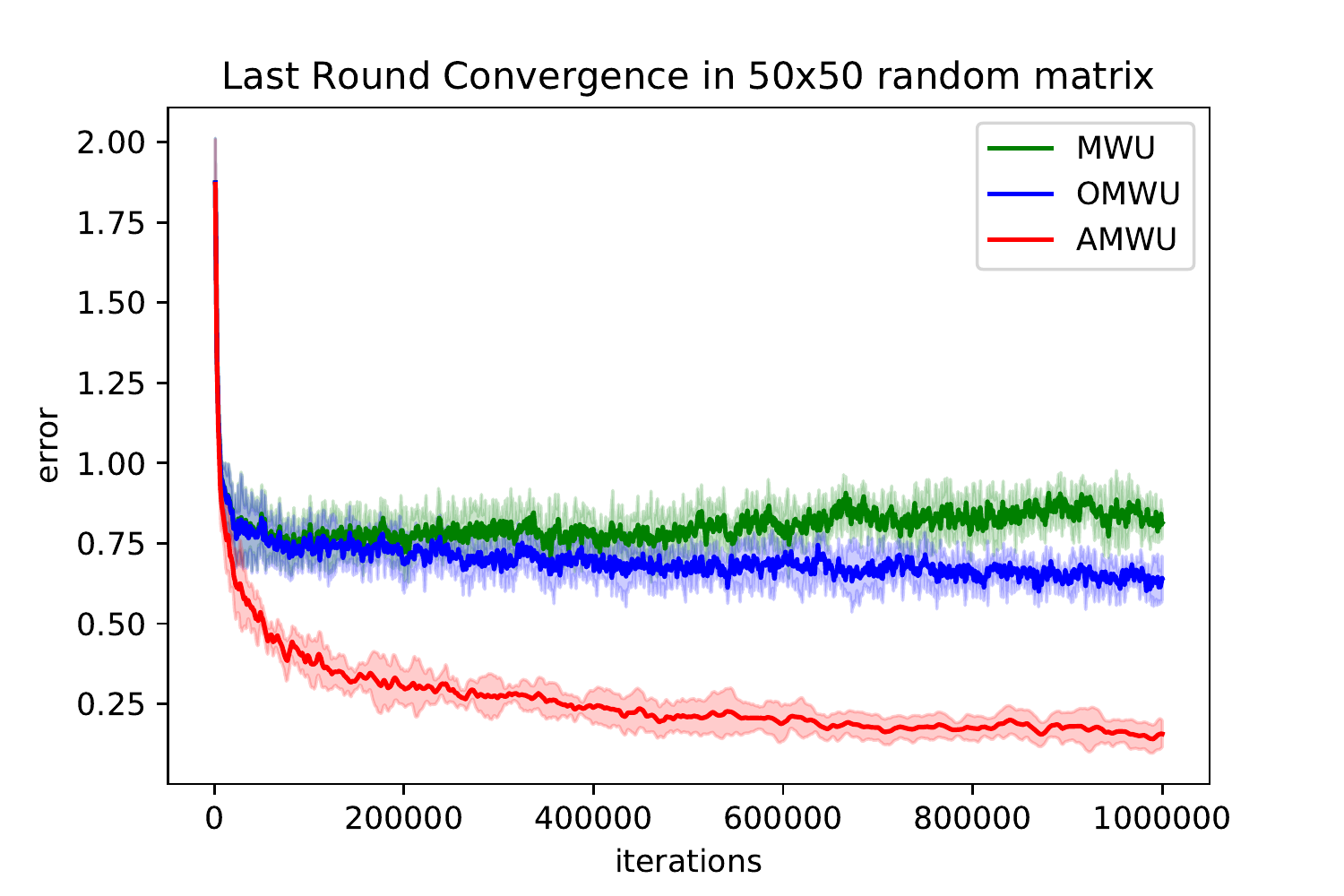}
                
\caption{$50 \times 50$ random games}
       \label{fig:last round convergence in 50x50 random games}
\end{subfigure}
\begin{subfigure}[l]{.49\textwidth}
         \centering
         \includegraphics[width=1.\textwidth]{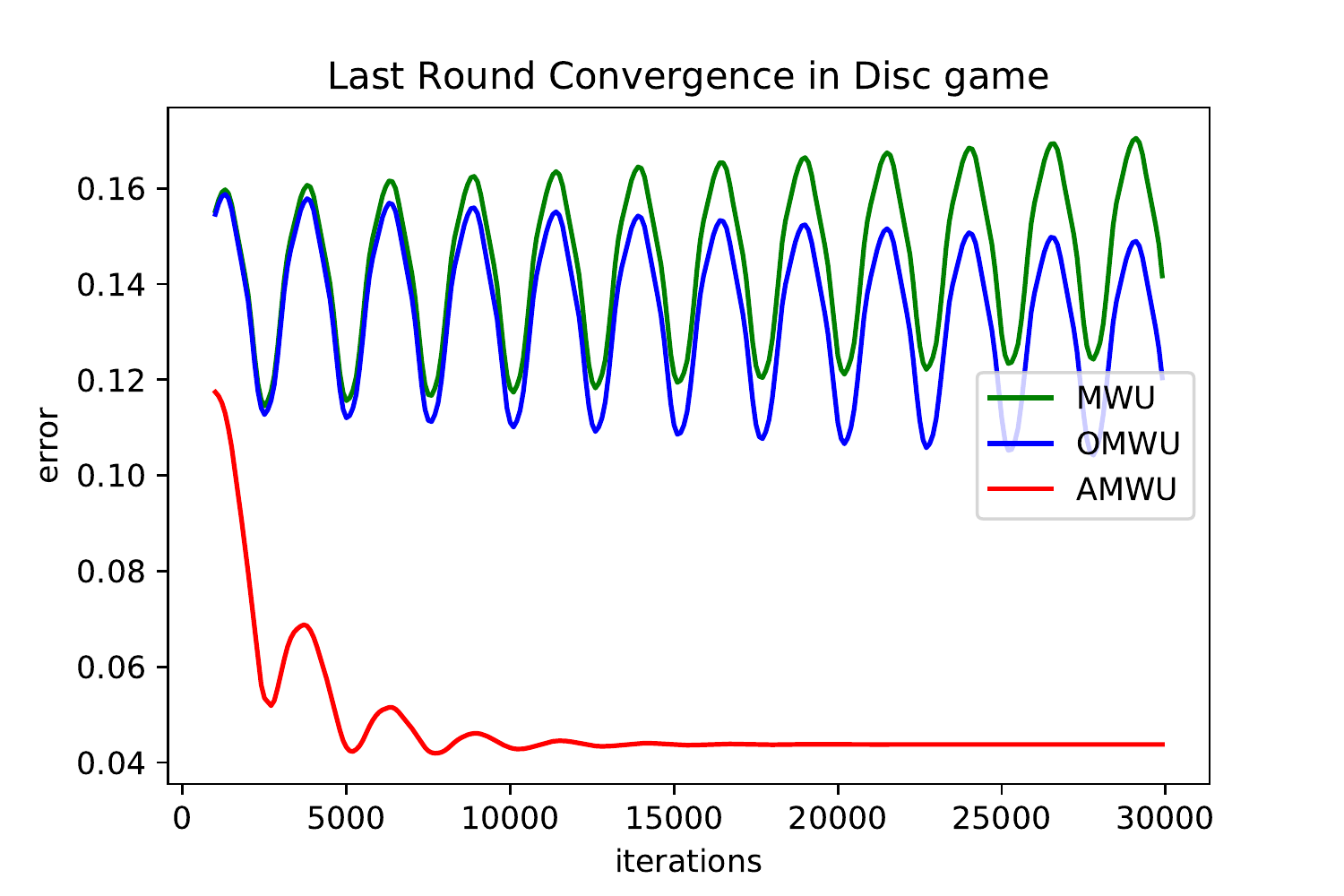}
                
\caption{Disc meta game}
       \label{fig:last round convergence in Disc}
\end{subfigure}
\caption{Last Round Convergence}
  \label{fig:last round convergence in games}
\end{figure*}
\vspace{-10pt}
\section{Experiments}

In this section, we test the performance of our algorithms AMWU and Prod-BR in several setting. Firstly, we consider an oblivious no-external regret adversary and measure the average loss performance of our algorithms against baselines (i.e., MWU, OMWU) in both random and meta games~\cite{czarnecki2020real}. Secondly, we test AMWU and Prod-BR against a non-oblivious no-external regret adversary and measure the average dynamic regret performance. Finally, we test AMWU in self-play setting and measure the last round convergence rate to the NE.

\textbf{Performance against oblivious adversary:} for a fair average loss performance comparison between AMWU, Prod-BR and the baselines, we consider oblivious MWU adversaries: the agent's historical strategies does not effect the strategy of the MWU adversary. In order to create this non-oblivious adversary, we assume the adversary follows MWU to play against a different opponent rather than the agent and therefore the agent's strategies do not effect the adversary's behaviour.~\footnote{The detail setting can be found in Appendix \ref{Appendix: exp oblivious adversary}}. 
As we can see in Figure \ref{fig:average loss in games}, AMWU and Prod-BR outperforms other baselines by a large margin. In particular, Prod-BR achieves a smallest average loss compared to AMWU and other baselines. Intuitively, since the agent plays against an oblivious adversary, a better theoretical regret guarantee of AMWU and Prod-BR can imply a better average loss performance as we have shown in this experiment. Therefore, Prod-BR with the best regret bound measure (i.e., dynamic regret) achieves the best performance, following by AMWU with forward regret guarantee. An interesting observation is that the performance of MWU is almost identical to OMWU with the same learning rate in our setting, reassuring the point in which OMWU does not exploit enough the extra knowledge.

\textbf{Performance against non-oblivious adversary:} we now test our algorithms against non-oblivious adversaries (i.e., the agent's behaviour can change the adversary's strategy) and answer the question: can better theoretical regret bound of AMWU and Prod-BR lead to better regret performance against no-external regret adversary in practice? As we can see in Figure \ref{fig:average loss against non-oblivious in games}, AMWU and Prod-BR achieve much smaller average dynamic regret compared to the baselines. This further assures our theoretical results as both AMWU and Prod-BR have better regret bound guarantee against no-external regret adversary compared to the baselines, leading to better regret bound in practice. 

\textbf{Last round convergence:} we compare the rate of convergence of AMWU against OWMU and MWU. For a fair comparison, we use a common learning rate $\mu=0.01$ for all 3 algorithms~\footnote{The results for other values have similar broad view. See Appendix \ref{Appendix: exp Last round convergence} for more details.}. 
As we can see in Figure \ref{fig:last round convergence in games}, AMWU outperforms OMWU and MWU by a large margin in convergence to the NE. Interestingly, in Connect Four and Disc meta games, AMWU shows clear convergence pattern whereas OMWU and MWU fluctuate under the same setting (Figure~\ref{fig:last round convergence in games}b). 

\textbf{AMWU vs OMWU}: in order to highlight the difference between AMWU and OMWU, we test OWMU$_1$ with the same relative weight between the predictable sequence $\vx_{t-1}$ and the regularizer $R(\vf)$ as AMWU (i.e., $\eta_{OMWU}=\eta_{AMWU} \times \alpha_{AMWU}$). 
As we can clearly see in Figure \ref{fig:average loss in games}, 
AMWU outperform OWMU$_1$ in every game that we consider. We can confirm that AMWU and OMWU are two very different algorithm due to its level of exploiting extra knowledge.
\end{section}
\vspace{-5pt}
\section{Conclusion}
We study online learning problems in which the learner has extra knowledge about the adversary's behaviour (i.e., no-external regret adversary). Under this setting, our algorithms AFTRL and Prod-BR can intensively exploit this extra knowledge to achieve $O(1)$ forward regret and $O(\sqrt{T})$ dynamic regret, respectively. Furthermore, both AFTRL and Prod-BR remain no-regret properties in the worst case scenario of inaccurate extra knowledge. Finally, we show that AMWU, a special case of AFTRL, leads to last round convergence in two-player zero-sum games with a unique NE.
%
%
%



\bibliographystyle{ACM-Reference-Format}
\bibliography{sample-bibliography}

\clearpage
\appendix
\section{Missing Algorithms and Definitions}
\subsection{Follow the Regularized Leader}
Follow the Regularized Leader~\cite{abernethy2009competing}, a well-known no-external regret algorithm, plays an important role in the analysis of our main algorithm:
\begin{algorithm}[h]
\textbf{Input:} \text{learning rate $\eta >0$,
$\vf_1=\argmin_{\vf \in \mathcal{F}}R(\vf)$.}

\textbf{Output:} \text{next strategy update}
\[\vf_{t+1}=\argmin_{\vf \in \mathcal{F}} F_{t+1}(\vf)= \langle \vf, \sum_{s=1}^t \vx_s \rangle + \frac{R(\vf)}{\eta}  \]
\caption{Follow the Regularized Leader}
\label{alg:FTRL}
\end{algorithm}

\subsection{Accurate Mirror Descent}\label{appendix: AMD}

We now apply our accurate prediction technique into another well-known no-regret algorithm, Mirror Descent~\cite{nemirovskij1983problem}. Let $\mathcal{R}$ be a $\beta$-strongly convex function with respect to a norm $\|.\|_p$, and let $D_{\mathcal{R}(.,.)}$ denote the Bregman divergence with respect to $\mathcal{R}$. Let $\|.\|_{q}$ be dual to $\|.\|_p$. Then the Accurate Mirror Descent (AMD) algorithm can be described as follows:

\begin{algorithm}[h]
\textbf{Input:} \text{ learning rate $\eta > 0$, exploiting rate $\alpha \geq 1$,}

\text{$\vf_1=\vg_1=\argmin_{\vf \in \mathcal{F}}\mathcal{R}(\vf)$.}

\textbf{Output:} \text{next strategy update}
\begin{equation*}
    \begin{aligned}
    &\vg_{t+1}=\argmin_{\vg \in \mathcal{F}} G_{t+1}(\vg)= \eta \langle \vg, \vx_t \rangle + D_{\mathcal{R}}(\vg,\vg_t) \\
    &\vf_{t+1}=\argmin_{\vf \in \mathcal{F}} F_{t+1}(\vf)= \eta \langle \vf, \alpha M_{t+1} \rangle + D_{\mathcal{R}}(\vf,\vg_{t+1}) \\
    \end{aligned}
\end{equation*}
\caption{Accurate Mirror Descent}
\end{algorithm}
where the regularizer $R(\vf)$ is a
$\beta$-strongly convex function with respect of $l_p$ norm, $p\geq 1$.

\subsection{Missing Definitions}\label{appendix: missing definitions}
\begin{definition}[\textbf{$\beta$-closeness}~\cite{mehta2017mutation}]
Assume $\beta >0$. A point $(\vf,\vy) \in \Delta_n \times \Delta_m$ is $\beta$-close if for each $i \in [n]$, it holds $\vf_i \leq \beta$ or $|\vf^{\top}\mA \vy -(\mA \vy)_i|\leq \beta$ and for each $j \in [m]$, it holds $\vy_i \leq \beta$ or $|\vf^\top\mA \vy -(\mA^\top \vx)_i|\leq \beta$.
\end{definition}
Finally, in order to analyze the dynamic in matrix game $\mA$, we will use the Kullback-Leibler divergence to understand the behaviour of the players's strategies.
\begin{definition}[\cite{KL1951}]\label{RE definition}
The relative entropy or Kull-Leibler (K-L) divergence between two vectors 
${\vx}_1$ and ${\vx}_2$ in $\Delta_n$ is defined as 
$RE({\vx}_1\|{\vx}_2) = \sum_{i=1}^n {\vx}_1(i)\log \left(\frac{{\vx}_1(i)}{{\vx}_2(i)}\right).$
\end{definition}
This is always non-negative. We can also show that $RE({\vx}_1\|{\vx}_2)= 0$ if and only if ${\vx}_1={\vx}_2$ almost everywhere~\citep{Gibbs1970}.
\section{Missing Proofs}
\begin{lemma}[Lemma \ref{lemma: forward regret vs external regret}]\label{proof: lemma forward regret vs external regret}
Let $\vg_t$ be defined as above, then the following relationship holds for any $\vf \in \mathcal{F}$:
\[\sum_{t=1}^T \langle \vg_t, \vx_t \rangle \leq \langle \vf, \sum_{t=1}^T \vx_t \rangle + \frac{R(\vf)}{\eta}.\]
\end{lemma}
\begin{proof}[Proof of Lemma \ref{lemma: forward regret vs external regret}]
We prove this by induction. For $t =1$:
\[\langle \vg_1, \vx_1 \rangle \leq \langle \vg_1, \vx_1 \rangle + \frac{R(\vg_1)}{\eta} \leq \langle \vf, \vx_1 \rangle + \frac{R(\vf)}{\eta} \; \forall \vf \in \mathcal{F}.\]
Suppose the statement is true for $T$ such that 
\[\sum_{t=1}^T \langle \vg_t, \vx_t \rangle \leq \langle \vf, \sum_{t=1}^T \vx_t \rangle + \frac{R(\vf)}{\eta}\; \forall \vf \in \mathcal{F}.\]
Adding $\langle \vg_{T+1}, \vx_{T+1} \rangle$ on both sides we have
\begin{equation*}
    \begin{aligned}
    \sum_{t=1}^{T+1} \langle \vg_t, \vx_t \rangle &\leq \langle \vf, \sum_{t=1}^T \vx_t \rangle + \frac{R(\vf)}{\eta}+\langle \vg_{T+1}, \vx_{T+1} \rangle \; \forall \vf \in \mathcal{F} \\
    &\leq \langle \vg_{T+1}, \sum_{t=1}^T \vx_t \rangle + \frac{R(\vg_{T+1})}{\eta}+\langle \vg_{T+1}, \vx_{T+1} \rangle \\
    &\leq \langle \vf, \sum_{t=1}^{T+1} \vx_t \rangle + \frac{R(\vf)}{\eta} \;\forall \vf \in \mathcal{F}. 
    \end{aligned}
\end{equation*}
Thus the statement is true for $T+1$.

From the above Inequality, if an algorithm is a no-forward regret, i.e.:
\[\sum_{t=1}^\top   \left( \langle \vf_t,\vx_t \rangle - \langle \vg_t, \vx_t \rangle\right)= o(T),\]
then we also have:
\begin{equation*}
    \begin{aligned}
    &\min_{\vf \in \mathcal{F}}\sum_{t=1}^\top   \left( \langle \vf_t,\vx_t \rangle - \langle \vf, \vx_t \rangle\right) \leq \sum_{t=1}^\top   \left( \langle \vf_t,\vx_t \rangle - \langle \vg_t, \vx_t \rangle\right)+\frac{R(\vf)}{\eta}\\
    &=o(T)+\frac{R(\vf)}{\eta}=o(T).
    \end{aligned}
\end{equation*}
Thus, the algorithm is also a no-external regret algorithm.
\end{proof}
\begin{lemma}[Doubling Trick]\label{Doubling Trick}
The idea of the doubling trick is to divide the time interval into different phases and restart the algorithm (i.e., AFTRL) in each phase. We will prove that by considering different phases in the process, the AFTRL will still maintain the regret bound of $O\left(\sqrt{\sum_{t=1}^{T} (\|\vx_t-\vx_{t-1}\|_q)^2}\right)$.


Using Lemma \ref{lemma: forward regret vs external regret}, the regret bound in Equation \ref{ineq: regret bound of AFTRL} can be derived as:
\begin{equation*}
    \begin{aligned}
    \sum_{t=1}^T \langle \vf_t,\vx_t \rangle-\langle \vf^*, \sum_{t=1}^T \vx_t \rangle
    \leq \frac{\alpha}{\eta \alpha}R(\vf^*) +\frac{\eta \alpha}{\beta} \sum_{t=1}^T (\|\vx_t-\vx_{t-1}\|_q)^2\;\forall \vf^* \in \mathcal{F}. 
    \end{aligned}
\end{equation*}
Now, we break the time interval $T$ into different phases, in which phase $i$ has a constant learning rate $\eta_i=\eta_0 2^{-i}$. Define the starting point of phase $i+1$ such as
\begin{equation*}
    \begin{aligned}
     s_{i+1}=\min \{\tau: \frac{\eta_i\alpha}{\beta} \sum_{t=s_i}^{\tau} (\|\vx_t-\vx_{t-1}\|_t^*)^2 >\frac{\alpha}{\eta_i \alpha} R(\vf^*)\}.
    \end{aligned}
\end{equation*}
and $s_1=1$.
Let $N$ be the last phase of the game and let $s_{N+1}=T+1$. We then have:
\begin{equation*}
    \begin{aligned}
    \sum_{t=1}^T \langle \vf_t,\vx_t \rangle-\langle \vf^*, \sum_{t=1}^T \vx_t \rangle \leq \sum_{i=1}^N \frac{\alpha}{\eta_i \alpha} R(\vf^*)+\frac{\eta_i \alpha}{\beta} \sum_{t=s_i}^{s_{i+1}-1} (\|\vx_t-\vx_{t-1}\|_t^*)^2\\
    \leq 2 \sum_{i=1}^N \frac{\alpha}{\eta_i \alpha} R(\vf^*)  \leq \frac{2^{N+2}}{\eta_0} R(\vf^*),
    \end{aligned}
\end{equation*}
where the inequalities come from the definition of $s_i$. Note that we have:
\begin{equation*}
    \begin{aligned}
     \frac{1}{\eta_0}=\frac{1}{\eta_{N-1}2^{N-1}} \leq\frac{1}{2^{N-1}}\sqrt{\sum_{t=s_{N-1}}^{s_{N}} (\|\vx_t-\vx_{t-1}\|_q)^2} \sqrt{\frac{\alpha}{\beta R(\vf^*)}}\\
     \leq \frac{1}{2^{N-1}}\sqrt{\sum_{t=1}^{T} (\|\vx_t-\vx_{t-1}\|_q)^2} \sqrt{\frac{\alpha}{\beta R(\vf^*)}}.
    \end{aligned}
\end{equation*}
Thus we have:
\begin{equation*}
    \begin{aligned}
     \sum_{t=1}^T \langle \vf_t,\vx_t \rangle-\langle \vf^*, \sum_{t=1}^T \vx_t \rangle \leq \frac{2^{N+2}}{\eta_0}R(\vf^*)\\
     \leq 2^{N+2}\frac{1}{2^{N-1}}\sqrt{\sum_{t=1}^{T} (\|\vx_t-\vx_{t-1}\|_q)^2} \sqrt{\frac{\alpha}{\beta R(\vf^*)}} R(\vf^*)\\
     =8 \sqrt{\sum_{t=1}^{T} (\|\vx_t-\vx_{t-1}\|_q)^2} \sqrt{\frac{\alpha R(\vf^*)}{\beta}} 
     = O\left(\sqrt{\sum_{t=1}^{T} (\|\vx_t-\vx_{t-1}\|_q}\right).
    \end{aligned}
\end{equation*}
Thus, we derive the result.
\end{lemma}
\begin{lemma}[Lemma \ref{lemma: consecutive strategis of no-regret algorithm}]\label{Proof of lemma: consecutive strategis of no-regret algorithm}
Let $\vf_t$, $\vf_{t+1}$ be two consecutive strategies of no-regret algorithms (i.e., FTRL, OMD). Then we have for any norm $\|.\|_q$:
\[\|\vf_{t+1}-\vf_t\|_q =O(\frac{1}{\sqrt{T}}).\]
\end{lemma}
In order to prove Lemma \ref{lemma: consecutive strategis of no-regret algorithm}, we first need to have the following lemmas about the distance between two consecutive strategies of FTRL and OMD:
\begin{lemma}\label{lemma: consecutive strategies of FTRL}
Let $\vf_t$, $\vf_{t+1}$ be two consecutive strategies of FTRL algorithm. Then we have:
\[\|\vf_{t+1}-\vf_t\|_p\leq \eta\frac{2 n}{\beta}, \;\text{where $\|.\|_p$ denotes $l_p$ norm.}\]
\end{lemma}
\begin{proof}
Following the property of $\beta$-strongly convex function we have:
\begin{equation*}
    \begin{aligned}
     &F_t(\vf_{t+1})-F_t(\vf_t) \geq\frac{\beta}{2\eta} \|\vf_{t+1}-\vf_t\|_p^2\\
     &\iff \langle \vf_{t+1}, \sum_{s=1}^t \vx_s \rangle+\frac{R(\vf_{t+1})}{\eta}- \langle \vf_{t+1},\vx_t \rangle- \langle \vf_{t}, \sum_{s=1}^{t-1} \vx_s \rangle-\frac{R(\vf_{t})}{\eta}\geq \frac{\beta}{2\eta} \|\vf_{t+1}-\vf_t\|_p^2\\
     &\iff F_{t+1}(\vf_{t+1})-\langle \vf_{t+1},\vx_t \rangle- \langle \vf_{t}, \sum_{s=1}^{t-1} \vx_s \rangle-\frac{R(\vf_{t})}{\eta}\geq \frac{\beta}{2\eta} \|\vf_{t+1}-\vf_t\|_p^2.
    \end{aligned}
\end{equation*}
By definition, we have $F_{t+1}(\vf_{t+1}) \leq F_{t+1}(\vf_{t})$. Thus, substitute it in the above inequality we have:
\begin{equation*}
    \begin{aligned}
    &F_{t+1}(\vf_{t})-\langle \vf_{t+1},\vx_t \rangle- \langle \vf_{t}, \sum_{s=1}^{t-1} \vx_s \rangle-\frac{R(\vf_{t})}{\eta}\geq \frac{\beta}{2\eta} \|\vf_{t+1}-\vf_t\|_p^2\\
    &\iff \langle \vf_{t}, \sum_{s=1}^{t} \vx_s \rangle+\frac{R(\vf_{t})}{\eta}-\langle \vf_{t+1},\vx_t \rangle- \langle \vf_{t}, \sum_{s=1}^{t-1} \vx_s \rangle-\frac{R(\vf_{t})}{\eta}\geq \frac{\beta}{2\eta} \|\vf_{t+1}-\vf_t\|_p^2\\
    &\iff \langle \vf_t-\vf_{t+1}, \vx_t \rangle \geq \frac{\beta}{2\eta} \|\vf_{t+1}-\vf_t\|_p^2\\
    &\implies \|\vf_{t+1}-\vf_t\|_p \|\vx_t\|_q \geq \frac{\beta}{2\eta} \|\vf_{t+1}-\vf_t\|_p^2 \\
    &\implies \frac{2\eta n}{\beta} \geq \|\vf_{t+1}-\vf_t\|_p,
    \end{aligned}
\end{equation*}
since $\vx_t \in [0,1]^n$ then $\|\vx_t\|_q \leq n^{1/q}=n^{1-1/p} \leq n$. Thus, we derive the result. 

\end{proof}

A similar property can be found in other no-regret algorithm, such as Online Mirror Descent:
\begin{lemma}\label{lemma: consecutive strategies of OMD}
Let $\vg_{t}$, $\vg_{t+1}$ be two consecutive strategies of OMD algorithm. Then we have:
\[\|\vf_{t+1}-\vf_t\|_p \leq \frac{\eta}{\beta}\]
\end{lemma}
\begin{proof}
Following the property of $\beta$-strongly convex function we have:
\begin{equation*}
    \begin{aligned}
     &G_{t+1}(\vg_{t})-G_{t+1}(\vg_{t+1})\geq \frac{\beta}{2} \|\vg_{t+1}-\vg_t\|_p^2\\
     &\iff \eta \langle \vg_t-\vg_{t+1},\vx_t \rangle+D_{\mathcal{R}}(\vg_t,\vg_t)-D_{\mathcal{R}}(\vg_{t+1},\vg_t) \geq \frac{\beta}{2} \|\vg_{t+1}-\vg_t\|_p^2\\
     &\iff \eta \langle \vg_t-\vg_{t+1},\vx_t \rangle \geq D_{\mathcal{R}}(\vg_{t+1},\vg_t)+ \frac{\beta}{2} \|\vg_{t+1}-\vg_t\|_p^2 \\
     &\implies\eta \langle \vg_t-\vg_{t+1},\vx_t \rangle \geq \frac{\beta}{2}\|\vg_{t+1}-\vg_t\|_p^2+\frac{\beta}{2} \|\vg_{t+1}-\vg_t\|_p^2\\
     &\implies \eta \|\vg_t-\vg_{t+1}\|_p \|\vx_t\|_q \geq \beta \|\vg_{t+1}-\vg_t\|_p^2 \\
     &\implies \frac{\eta}{\beta} n \geq \|\vg_{t+1}-\vg_t\|_p, 
    \end{aligned}
\end{equation*}
since $D_{\mathcal{R}}(\vg_t,\vg_t)=0$ and $\vx_t \in [0,1]^n$.
\end{proof}

Now we can prove Lemma \ref{lemma: consecutive strategis of no-regret algorithm}:

\begin{proof}[Proof of Lemma \ref{lemma: consecutive strategis of no-regret algorithm}]
From Lemma \ref{lemma: consecutive strategies of FTRL} and Lemma \ref{lemma: consecutive strategies of OMD} along with the property of no-regret algorithm such as $\eta = O(\frac{1}{\sqrt{T}})$, we have:
\[\|\vf_{t+1}-\vf_t\|_p=O(\frac{1}{\sqrt{T}}).\]
Now for $q>p$, it is easy to show that:
\begin{equation*}
    \begin{aligned}
    \|\vf_{t+1}-\vf_t\|_q \leq \|\vf_{t+1}-\vf_t\|_p \\
    \implies \|\vf_{t+1}-\vf_t\|_q=O(\frac{1}{\sqrt{T}}).
    \end{aligned}
\end{equation*}
For $q<p$, using the Holder's Inequality, we then have:
\begin{equation*}
    \begin{aligned}
    &\|\vf_{t+1}-\vf_t\|_q \leq n^{1/q-1/p} \|\vf_{t+1}-\vf_t\|_p = n^{1/q-1/p} O(\frac{1}{\sqrt{T}}) \\
    &\implies \|\vf_{t+1}-\vf_t\|_q=O(\frac{1}{\sqrt{T}}). 
    \end{aligned}
\end{equation*}
We complete the proof.
\end{proof}
\begin{theorem}[Theorem \ref{thm: AFTRL for convec regularizer}]\label{Appendix: thm AFRL for convec regularizer}
Let $ \mathcal{F} \subset \mathbb{R}^n$ be a convex compact set and let R be a $\beta$-strongly convex function with $\min_{\vf \in \mathcal{F}} R(\vf)=0.$ For any strategy of the environment, the AFTRL algorithm satisfies:
\begin{equation*}
    \begin{aligned}
    \sum_{t=1}^T \langle \vf_t,\vx_t \rangle-\frac{1}{\alpha}\langle \vf^*, \sum_{t=1}^T \vx_t \rangle -\frac{\alpha-1}{\alpha}\sum_{t=1}^T\langle \vg_{t}, \vx_t \rangle
    \leq \frac{1}{\eta \alpha}R(\vf^*) +\frac{\eta \alpha}{\beta} \sum_{t=1}^T (||\vx_t-\vx_{t-1}||_{q})^2. 
    \end{aligned}
\end{equation*}
\end{theorem}
\begin{proof}[Proof of Theorem \ref{thm: AFTRL for convec regularizer}]
Let us first define $\vh_{t+1}$ as follow
\[\vh_{t+1}= \argmin_{\vf \in \mathcal{F}} H_{t+1}(\vf)=\langle \vf, \sum_{s=1}^t \vx_s +\alpha \vx_{t+1}\rangle +\frac{R(\vf)}{\eta}.\]
Observe that for any sequence of $\vf_t \in \mathcal{F}$,
\begin{equation*}
\begin{aligned}
 &\sum_{t=1}^T \langle \vf_t,\vx_t \rangle= \sum_{t=1}^T\langle \vf_t-\vh_t,\vx_t-\vx_{t-1}\rangle   \\
 &+ \sum_{t=1}^T\langle \vf_t-\vh_t, \vx_{t-1} \rangle +\sum_{t=1}^T\langle \vh_t, \vx_t \rangle.
\end{aligned}
\end{equation*}
We now prove by induction that
\begin{equation}\label{inequation 1.1}
\begin{aligned}
&\sum_{t=1}^T\langle \vf_t-\vh_t, \vx_{t-1} \rangle +\sum_{t=1}^T\langle \vh_t, \vx_t \rangle \leq \\
&\frac{1}{\alpha}\langle \vf', \sum_{t=1}^T \vx_t \rangle +\frac{\alpha-1}{\alpha}\sum_{t=1}^T\langle \vg_{t}, \vx_t \rangle + \frac{1}{\eta \alpha}R(\vf'), \; \forall \vf' \in \mathcal{F}.
\end{aligned}
\end{equation}
For $t=1, M_1=0$, we have
\begin{equation*}
    \begin{aligned}
    \langle \vh_1, \vx_1 \rangle +\frac{R(\vh_1)}{\eta \alpha} \leq \langle \vf', \vx_1 \rangle +\frac{R(\vf')}{\eta \alpha},
    \implies \frac{1}{\alpha} \langle \vh_1, \vx_1 \rangle \leq \frac{1}{\alpha} \langle \vf', \vx_1 \rangle+ \frac{R(\vf')}{\eta \alpha},
    \end{aligned}
\end{equation*}
since $\alpha \geq 1$ and $R(\vf)\geq 0\;\forall \vf \in \mathcal{F}$.
We also have
\begin{equation*}
\begin{aligned}
&\langle \vh_1, \alpha \vx_t \rangle +\frac{R(\vh_1)}{\eta} \leq \langle \vg_1, \alpha \vx_t \rangle +\frac{R(\vg_1)}{\eta} \\
&= \langle \vg_1, \vx_t \rangle +\frac{R(\vg_1)}{\eta} + (\alpha-1)\langle \vg_1, \vx_t \rangle\\
&\leq \langle \vh_1, \vx_t \rangle +\frac{R(\vh_1)}{\eta} + (\alpha-1)\langle \vg_1, \vx_t \rangle\\
&\implies \langle \vh_1, \vx_1 \rangle \leq \langle \vg_1, \vx_1 \rangle.
\end{aligned}
\end{equation*}
Then, we have:
\begin{equation*}
    \begin{aligned}
     \langle \vh_1, \vx_1 \rangle \leq \frac{1}{\alpha} \langle \vf', \vx_1 \rangle+ \frac{R(\vf')}{\eta \alpha}+ \frac{\alpha-1}{\alpha} \langle \vg_1, \vx_t \rangle.
    \end{aligned}
\end{equation*}
Thus, the first step in the induction for $t=1$ is correct.

For the purpose of induction, suppose that the above inequality holds for $\tau = T-1$. Using $\vf' = \vf_T$ and add $\langle \vf_T-\vh_T, \vx_{t-1} \rangle + \langle \vh_T, \vx_{t-1} \rangle$ on both sides we have
\begin{equation}
    \begin{aligned}
     & \sum_{t=1}^T\langle \vf_t-\vh_t, \vx_{t-1} \rangle +\sum_{t=1}^T\langle \vh_t, \vx_t \rangle \\ &\leq \frac{1}{\alpha} \langle \vf_T , \sum_{t=1}^{T-1}\vx_t\rangle +\frac{\alpha-1}{\alpha} \sum_{t=1}^{T-1}\langle \vg_t,\vx_t\rangle +\frac{1}{\eta \alpha} R(\vf_T) +\langle \vf_T-h_T, \vx_{t-1} \rangle +\langle h_T, \vx_t \rangle \\
     &=\frac{1}{\alpha} ( \langle \vf_T , \sum_{t=1}^{T-1}\vx_t +\alpha \vx_{t-1}\rangle + \frac{1}{\eta} R(\vf_T))
     +\frac{\alpha-1}{\alpha} \sum_{t=1}^{T-1}\langle \vg_t,\vx_t\rangle +\langle h_T, \vx_t-\vx_{t-1} \rangle \\
     &\leq \frac{1}{\alpha} ( \langle h_T , \sum_{t=1}^{T-1}\vx_t +\alpha \vx_{t-1}\rangle + \frac{1}{\eta} R(h_T))
     +\frac{\alpha-1}{\alpha} \sum_{t=1}^{T-1}\langle \vg_t,\vx_t\rangle +\langle h_T, \vx_t-\vx_{t-1} \rangle \\
     &= \frac{1}{\alpha} ( \langle h_T , \sum_{t=1}^{T-1}\vx_t +\alpha \vx_t\rangle + \frac{1}{\eta} R(h_T))+\frac{\alpha-1}{\alpha} \sum_{t=1}^{T-1}\langle \vg_t,\vx_t\rangle\\
     &\leq \frac{1}{\alpha} ( \langle g_T , \sum_{t=1}^{T-1}\vx_t +\alpha \vx_t\rangle + \frac{1}{\eta} R(g_T))+\frac{\alpha-1}{\alpha} \sum_{t=1}^{T-1}\langle \vg_t,\vx_t\rangle\\
     &\leq \frac{1}{\alpha} ( \langle \vf' , \sum_{t=1}^{T-1}\vx_t +\vx_t\rangle + \frac{1}{\eta} R(\vf'))+\frac{\alpha-1}{\alpha} \sum_{t=1}^{T}\langle \vg_t,\vx_t\rangle \; \forall \vf'.
    \end{aligned}
\end{equation}
The proof is derived from the optimality of $\vf_t, \vg_t$ and $\vh_t$. This concludes the inductive argument.

Now, we only need to bound the sum:
\[\sum_{t=1}^T\langle \vf_t-\vh_t,\vx_t-\vx_{t-1}\rangle.\]
Using the property of strongly convex function we have:
\begin{equation*}
    \begin{aligned}
    F_t(\vh_t)-F_t(\vf_t) \geq \frac{\beta}{2\eta} ||\vh_t-\vf_t||_p^2 \\
    H_t(\vf_t)-H_t(\vh_t) \geq \frac{\beta}{2\eta} ||\vh_t-\vf_t||_p^2 \\
    \implies \alpha \langle \vh_t-\vf_t, \vx_{t-1}-\vx_t \rangle \geq \frac{\beta}{\eta} ||\vh_t-\vf_t||_p^2 \\
    \implies ||\vh_t-\vf_t||_p ||\vx_{t-1}-\vx_t||_{q} \geq \frac{\beta}{\eta \alpha}||\vh_t-\vf_t||_p^2 \\
    \implies ||\vx_{t-1}-\vx_t||_{q} \geq \frac{\beta}{\eta \alpha}||\vh_t-\vf_t||_p.
    \end{aligned}
\end{equation*}
Thus, we have
\begin{equation*}
    \begin{aligned}
     \sum_{t=1}^T\langle \vf_t-\vh_t,\vx_t-\vx_{t-1}\rangle &\leq \sum_{t=1}^T ||\vf_t-\vh_t||p ||\vx_t-\vx_{t-1}||_{q}\\
     &\leq \frac{\eta \alpha}{ \beta}\sum_{t=1}^T (||\vx_{t-1}-\vx_t||_{q})^2.
    \end{aligned}
\end{equation*}
Along with the Inequality \ref{inequation 1.1} gives the complete proof.
\end{proof}

\begin{lemma}[Lemma \ref{lm: BR performance each round}]\label{proof of lm: BR performance each round}
Let $\vx_t$, $\vx_{t+1}$ be two consecutive strategies of a no-regret algorithm (i.e., FTRL, OMD). Then, we have:
 \[\langle \vb,\vx_{t+1} \rangle- \langle \vc, \vx_{t+1}\rangle = O(\frac{1}{\sqrt{T}}),\; \text{where}\; \vb=\argmin_{\vf \in \mathcal{F}} \langle \vf, \vx_{t}\rangle \;,\;\vc=\argmin_{\vf \in \mathcal{F}} \langle \vf, \vx_{t+1}\rangle.\]
\end{lemma}
\begin{proof}[Proof of Lemma \ref{lm: BR performance each round}]
Since $\vb=\argmin_{\vf \in \mathcal{F}} \langle \vf, \vx_{t}\rangle$, we then have: $\langle \vb,\vx_{t} \rangle \leq \langle \vc,\vx_{t} \rangle$. Thus, we can derive that:
\begin{equation*}
    \begin{aligned}
     &\langle \vb,\vx_{t+1} \rangle- \langle \vc, \vx_{t+1}\rangle=\langle \vb,\vx_{t+1} \rangle-\langle \vb,\vx_{t} \rangle+ \langle \vb,\vx_{t} \rangle-\langle \vc,\vx_{t+1} \rangle\\
     &\leq  \langle \vb,\vx_{t+1}-\vx_t \rangle + \langle \vc,\vx_{t} \rangle-\langle \vc,\vx_{t+1} \rangle= \langle \vb,\vx_{t+1}-\vx_t \rangle+ \langle \vc,\vx_{t}-\vx_{t+1} \rangle.
    \end{aligned}
\end{equation*}
Using Lemma \ref{lemma: consecutive strategis of no-regret algorithm} such that $\|\vx_{t+1}-\vx_t\|_1=O(\frac{1}{\sqrt{T}})$ and $\vb, \vc \in [0,1]^n$ we then have:
\begin{equation*}
    \begin{aligned}
     &\langle \vb,\vx_{t+1} \rangle- \langle \vc, \vx_{t+1}\rangle=\langle \vb,\vx_{t+1}-\vx_t \rangle+ \langle \vc,\vx_{t}-\vx_{t+1} \rangle\\
     &\leq \|\vx_{t+1}-\vx_t\|_1+\|\vx_{t}-\vx_{t+1}\|_1 \leq 2 O(\frac{1}{\sqrt{T}}) = O(\frac{1}{\sqrt{T}}).
    \end{aligned}
\end{equation*}
The proof is complete.
\end{proof}
\subsection{Missing Proofs of Last Round Convergence of AMWU}\label{appendix: missing proof of AMWU}
\subsubsection{Decreasing K-L distance}
In this subsection, part of our analysis bases on the linear variant of AMWU with the following update rule:
\[\vf_{t+1}(i)= \frac{\vf_{t}(i)(1+\eta ((\alpha+1){e_i}^\top\mA \vy_t-\alpha{e_i}^\top \mA\vy_{t-1}))}{\sum_{j} \vf_{t}(j)(1+\eta ((\alpha+1){e_j}^\top\mA \vy_t-\alpha{e_j}^\top \mA\vy_{t-1}))}.\]
Since the variant' update rule does not contain the exponential part, it reduces the complexity in the analysis.
We first quantify the distance between two consecutive updates of AMWU by the following lemma:
\begin{lemma}\label{proof of distance between two updates}
Let $\vf \in \Delta_n$ be the vector of the max player, $\vw,\vz \in \Delta_m$ such that $\|\vw-\vz\|_1=O(\eta)$, $\eta \alpha=O(1)$ and suppose $\vf', \vf''$ are the next iterates of AMWU and its linear variant with current vector $\vf$ and vectors $\vw, \vz$ of the min player. It holds that
\[\|\vf'-\vf''\|_1\; \text{is}\; O(\eta^2)\;\text{and}\; \|\vf'-\vf\|_1, \; \|\vf''-\vf\|_1\;\text{are}\; O(\eta).\]
Analogously, it holds for vector $\vy \in \Delta_m$ of the min player and its next iterates.
\end{lemma}
\begin{proof} 
Let $\eta$ be sufficiently small (smaller than maximum in absolute value entry of $\mA$). From the assumption that $\|\vw-\vz\|_1=O(\eta)$and $O(\eta \alpha)=O(1)$ we have:
\[(\alpha+1)(\mA\vw)_i-\alpha(\mA\vz)_i=(\mA\vw)_i+O(1).\]
Thus, we can derive the following equalities:
\begin{equation*}
    \begin{aligned}
     &|\vf'_i-\vf''_i|=\vf_i\left|\frac{e^{\eta((\alpha+1)(\mA\vw)_i-\alpha(\mA \vz)_i)}}{\sum_{j}\vf_j e^{\eta((\alpha+1)(\mA\vw)_j-\alpha(\mA \vz)_j)}}-\frac{1+\eta((\alpha+1)(\mA\vw)_i-\alpha (\mA\vz)_i)}{\sum_{j}\vf_j(1+\eta((\alpha+1)(\mA\vw)_j-\alpha (\mA\vz)_j))}\right| \\
     &=\vf_i \left|\frac{1+\eta((\alpha+1)(\mA\vw)_i-\alpha (\mA\vz)_i)\pm O(\eta^2)}{\sum_{j}\vf_j(1+\eta((\alpha+1)(\mA\vw)_j-\alpha (\mA\vz)_j))\pm O(\eta^2)}-\frac{1+\eta((\alpha+1)(\mA\vw)_i-\alpha (\mA\vz)_i)}{\sum_{j}\vf_j(1+\eta((\alpha+1)(\mA\vw)_j-\alpha (\mA\vz)_j))}\right|\\
     &=\vf_i O(\eta^2).
    \end{aligned}
\end{equation*}
and hence $\|\vf'-\vf''\|_1$ is $O(\eta^2)$. Moreover we have that
\begin{equation*}
    \begin{aligned}
     &|\vf_i-\vf''_i|=\vf_i\left|1-\frac{1+\eta((\alpha+1)(\mA\vw)_i-\alpha (\mA\vz)_i)}{\sum_{j}\vf_j(1+\eta((\alpha+1)(\mA\vw)_j-\alpha (\mA\vz)_j))}\right| \\
     &=\vf_i \left|\frac{\sum_{j}\vf_j(1+\eta((\alpha+1)(\mA\vw)_j-\alpha (\mA\vz)_j))-(1+\eta((\alpha+1)(\mA\vw)_i-\alpha (\mA\vz)_i))}{\sum_{j}\vf_j(1+\eta((\alpha+1)(\mA\vw)_j-\alpha (\mA\vz)_j))}\right| \\
     &=\vf_i \left| \frac{\sum_{j}\vf_j(\eta((\alpha+1)(\mA\vw)_j-\alpha (\mA\vz)_j))-\eta((\alpha+1)(\mA\vw)_i-\alpha (\mA\vz)_i)}{\sum_{j}\vf_j(1+\eta((\alpha+1)(\mA\vw)_j-\alpha (\mA\vz)_j))}\right|\\
     &=\vf_i \left| \frac{\eta\left(\sum_{j}\vf_j((\alpha+1)(\mA\vw)_j-\alpha (\mA\vz)_j)-((\alpha+1)(\mA\vw)_i-\alpha (\mA\vz)_i)\right)}{\sum_{j}\vf_j(1+\eta((\alpha+1)(\mA\vw)_j-\alpha (\mA\vz)_j))}\right| \\
     &=\vf_i O(\eta).
    \end{aligned}
\end{equation*}
We can derive the third part of the lemma by using the triangle inequality with the two above proofs.
\end{proof}

\begin{lemma}\label{lemma 3}
Let $(\vf_t,\vy_t)$ denote the t-th iterate of AMWU dynamics. It holds for $t \geq 2$ that
\begin{equation*}
    \begin{aligned}
     &{\vf^*}^\top \mA ((\alpha+1)\vy_t-\alpha \vy_{t-1})\geq {\vf^*}^\top \mA \vy^* \; \text{and}\\
     &((\alpha+1)\vf_t-\alpha \vf_{t-1})^\top \mA \vy^* \leq {\vf^*}^\top \mA \vy^*
    \end{aligned}
\end{equation*}
\end{lemma}

\begin{proof}

It is sufficient to show that $((\alpha+1)\vy_t-\alpha \vy_{t-1}) \in \Delta_m$ and $((\alpha+1)\vf_t-\alpha \vf_{t-1}) \in \Delta_n$.

From Lemma \ref{distance between two updates} we have $\vf_t(i)=(1-O(\eta))\vf_{t-1}(i)$. Thus, in order to show that $((\alpha+1)\vf_t(i)-\alpha \vf_{t-1}(i)) \geq 0$ we need to show that:
\begin{equation}
    \begin{aligned}
     (1-O(\eta)) \geq{ \frac{\alpha}{\alpha+1}}\\
     \iff 1\geq (\alpha+1) O(\eta),
    \end{aligned}
\end{equation}
which is true since $\alpha \eta =\eta^b, b \in [0,1]$ and $\eta$ is small enough.
\end{proof}
\begin{theorem}\label{proof of thm: theorem KL decrease}
Let $(\vf^*,\vy^*)$ be the unique optimal minimax equilibrium and $\eta$ suffciently small. Assume that $\alpha \eta=\eta^b$ where $b\in [0,1]$.Then
\[RE((\vf^*,\vy^*)||(\vf_t,\vy_t))\]
is decreasing with time $t$ by $\eta^{2+b}$ unless $(\vf_t, \vy_t)$ is $O(\eta^{b/3})-close$.
\end{theorem}
\begin{proof}
We compute the difference in relative entropy distance between two connected strategies:
\begin{equation}
    \begin{aligned}
     &RE((\vf^*,\vy^*)||(\vf_{t+1},\vy_{t+1}))-RE((\vf^*,\vy^*)||(\vf_t,\vy_t))\\
     &=-\left( \sum_{i}\vf^*(i)\log(\frac{\vf_{t+1}(i)}{\vf_{t}(i)})+\sum_{i}\vy^*(i)\log(\frac{\vy_{t+1}(i)}{\vy_{t}(i)})\right)\\
     &=-\left(\sum_{i} \vf^*(i) \log(e^{\eta((\alpha+1)\mA\vy_t-\alpha \mA \vy_{t-1})(i)})+\sum_{i}\vy^*(i)\log(e^{\eta (-(\alpha+1)\mA^\top\vf_t+\alpha\mA^\top \vf_{t-1})(i)})\right)\\
     &+\log\left(\sum_{i}\vf_t(i)e^{\eta((\alpha+1)\mA\vy_t-\alpha \mA \vy_{t-1})(i)}\right)+\log\left(\sum_{i}\vy_t(i)e^{\eta (-(\alpha+1)\mA^\top\vf_t+\alpha\mA^\top \vf_{t-1})(i)}\right)\\
     &=-\eta {x^*}^\top \mA((\alpha+1)\vy_t-\alpha \vy_{t-1})-\eta {y^*}^\top \mA^\top (-(\alpha+1)\vf_t+\alpha \vf_{t-1})+\\
     &\log\left(\sum_{i}\vf_t(i)e^{\eta((\alpha+1)\mA\vy_t-\alpha \mA \vy_{t-1})(i)}\right)+\log\left(\sum_{i}\vy_t(i)e^{\eta (-(\alpha+1)\mA^\top\vf_t+\alpha\mA^\top \vf_{t-1})(i)}\right).
    \end{aligned}
\end{equation}
From Lemma \ref{lemma 3} we have:
\[-\eta {\vf^*}^\top \mA((\alpha+1)\vy_t-\alpha \vy_{t-1})-\eta {\vy^*}^\top \mA^\top (-(\alpha+1)\vf_t+\alpha \vf_{t-1})\leq 0.\]
Thus we have:
\begin{equation}
    \begin{aligned}
     &RE((\vf^*,\vy^*)||(\vf_{t+1},\vy_{t+1}))-RE((\vf^*,\vy^*)||(\vf_t,\vy_t))\\
     &\leq \log\left(\sum_{i}\vf_t(i)e^{\eta((\alpha+1)\mA\vy_t-\alpha \mA \vy_{t-1})(i)}\right)+\log\left(\sum_{i}\vy_t(i)e^{\eta (-(\alpha+1)\mA^\top\vf_t+\alpha\mA^\top \vf_{t-1})(i)}\right) \\
     &=\log\left(\sum_{i}\vf_t(i)e^{\eta((\alpha+1)((\mA\vy_t)(i)-\vf_t^{\top}\mA \vy_t)-\alpha ((\mA \vy_{t-1})(i)-\vf_t^{\top}\mA\vy_{t-1}))} \right)\\
     &+\log\left(\sum_{i}\vy_t(i)e^{\eta (-(\alpha+1)((\mA^\top\vf_t)(i)-\vf_t^{\top}\mA \vy_t)+\alpha((\mA^\top \vf_{t-1})(i)-\vf_{t-1}^{\top}\mA\vy_t))}\right) +\alpha\eta (\vf_{t-1}^{\top}\mA\vy_t-\vf_t^{\top}\mA\vy_{t-1}) \\
     &=\log\left(\sum_{i}\vf_t(i)e^{\eta((e_i-\vf_t)^\top\mA ((\alpha+1)\vy_t-\alpha \vy_{t-1}))} \right)+
     \log\left(\sum_{i}\vy_t(i)e^{\eta ((-(\alpha+1)\vf_t+\alpha \vf_{t-1})^\top \mA (e_i-\vy_t))}\right)\\
     &+\eta^b (\vf_{t-1}^{\top}\mA\vy_t-\vf_t^{\top}\mA\vy_{t-1}).
    \end{aligned}
\end{equation}
Using the Taylor approximation ($\eta$ is suffciently small) to the function $e^x$ (i.e., $e^x=1+x+\frac{1}{2}x^2$) and $\log(1+x)<x$ for $x>0$, we then have:
\begin{equation}
    \begin{aligned}
     &=\log\left(\sum_{i}\vf_t(i)e^{\eta((e_i-\vf_t)^\top\mA ((\alpha+1)\vy_t-\alpha \vy_{t-1}))} \right)\\
     &\leq \log(\sum_{i}x_t(i)(1+\eta((e_i-\vf_t)^\top\mA ((\alpha+1)\vy_t-\alpha \vy_{t-1})))+\\
     &\sum_{i}\vf_t(i)((\frac{1}{2}+O(\eta^b)\eta^2((e_i-\vf_t)^\top\mA ((\alpha+1)\vy_t-\alpha \vy_{t-1}))^2 )\\
     &=\log \left(1+ \sum_{i}\vf_t(i)((\frac{1}{2}+O(\eta^b)\eta^2((e_i-\vf_t)^\top\mA ((\alpha+1)\vy_t-\alpha \vy_{t-1}))^2\right)\\
     &\leq \sum_{i}\vf_t(i)((\frac{1}{2}+O(\eta^b)\eta^2((e_i-\vf_t)^\top\mA ((\alpha+1)\vy_t-\alpha \vy_{t-1}))^2.
    \end{aligned}
\end{equation}
Along with Lemma \ref{Lemma 10}, we then have:
\begin{equation}
    \begin{aligned}
     &RE((\vf^*,\vy^*)||(\vf_{t+1},\vy_{t+1}))-RE((\vf^*,\vy^*)||(\vf_t,\vy_t))\\
     &\sum_{i}(\frac{1}{2}+O(\eta^b))\eta^2\vf_t(i)((e_i-\vf_t)^\top\mA ((\alpha+1)\vy_t-\alpha \vy_{t-1}))^2+\\
     &\sum_{i} (\frac{1}{2}+O(\eta^b))\eta^2\vy_{t}(i)((\vy_t-e_i)^\top \mA^\top((\alpha \vy_{t-1}-(\alpha+1)\vy_{t}))^2\\
     &-\frac{\eta^b}{\eta}(1-O(\eta))\eta^2 \sum_{i}\vf_t(i)((\vf_t-e_i)^\top \mA ((\alpha+1)\vy_{t}-\alpha \vy_{t-1}))^2-\\
     &\frac{\eta^b}{\eta}(1-O(\eta))\eta^2 \sum_{i} \vy_{t}(i)((\vy_t-e_i)^\top \mA^\top((\alpha \vy_{t-1}-(\alpha+1)\vy_{t}))^2+\frac{\eta^b}{\eta}\eta^b O(\eta^2)\\
     &\leq -(\frac{1}{2}-O(\eta^b))\eta^2 \sum_{i}\vf_t(i)((\vf_t-e_i)^\top \mA ((\alpha+1)\vy_{t}-\alpha \vy_{t-1}))^2-\\
     &(\frac{1}{2}-O(\eta^b))\eta^2 \sum_{i} \vy_{t}(i)((\vy_t-e_i)^\top \mA^\top((\alpha \vy_{t-1}-(\alpha+1)\vy_{t}))^2+\eta^b O(\eta^2).
    \end{aligned}
\end{equation}
Since $\frac{\eta^b}{\eta} >1$. 
Now, it is clear that as long as $(\vf_t,\vy_t)$ and thus $(\vf_{t-1},\vy_{t-1})$ is not $O(\eta^{b/3})$-close, from the above inequalities we get:
\[RE((\vf^*,\vy^*)||(\vf_{t+1},\vy_{t+1}))-RE((\vf^*,\vy^*)||(\vf_t,\vy_t)) \leq -\Omega(\eta^{b+2}),\]
or the relative entropy distance decreases at least a factor of $\eta^{b+2}$ and the claim follows.
\end{proof}
\subsubsection{$\eta^{b/3}$-closeness implies closeness to optimum}
We first need the following lemma:
\begin{lemma}
Let $i \in \operatorname{Supp}(\vf^*)$ and $j \in \operatorname{Supp}(\vy^*)$. It holds that $x_T(i)\geq \frac{1}{2}\eta^{b/3}$ and $y_T(i)\geq \frac{1}{2}\eta^{b/3}$ as long as
\[\eta^{b/3} \leq \min_{s \in \operatorname{Supp}(\vf^*)}\frac{1}{(nm)^{1/\vf^*(s)}},\; \min_{s \in \operatorname{Supp}(\vy^*)}\frac{1}{(nm)^{1/\vy^*(s)}}.\]
\end{lemma}

Using the above lemma, we can follow the same argument as in Theorem 3.2 of \cite{Daskalakis2018c} to prove the following theorem:
\begin{theorem}\label{proof of theorem close implies close to optimum}
Assume $(\vf^*, \vy^*)$ is unique optimal solution of the problem. Let T be the first time KL divergence does not decrease by $\Omega(\eta^{b+2})$. It follows that as $\eta \to 0$, the $\eta^{b/3}$-close point $(\vf_T,\vy_T)$ has distance from $(\vf^*,\vy^*)$ that goes to zero:
\[\lim_{\eta \to 0} \|(\vf^*, \vy^*)-(\vf_T, \vy_T)\|_1=0.\]
\end{theorem}
\begin{proof}
From Lemma \ref{lemma: close lemma 1} and the definition of $T$ we have $|(\mA \vy_T)_i-{\vf_T}^\top \mA \vy_T|$ is $O(\eta^{1/3})$ for $i$ in support of $\vf^*$ and $|(\vf_T^\top \mA)_j-\vf_T^\top \mA \vy_T|$ is $O(\eta^{1/3})$ for $j$ in support of $\vy^*$. We consider $(\vw_T,\vz_T)$ the project of $(\vf_T,\vy_T)$ by removing all the coordinates with mass less than $\frac{1}{2}\eta^{b/3}$ and rescales it. We have the following relationship:
\begin{equation}\label{eq: distance-close 1}
\lim_{\eta \to 0}\|(\vf_T, \vy_T)- (\vw_T,\vz_T)\|=0.    
\end{equation}
Since for all the cordinates in $\vw$ and $\vz$, it holds that $|(\hat{\mA} \vz_T)_i-{\vw_T}^\top \hat{\mA} \vz_T|$ and $|(\vw_T^\top \hat{\mA})_j-\vw_T^\top \hat{\mA} \vz_T|$ are $O(\eta^{b/3})$, thus $(\vw, \vz)$ is $O(\eta^{b/3})$-approximate solution of the game $\hat{\mA}$.
Using the following lemma:
\begin{lemma}[Claim 3.5 in \cite{Daskalakis2018c}]
Let $(\vx^*,\vy^*)$ be the unique optimal solution of the game. For every $\epsilon>0$, there exists an $\gamma$ so that for every $\gamma$-approximate solution $(\vx,\vy)$ we get that $|x_i-x^*_1| < \epsilon$ for all $i \in [n]$. Analogously holds for player $\vy$.
\end{lemma}
Using the above lemma with $\epsilon=\eta^{b/3}$ and sufficiently small $\eta$, we have $|w_i|<\eta^{b/3}$ for every $i$ not in the support of $\vx^*$. 
Since the subgame $\hat{\mA}$ contains all the pure strategy in the NE support of game $\mA$, subgame $\hat{\mA}$ will also have a unique NE with the same weight as in the game $\mA$. Thus we have:
\begin{equation}\label{eq: distance-close 2}
\lim_{\eta \to 0}  \|(\vw_T,\vz_T)-(\vf^*, \vy^*)\|=0.    
\end{equation}
Combining Equation (\ref{eq: distance-close 1}) and (\ref{eq: distance-close 2}) gives us the proof.
\end{proof}
\subsubsection{Proof of local convergence}
We use the following well-known fact in dynamical systems to prove the local convergence:
\begin{proposition}[see \cite{galor2007discrete}]\label{prop: dynamical system convergence}
If the Jacobian of the continuously differential update rule $w$ at a fixed point $\vz$ has spectral radius less than one, then there exists a neighborhood $U$ around $\vz$ such that for all $\vx \in U$, the dynamic converges to $\vz$.
\end{proposition}
Given this, our local convergence theorem states:
\begin{theorem}\label{proof of theorem contraction}
Let$(\vf^*, \vy^*)$ be the unique minimax equilibrium of the game $\mA$. There exists a neighborhood of $(\vf^*, \vy^*)$ such that the E-OMWU dynamics converge.
\end{theorem}
\begin{proof}
The update rule of AMWU can be described as the following dynamical system:
\begin{equation}\label{eq: dynamical system update rule}
    \begin{aligned}
     &g(\vf,\vy,\vz,\vw):= (g_1(\vf,\vy,\vz,\vw),g_2(\vf,\vy,\vz,\vw),g_3(\vf,\vy,\vz,\vw)g_4(\vf,\vy,\vz,\vw)) \\
     &g_{1,i}(\vf,\vy,\vz,\vw):= (g_1(\vf,\vy,\vz,\vw))_i:= \vf_i\frac{e^{\eta ((\alpha+1){e_i}^\top\mA\vy-\alpha{e_i}^\top\mA\vw)}}{\sum_i f_i e^{\eta ((\alpha+1){e_i}^\top\mA \vy-\alpha{e_i}^\top\mA\vw)}}\; \forall i \in [n] \\
     &g_{2,i}(\vf,\vy,\vz,\vw):= (g_2(\vf,\vy,\vz,\vw))_i:= \vy_i\frac{e^{-\eta ((\alpha+1){e_i}^\top\mA \vx-\alpha{e_i}^\top\mA \vz)}}{\sum_i \vy_i e^{-\eta ((\alpha+1){e_i}^\top\mA \vx-\alpha{e_i}^\top\mA \vz)}}\; \forall i \in [m] \\
     &g_3(\vf,\vy,z,w):=\mI_{n \times n} \vf \\
     &g_4(\vf,\vy,z,w):=\mI_{m\times m}\vy.
    \end{aligned}
\end{equation}
It is easy to show that $(\vf^*, \vy^*, \vf^*, \vy^*)$ is the stationary point.
Following Proposition \ref{prop: dynamical system convergence} it is sufficient to prove that that the eigenvalue of the Jacobian matrix of g at $(\vf^*, \vy^*, \vf^*, \vy^*)$ is less than $1$.

We now calculate the Jacobian matrix of g at the point $(\vf^*, \vy^*, \vf^*, \vy^*)$ and show that the spectral radius less than one. We study the Jacobian computed at the stationary point $(\vf^*, \vy^*, \vf^*, \vy^*)$.

Let $v$ be the value of the game and $\vf^*, \vy^*$ is the unique minimax equilibrium (i.e ${\vf^*}^\top\mA \vy^*=v$). For $i \notin Supp(\vf^*)$(e.g. $\vf^*_i=0$), we have
\[\frac{\partial g_{1,i}}{\partial f_i} = \frac{e^{\eta (\mA \vy^*)_i}}{\sum_t \vf^*_t e^{\eta (\mA \vy^*)_t}}= \frac{e^{\eta (\mA \vy^*)(i)}}{e^{\eta v}}\]

and other partial derivatives equal to zero. Therefore, $\frac{e^{\eta (\mA \vy^*_{i})}}{e^{\eta v}}<1$ is an eigenvalue of the Jacobian computed at the optimal solution(e.g. Due to the uniqueness, $\mA \vy^*_i < v$). Similarly, we have for $j \notin Supp(\vy^*)$, $\frac{\partial \vg_{2,j}}{\partial \vy_j}= \frac{e^{-\eta (\mA^\top \vx^*)_j }}{e^{-\eta v}} <1$ is an eigenvalue of the Jacobian matrix. By removing the row and columns corresponding to above eigenvalue, we create a matrix J containing only the elements in the support of $\vf^*$ and $\vy^*$. From above, it is clear that the spectral radius of the Jacobian matrix less than 1 iff the spectral of the new matrix $J$ less than 1. Denote $D_x, D_y$ be the diagonal matrix containing non-zero element of $\vf^*$ and $\vy^*$ respectively. Let $\mB$ be the submatrix of of payoff $\mA$ corresponding to non-zero element of $\vf^*, \vy^*$. We then have the matrix J as follow:
\[ \tiny{ A=\begin{bmatrix}
    \mI_{k_1 \times k_1}-D_x 1_{k} 1_{k}^T & \eta (\alpha+1) D_x(\mB-v 1_{k_1} 1_{k_2}^{\top}) & 0_{k1 \times k1} & -\eta \alpha D_x(\mB-v 1_{k_1} 1_{k_2}^T)\\
    (\alpha+1)\eta D_y(v 1_{k_2}1_{k_1}^{\top}-\mB^{\top}) & \mI_{k_2 \times k_2}-D_y 1_{k_2}1_{k_2}^{\top} & -\eta \alpha D_y(v 1_{k_2}1_{k_1}^{\top}-\mB^{\top})& 0_{k_2 \times k_2}\\
    \mI_{k_1 \times k_1}& 0_{k_1 \times k_2}& 0_{k_1 \times k_1}& 0_{k1 \times k_2}\\
    0_{k_2 \times k_1} & \mI_{k_2 \times k_2}& 0_{k_2 \times k_1}& 0_{k_2 \times k_2}
  \end{bmatrix}
  }
\]
It is clear that $(1_{k_1},0_{k_2}, 0_{k_1},0_{k_2}),(0_{k_1},1_{k_2}, 0_{k_1},0_{k_2})$ are left eigenvectors with eigenvalues zero and thus any right eigenvector $(\vf,\vy,\vz,\vw)$ with nonzero eigenvalue has the property that $\vf^\top 1_{k_1}=0$ and $\vy^\top 1_{k_2}=0$. Thus, every nonzero eigenvalue of the matrix above is an eigenvalue of the following matrix:
\[J_{new}=\begin{bmatrix}
    \mI_{k_1 \times k_1}& \eta (\alpha+1) D_x\mB & 0_{k1 \times k1} & -\eta \alpha D_x\mB\\
    -(\alpha+1)\eta D_y\mB^{\top} & \mI_{k_2 \times k_2} & \eta \alpha D_y\mB^{\top}& 0_{k_2 \times k_2}\\
    \mI_{k_1 \times k_1}& 0_{k_1 \times k_2}& 0_{k_1 \times k_1}& 0_{k1 \times k_2}\\
    0_{k_2 \times k_1} & \mI_{k_2 \times k_2}& 0_{k_2 \times k_1}& 0_{k_2 \times k_2}
  \end{bmatrix}
\]
Using the determinant of block matrix we have the characteristic polynomial of the matrix:
\[J_{new}=(-1)^k \operatorname{det}\left(\begin{bmatrix}
    \lambda(1-\lambda)\mI_{k_1\times k_1} & \eta (\lambda(\alpha+1)-\alpha)D_x \mB \\
    -\eta(\lambda(\alpha+1)-\alpha)D_y \mB^\top & \lambda(1-\lambda)\mI_{k_2\times k_2}
  \end{bmatrix}\right)
\]
This equivalent to 
\[(\alpha-(\alpha+1)\lambda)^k q\left(\frac{\lambda(\lambda-1)}{(\alpha+1)\lambda-\alpha}\right),\]
where $q(\lambda)$ is the characteristic polynomial of
\[J_{small}= \left(\begin{bmatrix}
    0_{k_1\times k_1} & \eta D_x \mB \\
    -\eta D_y \mB^\top & 0_{k_2\times k_2}
  \end{bmatrix}\right)
\]
Following Lemma B.6 in \cite{Daskalakis2018c}, we then have $J_{small}$ has eigenvalues of the form $\pm i\eta \tau$ with $\tau \in \mathcal{R}$. Denote $\sigma:=\eta \tau$ and thus $\sigma$ and $\sigma \alpha$ can be sufficiently small in absolute value. We derive that any nonzero eigenvalue $\lambda$ of the matrix J will satisfy:
\begin{equation*}
    \begin{aligned}
     &\frac{\lambda(\lambda-1)}{(\alpha+1)\lambda-\alpha}=i\sigma\\
     &\iff \lambda^2-\lambda (1+i\sigma(\alpha+1))+i\sigma\alpha=0\\
     &\lambda=\frac{1+i\sigma(\alpha+1)\pm \sqrt{1-\sigma^2(\alpha+1)^2-i2\sigma(\alpha-1)}}{2}.
    \end{aligned}
\end{equation*}
Suppose that $\sqrt{1-\sigma^2(\alpha+1)^2-i2\sigma(\alpha-1)}=x+iy$, then we can derive that in order to maximize the magnitude of $\lambda$ when $\sigma$ is relatively small, we have:
\[x=\sqrt{\frac{1-\sigma^2(\alpha+1)^2+\sqrt{(1-\sigma^2(\alpha+1)^2)^2+4\sigma^2(\alpha-1)^2}}{2}},\;y=\frac{-\sigma(\alpha-1)}{x}\]
Thus, the square of magnitude of $\lambda$ will be:
\begin{equation*}
    \begin{aligned}
     \frac{(1+x)^2+(\sigma(\alpha+1)+y)^2}{4}
    \end{aligned}
\end{equation*}
We note that for sufficiently small $\sigma$:
\begin{equation*}
    \begin{aligned}
     &x=\sqrt{\frac{1-\sigma^2(\alpha+1)^2+\sqrt{(1+\sigma^2(\alpha+1)^2)^2-16 \sigma^2\alpha}}{2}}\\
     &\leq \sqrt{\frac{1-\sigma^2(\alpha+1)^2+(1+\sigma^2(\alpha+1)^2)-2 \sigma^2\alpha}{2}}\\
     &=\sqrt{1-\sigma^2\alpha}
    \end{aligned}
\end{equation*}
Furthermore, we have:
\begin{equation*}
    \begin{aligned}
    &x=\sqrt{\frac{1-\sigma^2(\alpha+1)^2+\sqrt{(1+\sigma^2(\alpha+1)^2)^2-16 \sigma^2\alpha}}{2}}\\
    &\geq{\sqrt{\frac{1-\sigma^2(\alpha+1)^2+(1+\sigma^2(\alpha+1)^2)-8 \sigma^2\alpha}{2}}}\\
    &=\sqrt{1-4\sigma^2\alpha}.
    \end{aligned}
\end{equation*}
Since $\sqrt{1-4\sigma^2\alpha}\leq x\leq 1$ we have:
\begin{equation*}
    \begin{aligned}
     \frac{-\sigma(\alpha-1)}{\sqrt{1-4\sigma^2\alpha}}\leq y=\frac{-\sigma(\alpha-1)}{x} \leq -\sigma(\alpha-1).
    \end{aligned}
\end{equation*}
We will prove that:
\begin{equation*}
\begin{aligned}
 &\sigma(\alpha+1)+ \frac{-\sigma(\alpha-1)}{\sqrt{1-4\sigma^2\alpha}} \geq 0\\
    &\iff(\alpha+1)\geq \frac{(\alpha-1)}{\sqrt{1-4\sigma^2\alpha}}\\
    &\iff(\alpha^2+2\alpha+1)(1-4\sigma^2\alpha) \geq (\alpha-1),
\end{aligned}
\end{equation*}
which is true since $\sigma$ and $\sigma \alpha$ can set sufficiently small.
Thus we have:
\[0 \leq \sigma(\alpha+1)+y \leq 2\sigma\]
We then have:
\begin{equation*}
    \begin{aligned}
     &\frac{(1+x)^2+(\sigma(\alpha+1)+y)^2}{4}\leq \frac{(1+\sqrt{1-4\sigma^2\alpha})^2+(2\sigma)^2}{4}\\
     &\leq \frac{2+2\sqrt{1-4\sigma^2\alpha}-4\sigma^2\alpha+4\sigma^2}{4}\leq 1,
    \end{aligned}
\end{equation*}
Since $\alpha\geq 1$ and the equality happens only when $\sigma=0$. For $\sigma=0$, it means that $J_{new}$ has an eigenvalue which is equal to one. Suppose $(\hat{\vx},\hat{\vy},\hat{\vz},\hat{\vw})$ is the corresponding eigenvector. We then have $\mI \hat{\vx}-\mI \hat{\vz}=0$ and $\mI \hat{\vy}-\mI \hat{\vw}=0$, thus we derive that: $\hat{\vx}=\hat{\vz}$ and $\hat{\vy}=\hat{\vw}$. Furthermore, we also have: $D_x \mB\hat{\vx}=0$ and $D_y \mB^\top \hat{\vy}=0$, thus we have $\mB\hat{\vx}=0$ and $\mB^\top \hat{\vy}=0$. From previous argument, we also have: $\hat{\vx}^\top 1_{k_1}=0$ and $\hat{\vy}^\top 1_{k_2}=0$. Thus, the strategy $(\vx^*,\vy^*)+t(\hat{\vx},\hat{\vy})$ also an optimal strategy for small enough $t$ to make every element non-negative. Since the assumption of uniqueness, we then have $\hat{\vx} =0,\hat{\vy}=0$, contradiction. Thus, every eigenvalue of matrix $J$ has magnitude of less than 1. The proof is complete.
\end{proof}
\textbf{Derivatives calculation}

Set $S_{\vx}=\sum_i f_i e^{\eta ((\alpha+1){e_i}^\top\mA \vy-\alpha{e_i}^\top\mA w)}$ and $S_{\vy}=\sum_i \vy_i e^{-\eta ((\alpha+1){e_i}^\top\mA \vx-\alpha{e_i}^\top\mA \vz)}$. The derivative at $(\vf^*, \vy^*, \vf^*, \vy^*)$ is as follow:
\begin{equation}
    \begin{aligned}
    &\frac{\partial g_{1,i}}{\partial f_i}=\frac{e^{\eta ((\alpha+1){e_i}^\top\mA \vy-\alpha{e_i}^\top\mA \vw)}}{S_{\vx}}-f_i \frac{e^{2\eta ((\alpha+1){e_i}^\top\mA \vy-\alpha{e_i}^\top\mA \vw)}}{{S_{\vx}}^2} \;\;\forall i \in [n],\\
    &\frac{\partial g_{1,i}}{\partial \vx_j}=f_i e^{\eta ((\alpha+1){e_i}^\top\mA \vy-\alpha{e_i}^\top\mA \vw)} \frac{-e^{\eta ((\alpha+1){e_j}^\top\mA \vy-\alpha{e_j}^\top\mA \vw)}}{{S_{\vx}}^2} \;\;\forall i \in [n], j \in [m], j \neq i,\\
    &\frac{\partial g_{1,i}}{\partial \vy_j}=f_i e^{\eta ((\alpha+1){e_i}^\top\mA \vy-\alpha{e_i}^\top\mA \vw)}\frac{\eta(\alpha+1)\mA _{i,j}  S_{\vx}- \eta (\alpha+1)\sum_{t} \mA_{tj}\vx_te^{\eta ((\alpha+1){e_t}^\top\mA  \vy-\alpha{e_t}^\top\mA \vw)}}{S_{\vx}^2} \;\;\forall i \in [n], j=i,\\
     &\frac{\partial g_{1,i}}{\partial \vz_j}=0 \;\;\forall i,j \in [n],\\
    &\frac{\partial g_{1,i}}{\partial \vw_j}=f_i e^{\eta ((\alpha+1){e_i}^\top\mA \vy-\alpha{e_i}^\top\mA \vw)} \frac{-\alpha \eta \mA_{ij}S_\vx+\eta \alpha \sum_{t} \mA_{tj}\vx_te^{\eta ((\alpha+1){e_t}^\top\mA \vy-\alpha{e_t}^\top\mA \vw)}}{S_\vx^2} \;\;\forall i \in [n], j \in [m].
    \end{aligned}
\end{equation}
\begin{equation}
    \begin{aligned}
     &\frac{\partial g_{2,i}}{\partial \vy_i}= \frac{e^{-\eta ((\alpha+1){e_i}^\top\mA \vx-\alpha{e_i}^\top\mA \vz)}}{S_\vy}-\vy_i \frac{e^{-2\eta ((\alpha+1){e_i}^\top\mA \vx-\alpha{e_i}^\top\mA \vz)}}{{S_\vy}^2} \;\;\forall i \in [m],\\
      &\frac{\partial g_{2,i}}{\partial \vy_j}=\vy_i e^{-\eta ((\alpha+1){e_i}^\top\mA \vx-\alpha{e_i}^\top\mA \vz)} \frac{-e^{-\eta ((\alpha+1){e_i}^\top\mA \vx-\alpha{e_i}^\top\mA \vz)}}{{S_\vy}^2} \;\;\forall i \in [n], j \in [m], j \neq i,\\
      &\frac{\partial g_{2,i}}{\partial \vx_j}=\vy_i e^{-\eta ((\alpha+1){e_i}^\top\mA \vx-\alpha{e_i}^\top\mA \vz)}\frac{-\eta(\alpha+1)\mA_{i,j}  S_\vy+\eta (\alpha+1)\sum_{t} \mA_{tj}\vy_te^{-\eta ((\alpha+1){e_i}^\top\mA \vx-\alpha{e_i}^\top\mA \vz)}}{S_\vy^2} \;\;\forall i \in [m], j\in[n],\\
      &\frac{\partial g_{2,i}}{\partial \vz_j}=\vy_i e^{-\eta ((\alpha+1){e_i}^\top\mA \vx-\alpha{e_i}^\top\mA \vz)}\frac{\eta\alpha \mA_{i,j}  S_\vy-\eta \alpha \sum_{t} \mA_{tj}\vx_te^{-\eta ((\alpha+1){e_i}^\top\mA \vx-\alpha{e_i}^\top\mA \vz)}}{S_\vy^2} \;\;\forall i \in [m], j\in[n],\\
      &\frac{\partial g_{2,i}}{\partial \vw_j}=0 \;\;\forall i,j \in [m],\\
      &\frac{\partial g_{3,i}}{\partial f_i}=1 \;\;\forall i \in [n],\; \\
      &\frac{\partial g_{4,i}}{\partial \vy_i}=1 \;\;\forall i \in [m].\; 
    \end{aligned}
\end{equation}
\section{Experiment}\label{Appendix: experiment}
\subsection{Oblivious adversary}\label{Appendix: exp oblivious adversary} We specify our experiment setting as follow. In a chosen random matrix game, we first let the agent follows a fixed MWU against the adversary follows MWU with a chosen learning rate in the set: $[0.5,0.45,0.4,\dots,0.05]$~\footnote{Each learning rate will create different oblivious adversary.}. Then, we record the strategies of the adversary in each round and consider it as the oblivious adversary. To highlight the difference between AMWU and OMWU, we also test the performance of OWMU with learning rate $\eta = 1$. For the random games, we test it on $5$ random seeds for each matrix size. For the meta games, we run our algorithms against 5 different oblivious adversary (i.e., MWU with the learning rate in $[0.5, 0.4, 0.3, 0.2,0.1]$) and report the average performance as well as the standard deviation.

\begin{figure*}[t!]
     \centering
\begin{subfigure}[l]{.49\textwidth}
         \centering
         \includegraphics[width=1.\textwidth]{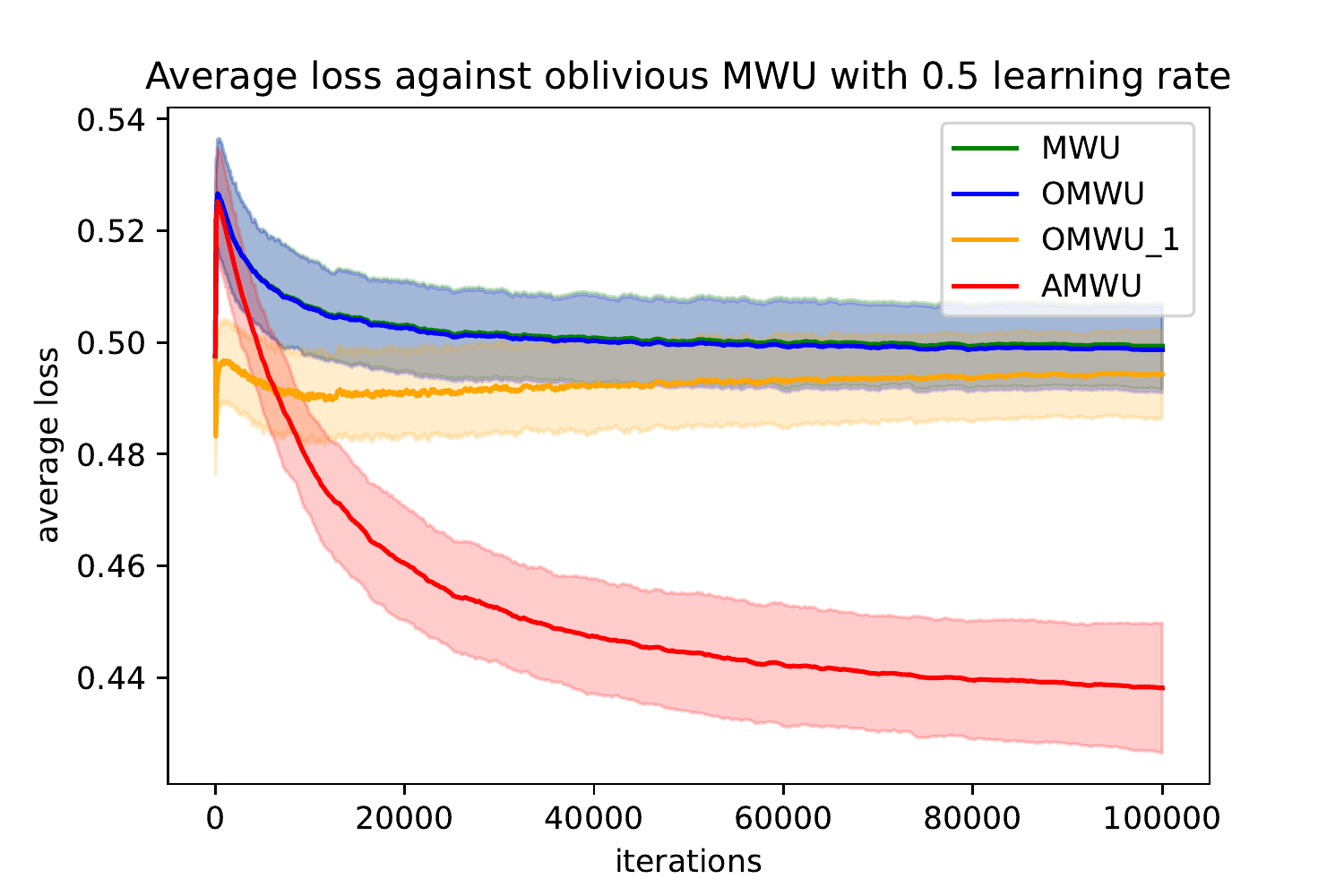}
\caption{0.5 learning rate MWU adversary in random game}
         
\end{subfigure}
\begin{subfigure}[l]{.49\textwidth}
         \centering
         \includegraphics[width=1.\textwidth]{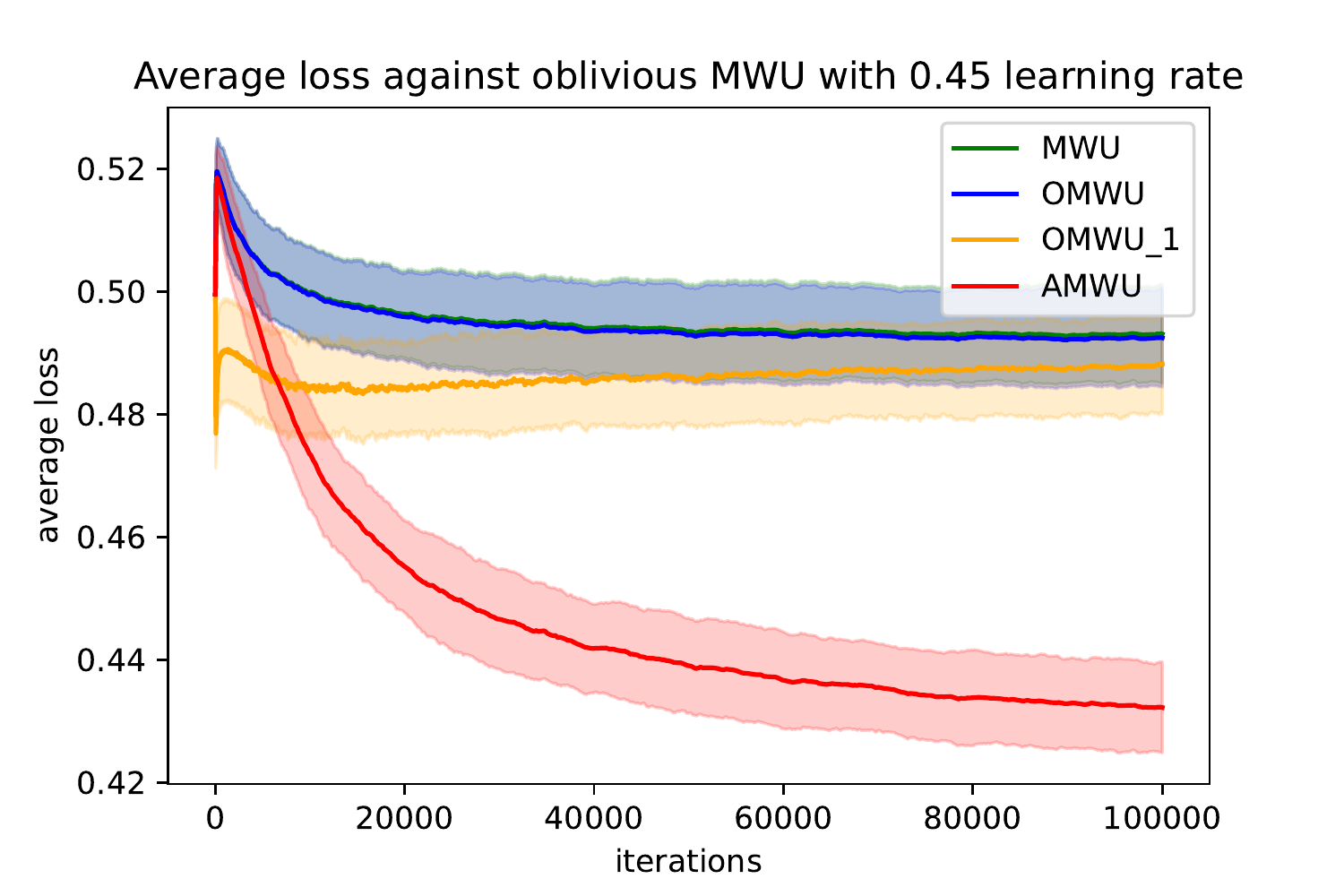}
                
\caption{0.45 learning rate MWU adversary}
\end{subfigure}
\begin{subfigure}[l]{.49\textwidth}
         \centering
         \includegraphics[width=1.\textwidth]{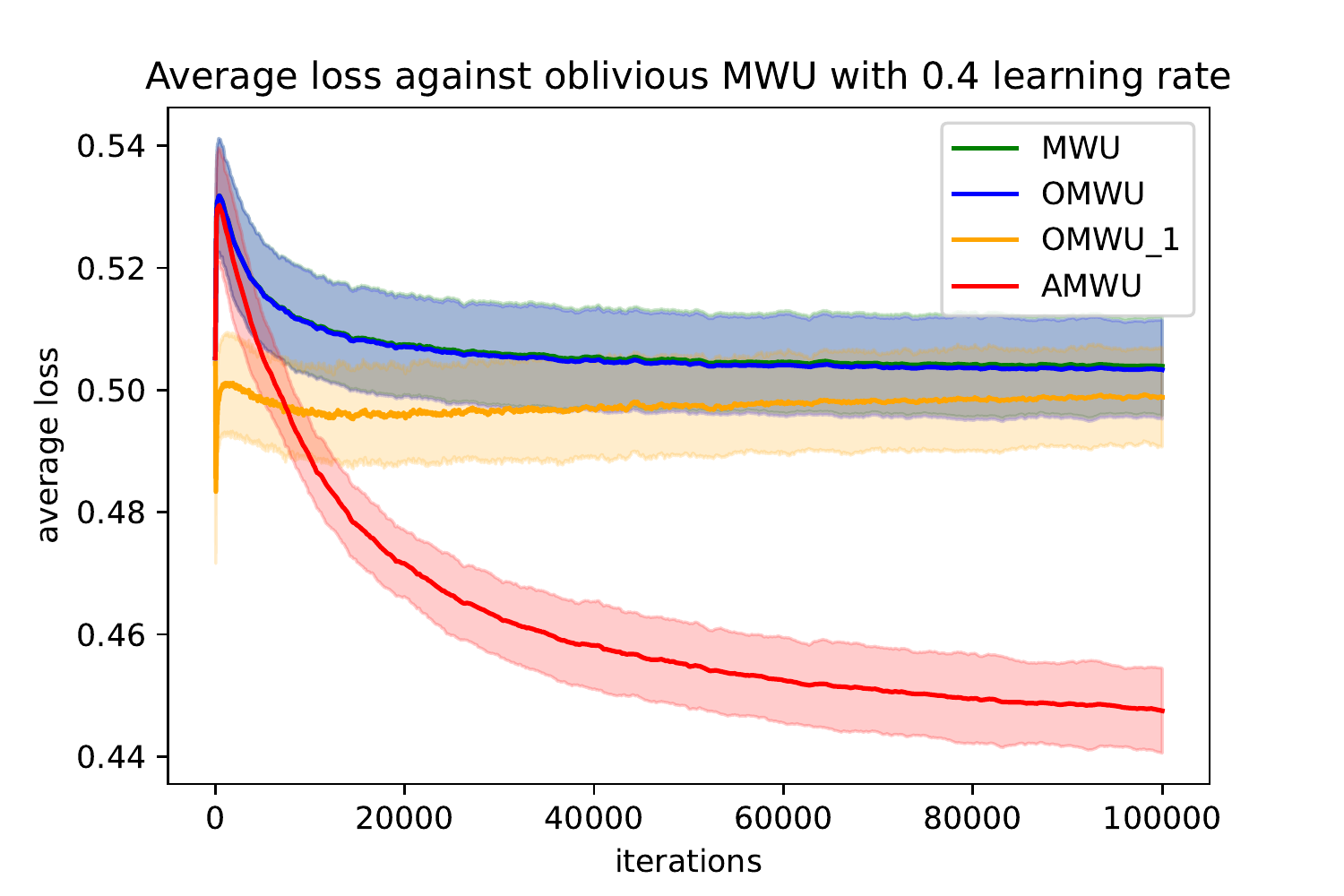}
                
\caption{0.4 learning rate MWU adversary}
\end{subfigure}
\begin{subfigure}[l]{.49\textwidth}
         \centering
         \includegraphics[width=1.\textwidth]{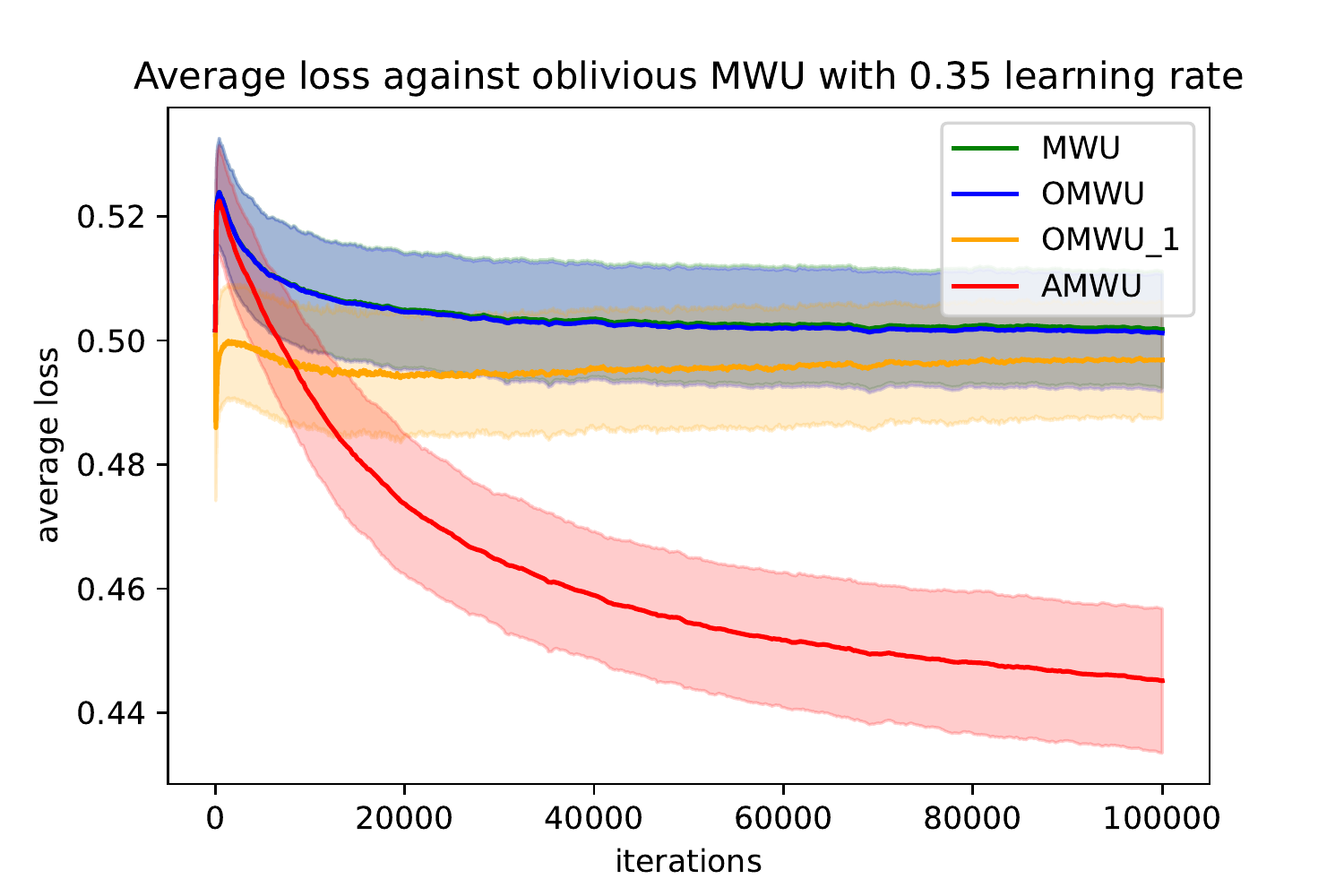}
                
\caption{0.35 learning rate MWU adversary}
\end{subfigure}
\begin{subfigure}[l]{.49\textwidth}
         \centering
         \includegraphics[width=1.\textwidth]{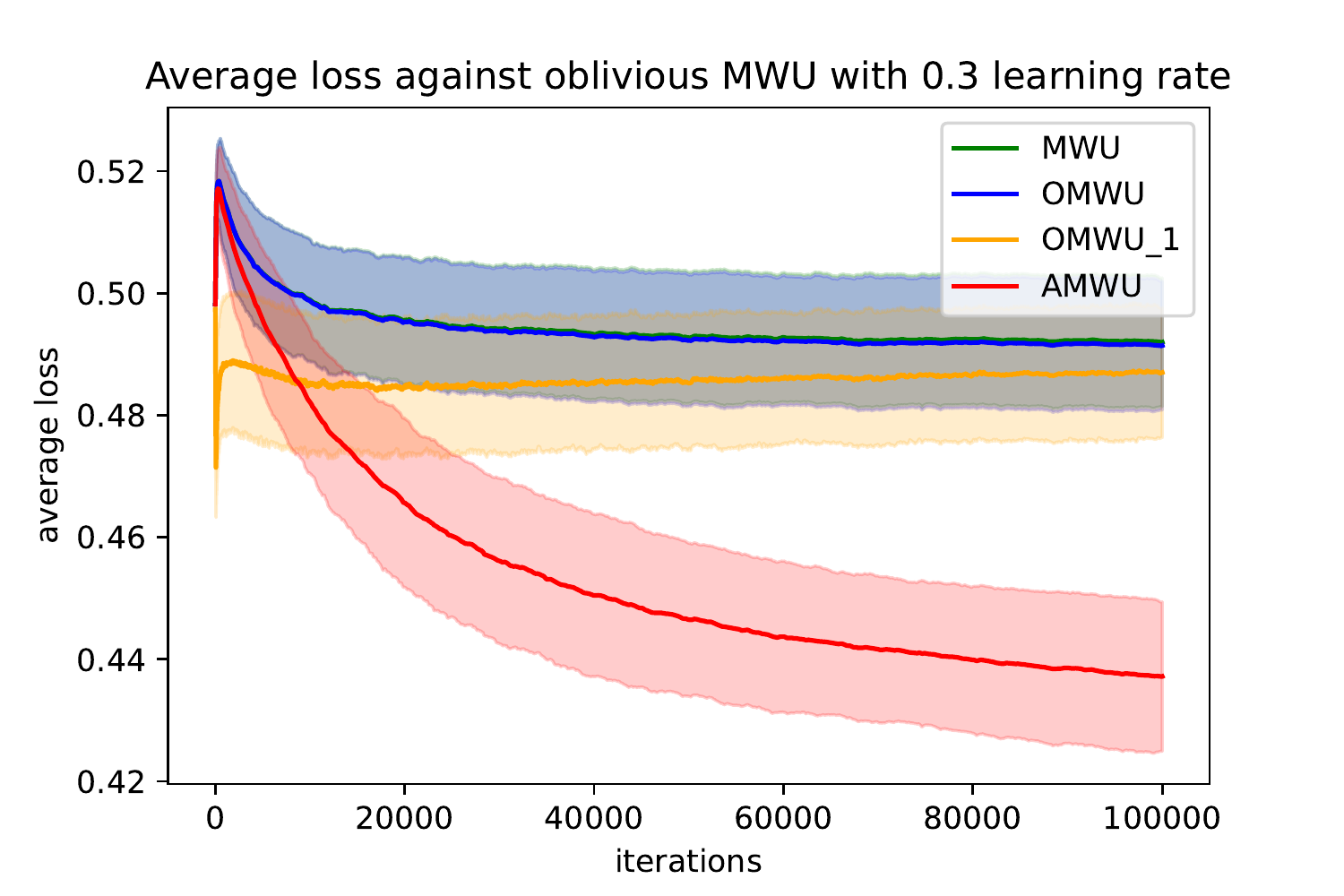}
                
\caption{0.3 learning rate MWU adversary}
\end{subfigure}
\begin{subfigure}[l]{.49\textwidth}
         \centering
         \includegraphics[width=1.\textwidth]{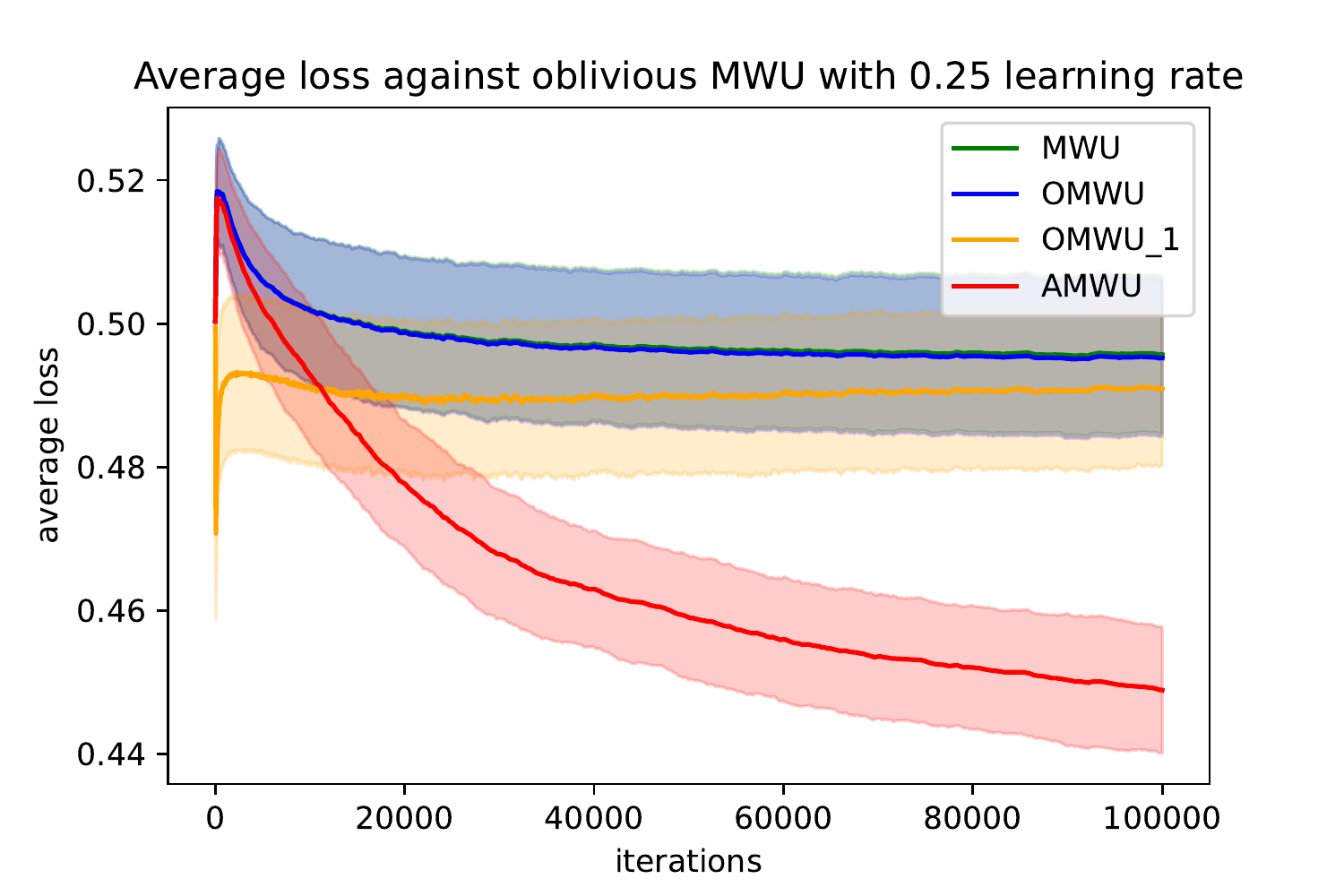}
                
\caption{0.25 learning rate MWU adversary}
\end{subfigure}
\caption{Against different Oblivious MWU adversary in random games}
\label{average loss for random games: 1}
\vspace{-10pt}
\end{figure*}

\begin{figure*}[t!]
     \centering
\begin{subfigure}[l]{.49\textwidth}
         \centering
         \includegraphics[width=1.\textwidth]{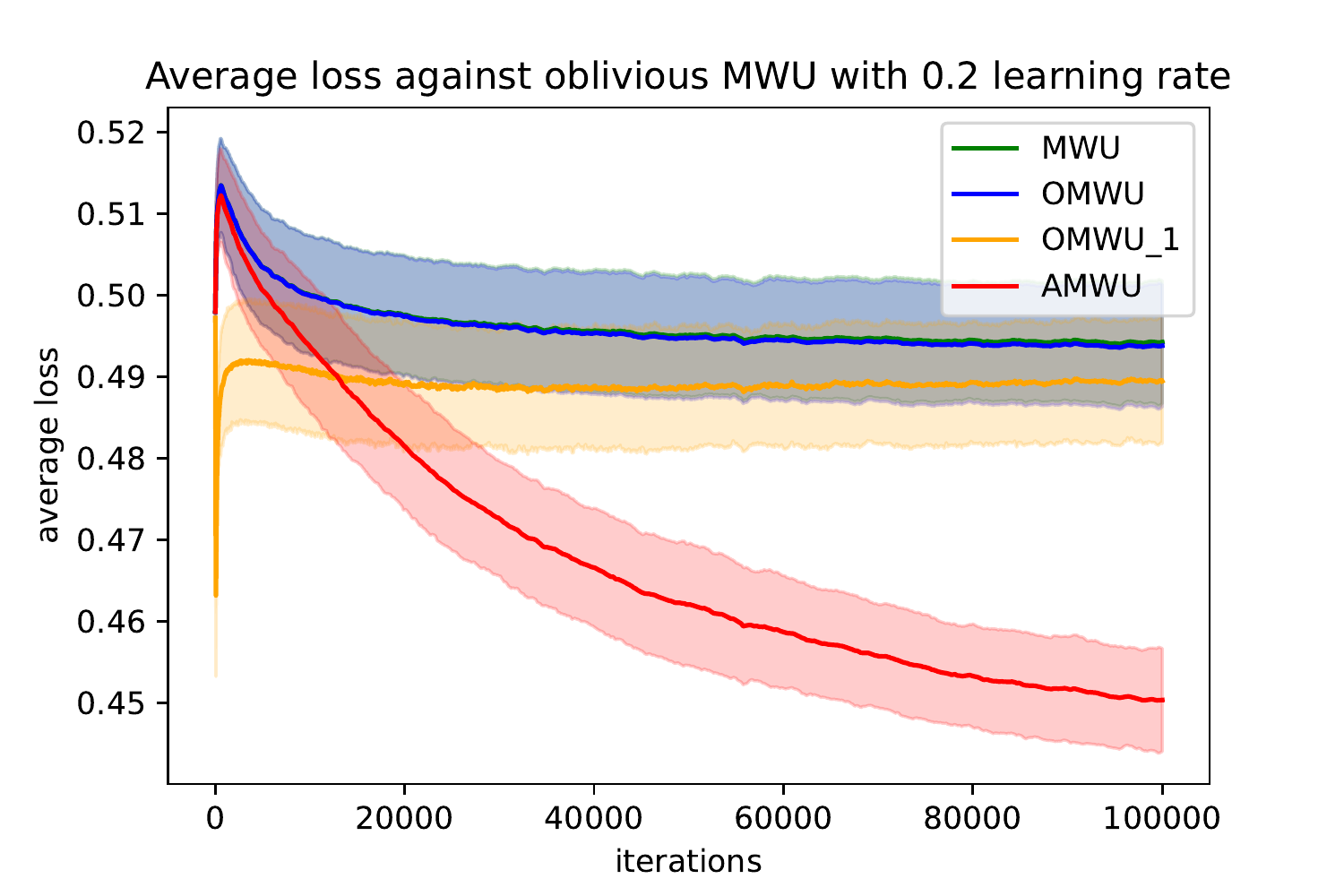}
                
\caption{0.2 learning rate MWU adversary}
\end{subfigure}
\begin{subfigure}[l]{.49\textwidth}
         \centering
         \includegraphics[width=1.\textwidth]{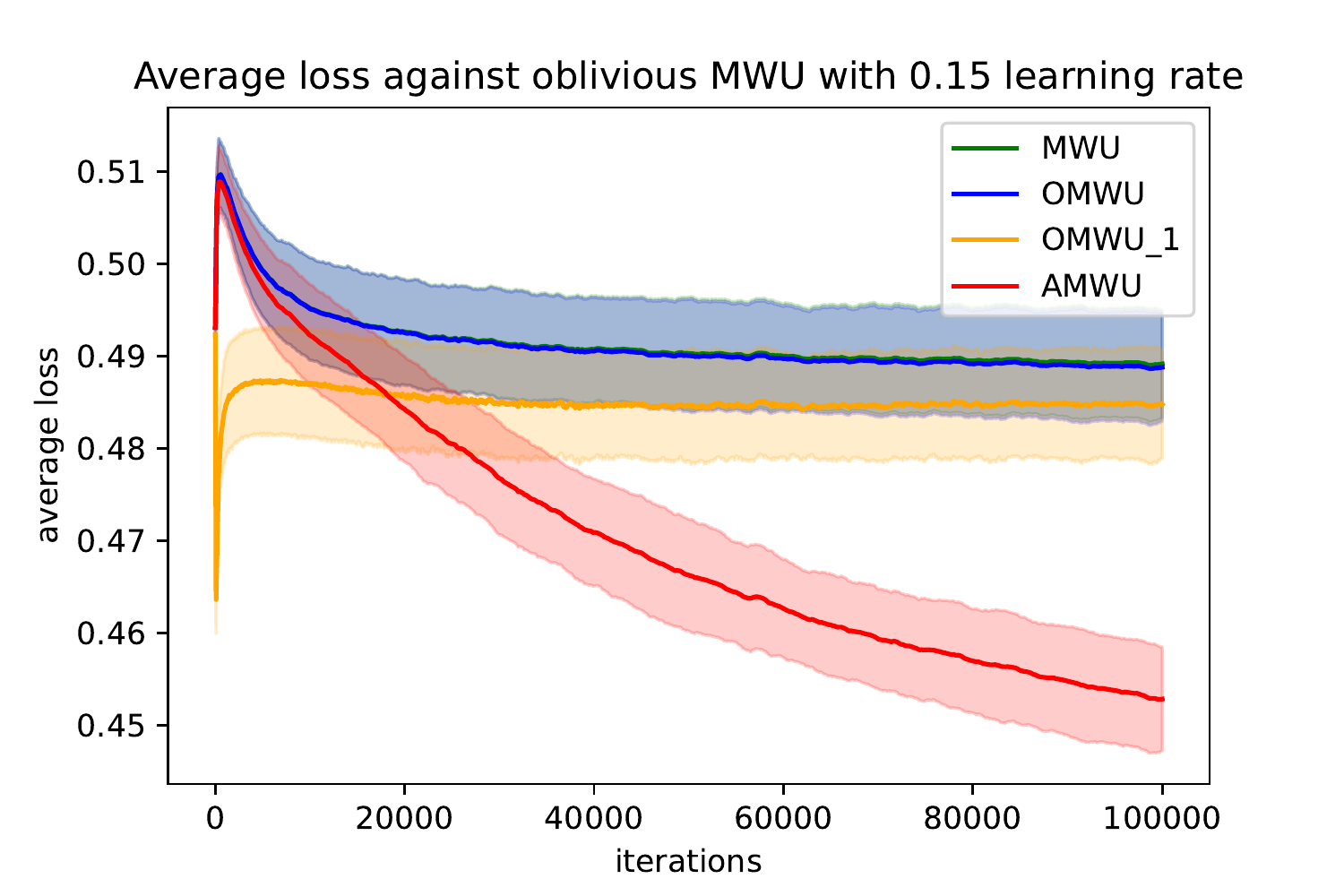}
                
\caption{0.15 learning rate MWU adversary}
\end{subfigure}
\begin{subfigure}[l]{.49\textwidth}
         \centering
         \includegraphics[width=1.\textwidth]{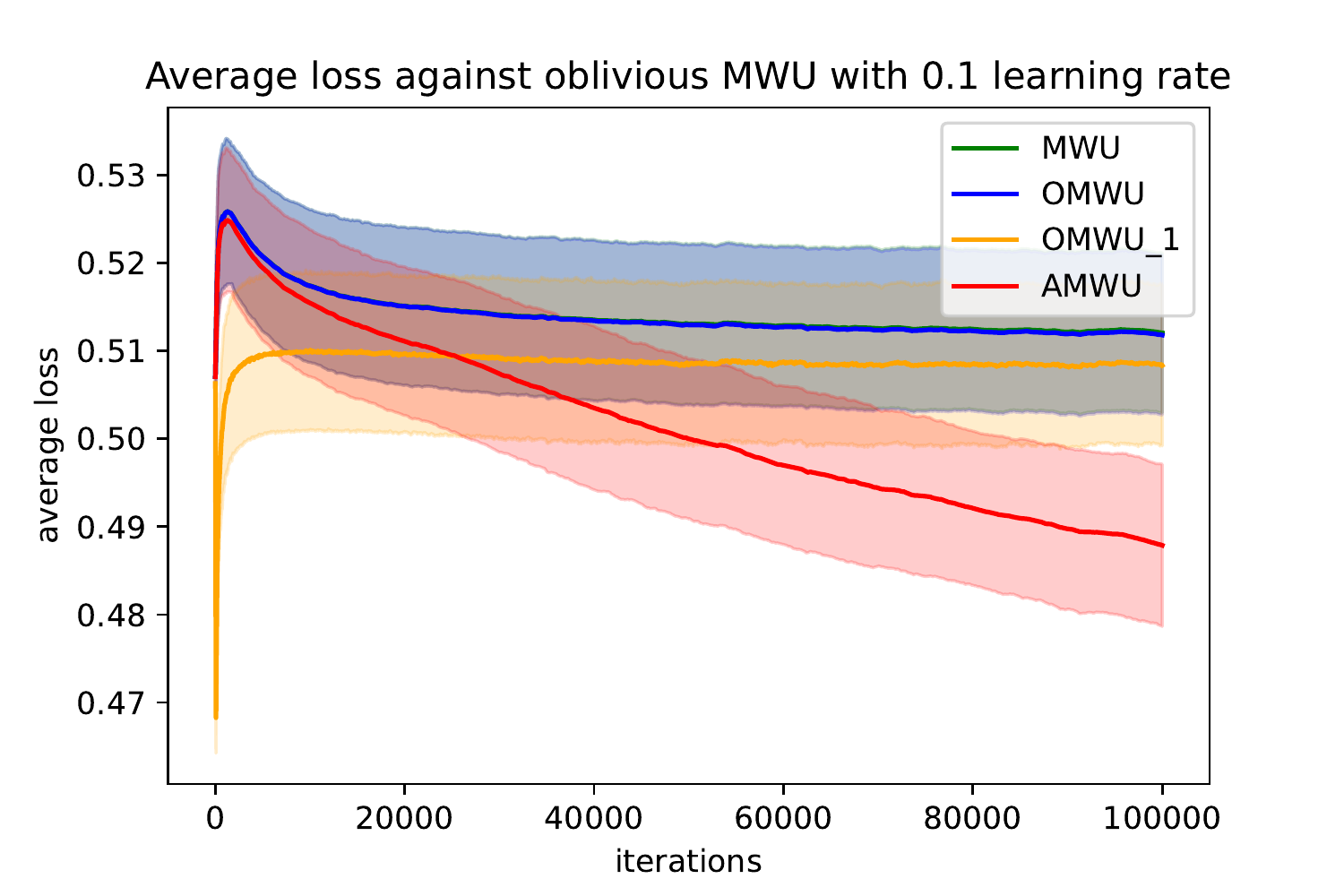}
                
\caption{0.1 learning rate MWU adversary}
\end{subfigure}
\begin{subfigure}[l]{.49\textwidth}
         \centering
         \includegraphics[width=1.\textwidth]{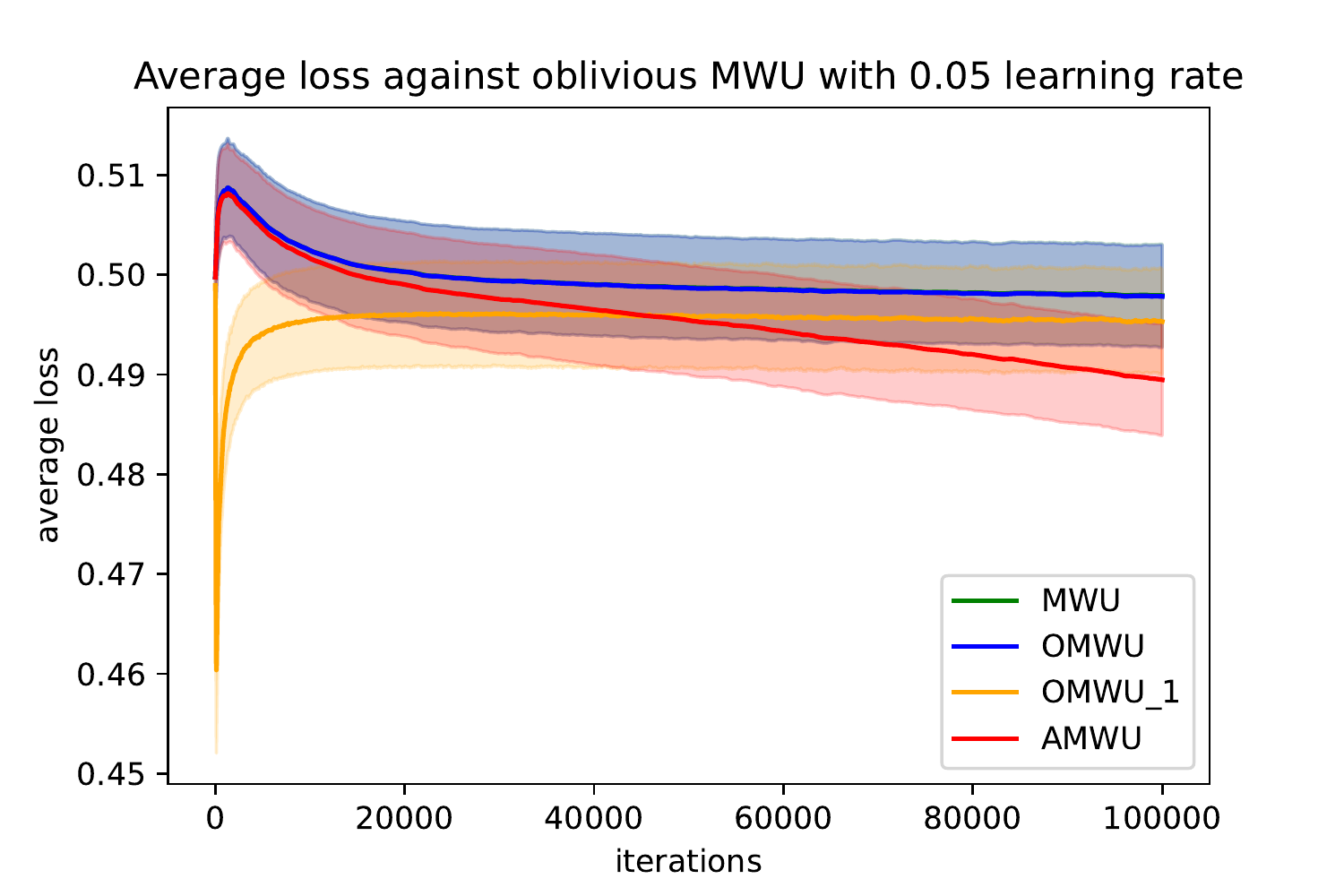}
                
\caption{0.05 learning rate MWU adversary}
\end{subfigure}
\caption{Against different Oblivious MWU adversary in random games}
\label{average loss for random games: 2}

\end{figure*}

\begin{figure*}[t!]
     \centering
\begin{subfigure}[l]{.49\textwidth}
         \centering
         \includegraphics[width=1.\textwidth]{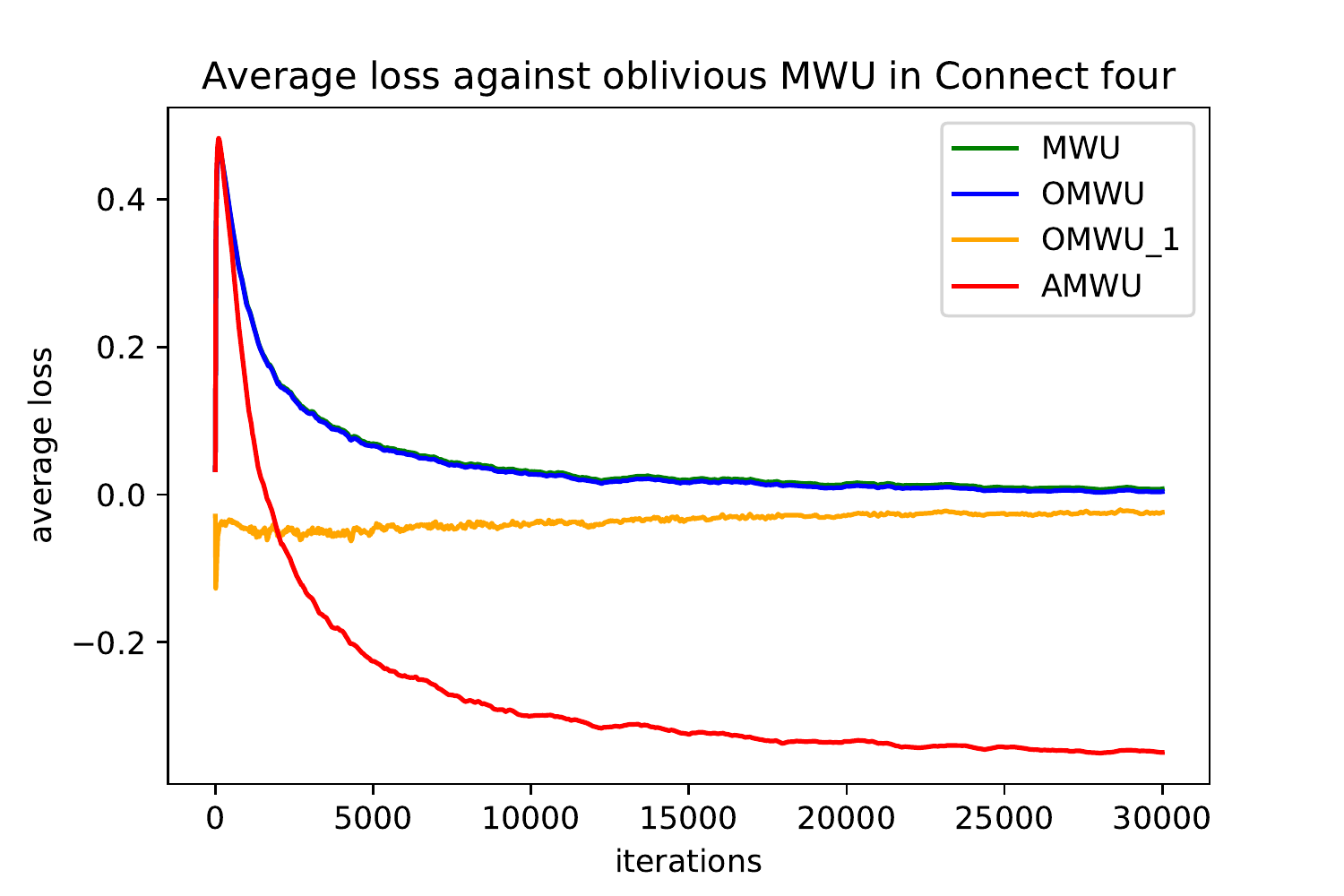}
\caption{Against MWU adversary in Connect Four}
        
\end{subfigure}
\begin{subfigure}[l]{.49\textwidth}
         \centering
         \includegraphics[width=1.\textwidth]{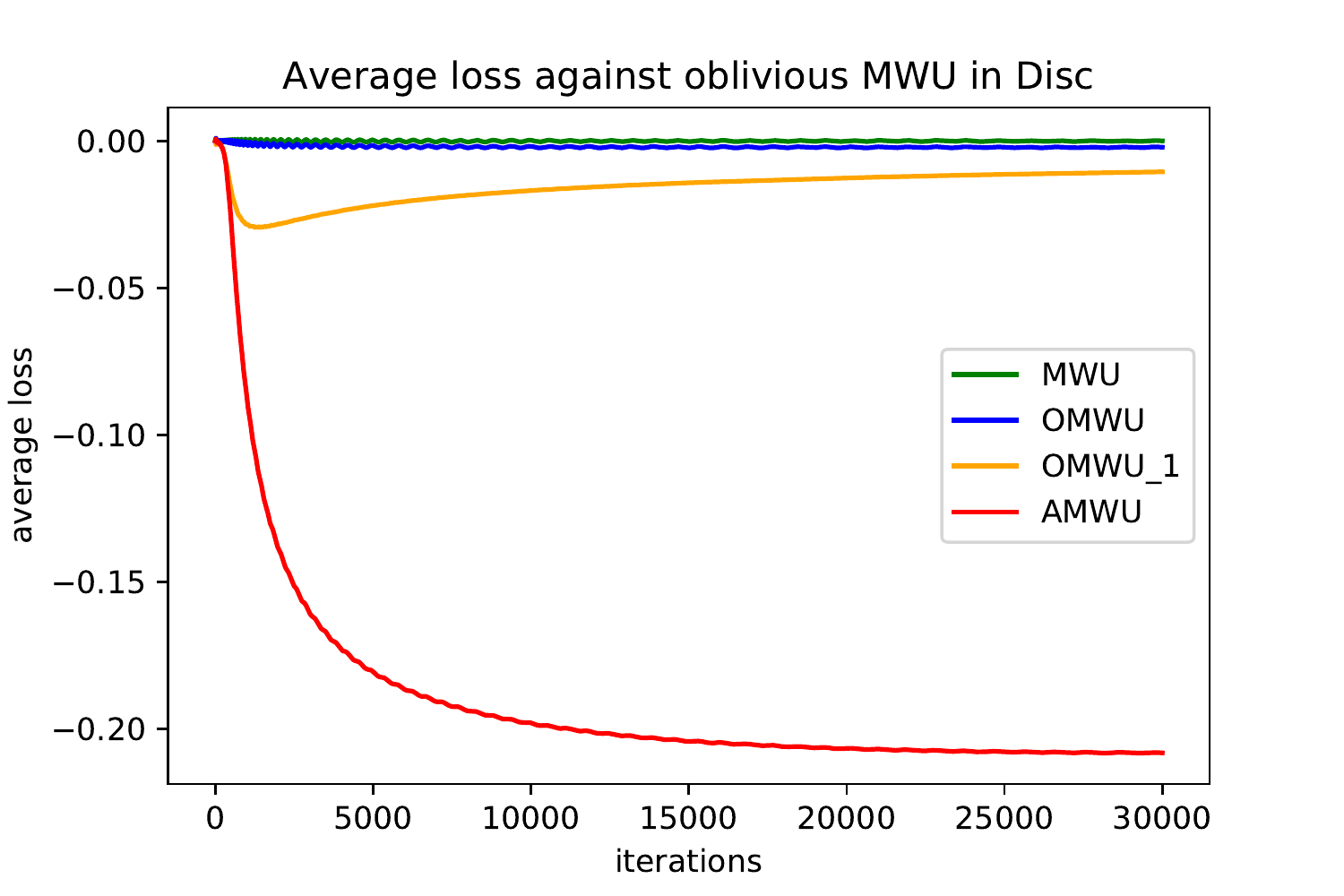}
                
\caption{Against MWU adversary in Disc}
\end{subfigure}
\caption{Against Oblivious MWU adversary in meta games}
\label{average performance in meta game}
\vspace{-10pt}

\end{figure*}

\textbf{Average performance against oblivious adversary:} we report performance of AMWU and other baselines against different oblivious adversaries, i.e., the MWU adversary with different learning rate $[0.5, 0.45, 0.4.\dots,0.05]$. As we can see in Figure \ref{average loss for random games: 1} and Figure \ref{average loss for random games: 2}, AMWU outperfoms other baselines by a large margin across all the adversary setting in random matrix games. A similar trend can be observed in the Connect Four and Disc experiments in Figure \ref{average performance in meta game}.
\subsection{Last round convergence of AMWU}\label{Appendix: exp Last round convergence}
For a fair comparison, we set up a common learning rate for our algorithm AMWU and the baselines MWU and OMWU. In the experiments of average performance, we first set the common learning rate $\eta=0.01$ and the exploiting rate $\alpha=100$. In order to highlight the difference between AMWU and OMWU, we also test the performance of OMWU with learning rate $\eta=1$. That is, the OWMU with the same relative weight between the predictable sequence $\vx_{t-1}$ and the regularizer $R(\vf)$ as AMWU (i.e., $\eta_{OMWU}=\eta_{AMWU} \times \alpha_{AMWU}$). In the experiments of last round convergence, we vary the common learning rate $\eta$ (i.e., $\eta=[0.01, 0.025, 0.05]$) to see whether the convergence trend we see is robust against the learning rate. In here we focus on the random matrix games ($20 \times 20$ and $50 \times 50$ dimensions) due to its nice property of unique Nash Equilibrium, which AMWU and OMWU require to convergence. Since there is no guarantee of convergence of OMWU with a large learning rate (e.g., $\eta=1$), we do not consider $OMWU_1$ as a baseline in this experiment.

\begin{figure*}[t!]
     \centering
\begin{subfigure}[l]{.49\textwidth}
         \centering
         \includegraphics[width=1.\textwidth]{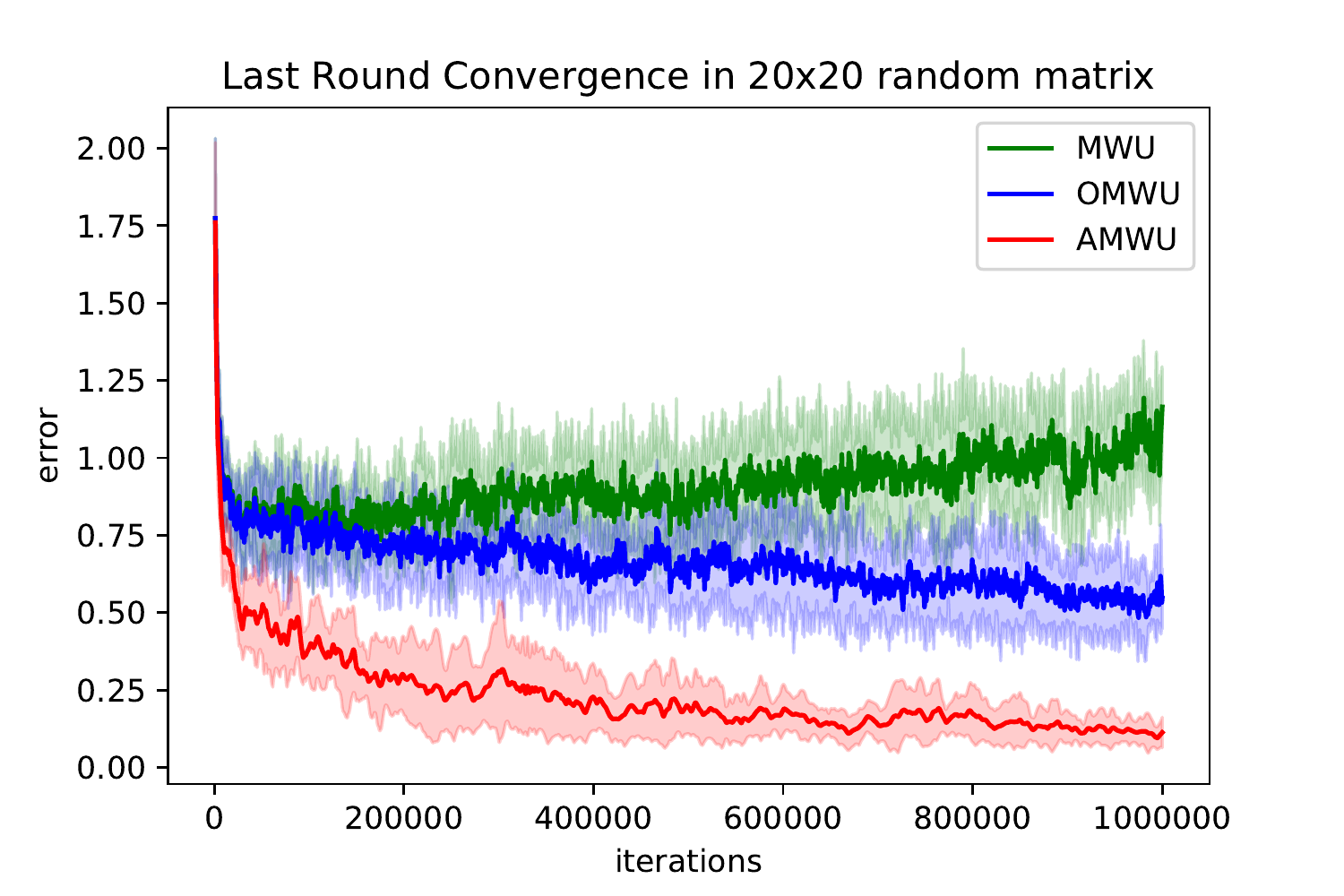}
\caption{$20 \times 20$ random games}
\end{subfigure}
\begin{subfigure}[l]{.49\textwidth}
         \centering
         \includegraphics[width=1.\textwidth]{figures/last_round_convergence_random_matrix_50x50.pdf}
                
\caption{$50 \times 50$ random games}
\end{subfigure}
\caption{Last round convergence in random games with 0.01 learning rate}
\label{fig: last round convergence in random games with 0.01 learning rate}
\vspace{-10pt}
\end{figure*}

\begin{figure*}[t!]
     \centering
\begin{subfigure}[l]{.49\textwidth}
         \centering
         \includegraphics[width=1.\textwidth]{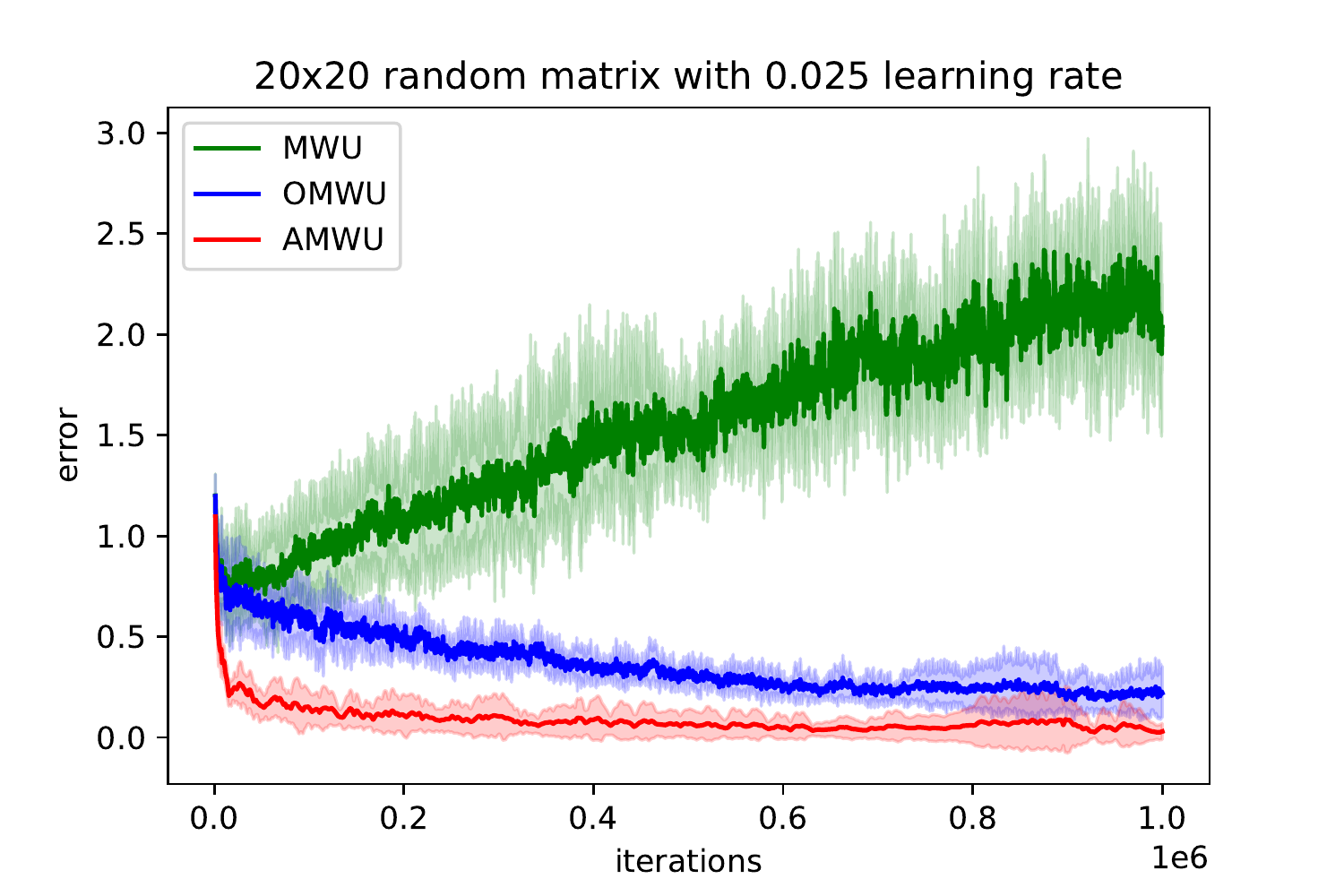}
\caption{$20 \times 20$ random games}
\end{subfigure}
\begin{subfigure}[l]{.49\textwidth}
         \centering
         \includegraphics[width=1.\textwidth]{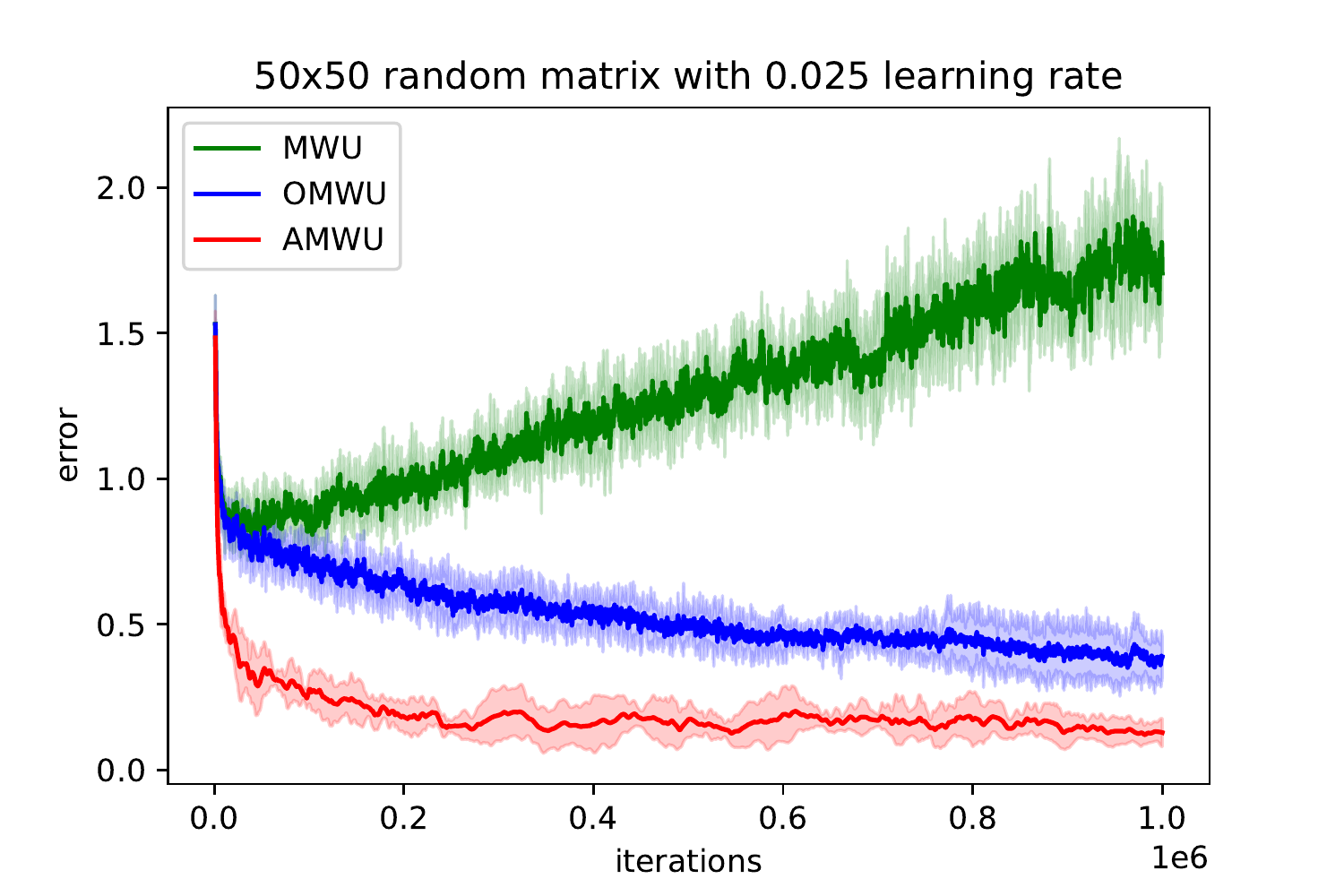}
                
\caption{$50 \times 50$ random games}
\end{subfigure}
\caption{Last round convergence in random games with 0.025 learning rate}
\label{fig: last round convergence in random games with 0.025 learning rate}

\vspace{-10pt}
\end{figure*}

\begin{figure*}[t!]
     \centering
\begin{subfigure}[l]{.49\textwidth}
         \centering
         \includegraphics[width=1.\textwidth]{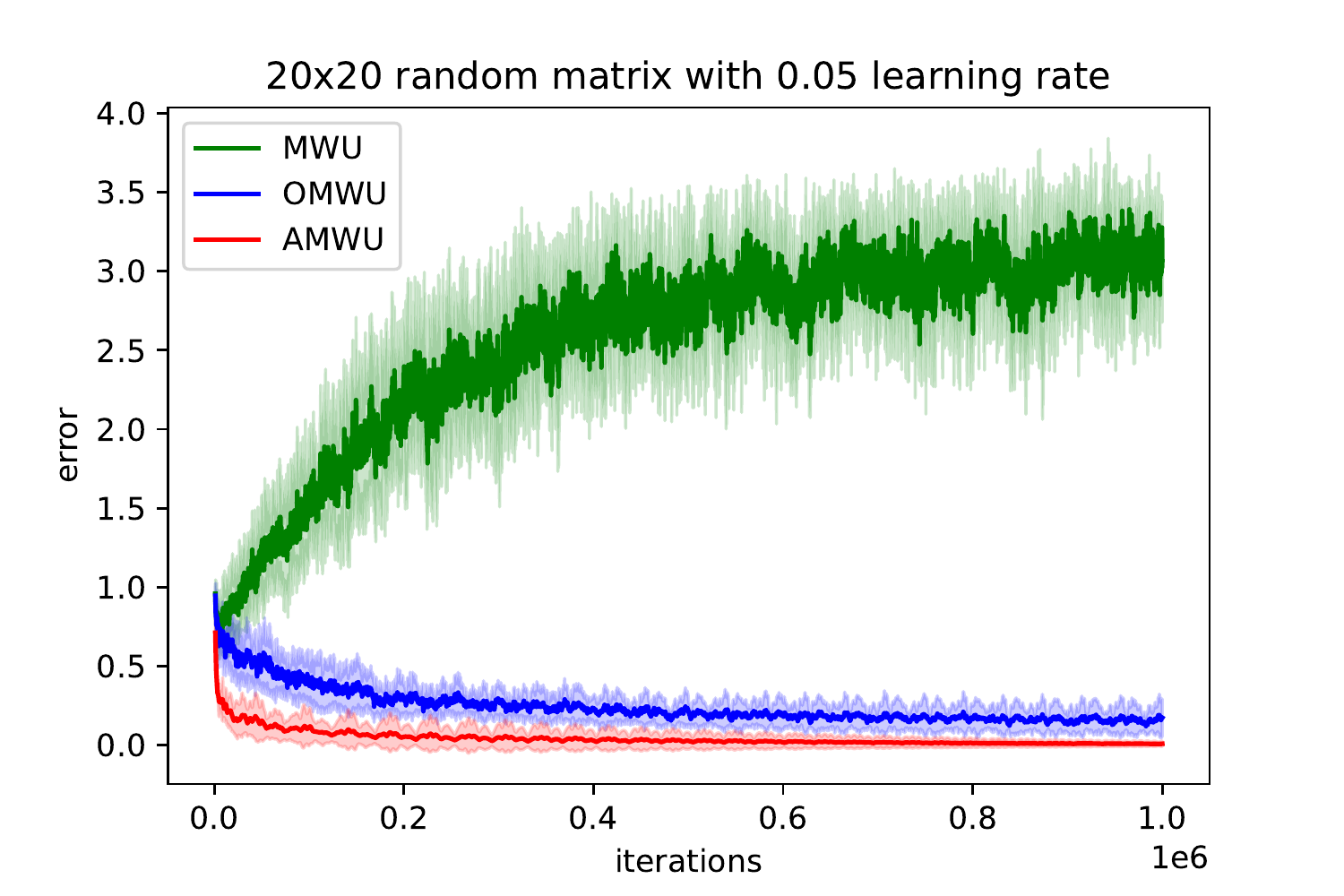}
\caption{$20 \times 20$ random games}
\end{subfigure}
\begin{subfigure}[l]{.49\textwidth}
         \centering
         \includegraphics[width=1.\textwidth]{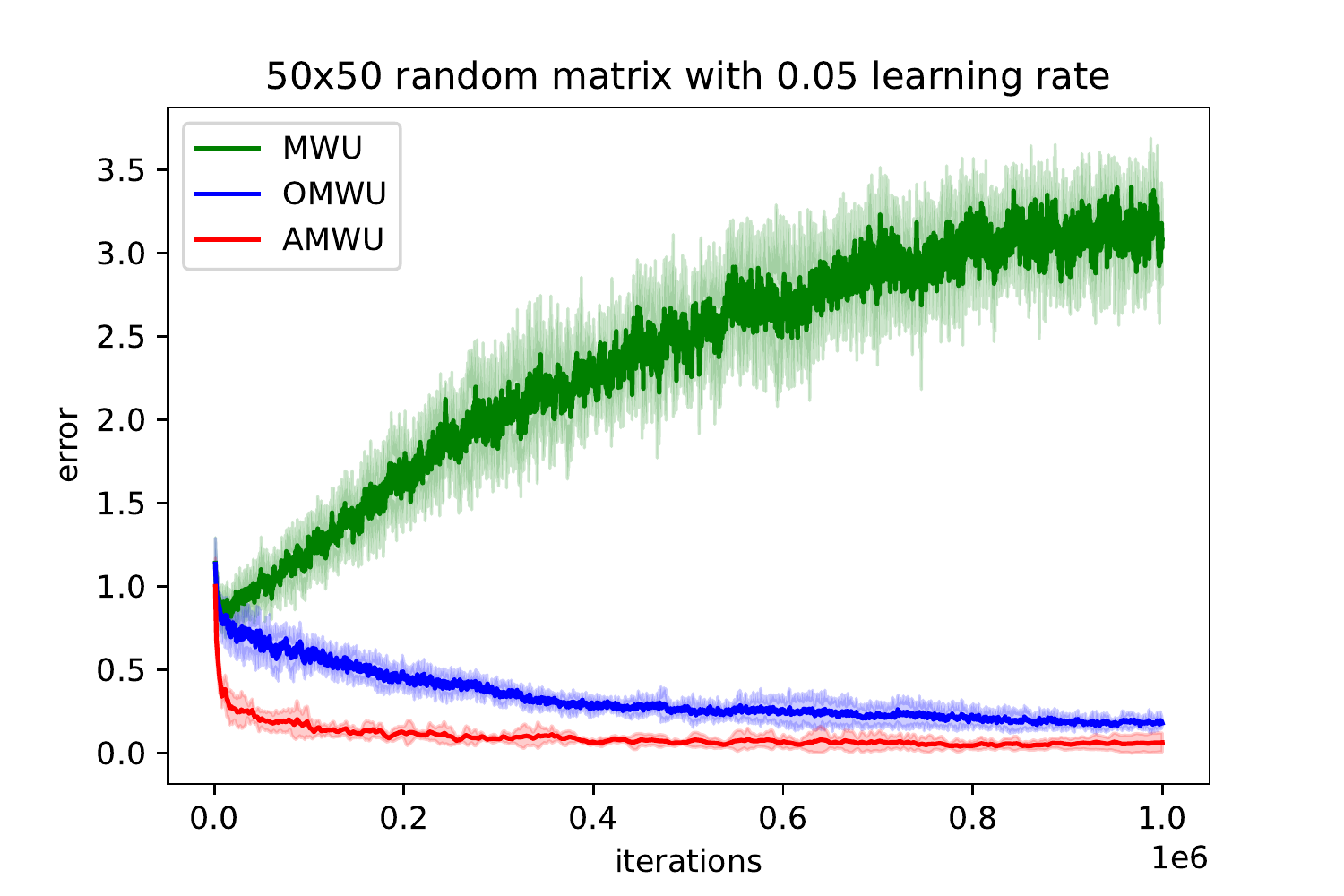}
                
\caption{$50 \times 50$ random games}
\end{subfigure}
\caption{Last round convergence in random games with 0.05 learning rate}
\label{fig: last round convergence in random games with 0.05 learning rate}
\vspace{-10pt}
\end{figure*}

\begin{figure*}[t!]
     \centering
\begin{subfigure}[l]{.49\textwidth}
         \centering
         \includegraphics[width=1.\textwidth]{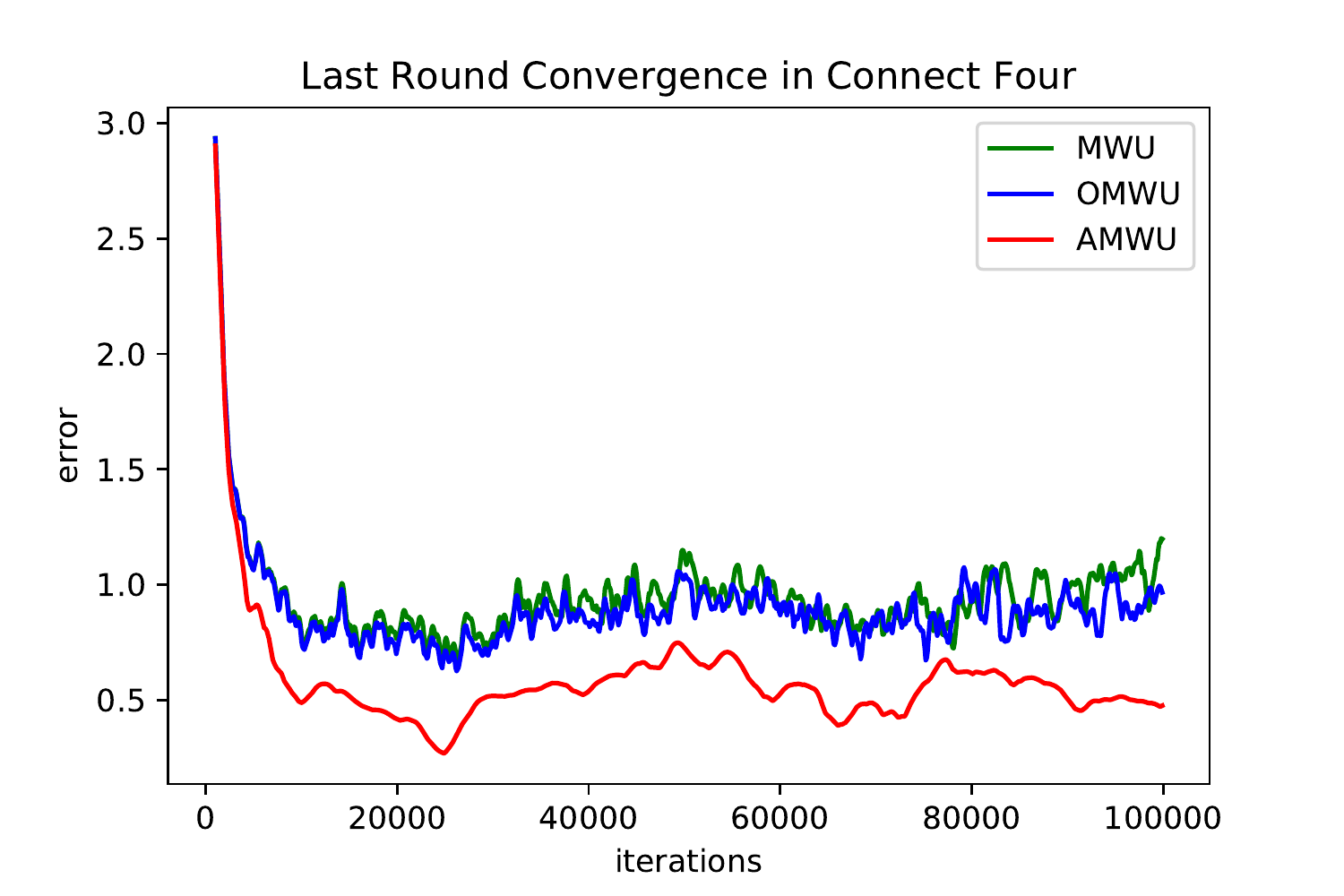}
\caption{Connect Four meta game}
\end{subfigure}
\begin{subfigure}[l]{.49\textwidth}
         \centering
         \includegraphics[width=1.\textwidth]{figures/last_round_convergence_Disc.pdf}
                
\caption{Disc meta game}
\end{subfigure}
\caption{Last round convergence in meta games}
\label{Last round convergence in meta games}
\vspace{-10pt}
\end{figure*}

\textbf{Last round convergence in self-play:} we report the performance of AMWU and other baselines in self-play setting. As we can see in Figure \ref{fig: last round convergence in random games with 0.01 learning rate}, Figure \ref{fig: last round convergence in random games with 0.025 learning rate} and Figure \ref{fig: last round convergence in random games with 0.05 learning rate}, AMWU outperforms OWMU and MWU by a large margin across all the 3 different learning rate setting. The MWU shows divergence in last round convergence in as expected in \cite{Bailey2018}. A similar trend can be observed in the Connect Four and Disc experiments in Figure \ref{Last round convergence in meta games}.


\end{document}